\pdfoutput=1
\documentclass[11pt]{article}
\usepackage{float}
\usepackage{acl}
\usepackage[utf8]{inputenc}

\usepackage{pgfplots}
\pgfplotsset{width=10cm,compat=1.9}
\usepackage[]{algpseudocode}
\usepackage[]{algorithm}
\algtext*{EndFor}%
\algtext*{EndProcedure}%
\DeclareUnicodeCharacter{2212}{−}
\usepgfplotslibrary{groupplots,dateplot}
\usetikzlibrary{patterns,shapes.arrows}
\pgfplotsset{compat=newest}

\usepackage{newfloat}
\usepackage{listings}
\floatstyle{ruled}
\newfloat{listing}{tb}{lst}{}
\floatname{listing}{Listing}

\usepackage{tikzscale} 
\usepackage{tabularx}
\usepackage{booktabs}
\usepackage{xcolor}
\usepackage{amsmath}
\usepackage{bbm}
\newcommand{\probP}{\text{I\kern-0.15em P}}
\usepackage{color}
\usepackage{amssymb}
\usepackage{pifont}
\usepackage{tabularx,booktabs}
\usepackage{enumitem}
\usepackage{placeins}
\usepackage[normalem]{ulem}
\usepackage{ragged2e}
\usepackage{cleveref}
\usepackage{microtype}
\usepackage{dsfont}
\usepackage{scalerel,xparse}
\usepackage{times}
\usepackage{latexsym}
\usepackage{adjustbox}
\usepackage{subfigure}
\usepackage{bbm}
\usepackage{twemojis}

\usepackage{CJKutf8}
\usepackage[T1]{fontenc}
\usepackage[russian,english]{babel}

\usepackage{soul}
\definecolor{lightyellow}{RGB}{255, 242, 176}
\sethlcolor{lightyellow}

\usepackage{multirow}
\usepackage{fontawesome5}
\usepackage{xspace}
\usepackage{todonotes}
\usepackage[most]{tcolorbox}
\usepackage[table]{xcolor}
\usepackage{array}
\usepackage{colortbl}
\usepackage{makecell}
\usepackage{graphicx}

\usepackage{caption}

\newcommand{\cmark}{\ding{51}} 
\newcommand{\xmark}{\ding{55}} 

\definecolor{zoey green}{rgb}{0.684,0.836,0.227}
\definecolor{zoey green}{RGB}{102, 158, 212}

\newcommand{\ignore}[1]{}

\definecolor{nns-self}{HTML}{F5C857}
\definecolor{nns-mcq}{HTML}{F39EB6}
\definecolor{ns-eval}{HTML}{A594F9}

\definecolor{boxteal}{HTML}{B7E2DA}
\definecolor{inlineteal}{HTML}{91C4C3}

\definecolor{bggray}{rgb}{0.95, 0.95, 0.95}
\usepackage[%
    framemethod=tikz,
    skipbelow=\topskip,
    skipabove=\topskip
]{mdframed}
\mdfsetup{%
    leftmargin=0pt,
    rightmargin=0pt,
    backgroundcolor=bggray,
    middlelinecolor=black,
    roundcorner=3
}
\newtcolorbox[list inside=prompt,auto counter,number within=section]{prompt}[1][]{
    colbacktitle=black!60,
    fonttitle=\small,
    coltitle=white,
    fontupper=\footnotesize,
    boxsep=4pt,
    left=0pt,
    top=0pt,
    bottom=0pt,
    boxrule=1pt,
    #1,
}

\makeatletter
\newcommand\starfootnote[1]{%
  \begingroup
    \renewcommand\thefootnote{*}%
    \let\@makefnmark\relax        
    \footnote{#1}%
    \addtocounter{footnote}{-1}%
  \endgroup
}
\makeatother

\title{\textit{Reheat Nachos} for Dinner? \\ Evaluating AI Support for Cross-Cultural Communication of Neologisms}

\author{\textbf{Dayeon Ki$^*$}      
        \hspace{0.5cm} 
        \textbf{Yu Hou$^*$} 
        \hspace{0.5cm} \\
        \textbf{Rachel Rudinger}
        \hspace{0.5cm}
        \vspace{0.1cm}
        \textbf{Hal Daumé III}
        \hspace{0.5cm} 
        \textbf{Marine Carpuat}
        \hspace{0.5cm} 
        \textbf{Fumeng Yang}
        \vspace{0.1cm}  \\
  University of Maryland \\
  \texttt{\{dayeonki,houyu\}@umd.edu}
}

\newif\ifnotes
\notestrue 

\definecolor{fycolor}{HTML}{9342c9}

\begin{document}
\maketitle

\begin{abstract}

Neologisms and emerging slang are central to daily conversation, yet challenging for non-native speakers (NNS) to interpret and use appropriately in cross-cultural communication with native speakers (NS). 
NNS increasingly make use of Artificial Intelligence (AI) tools to learn these words.
We study the utility of such tools in mediating an informal communication scenario through a human-subjects study ($N$=234): NNS participants learn English neologisms with AI support, write messages using the learned word to an NS friend, and judge contextual appropriateness of the neologism in two provided writing samples.
Using both NS evaluator-rated communicative competence of NNS-produced writing and NNS' contextual appropriateness judgments, we compare three AI-based support conditions:
AI Definition, AI Rewrite into simpler English, AI Explanation of meaning and usage, and Non-AI Dictionary for comparison.
We show that AI Explanation yields the largest gains over no support in NS-rated competence, while contextual appropriateness judgments show indifference across support.
NNS participants' self-reported perceptions tend to overestimate NS ratings, revealing a mismatch between perceived and actual competence.
We further observe a significant gap between NNS- and NS-produced writing, highlighting the limitations of current AI tools and informing design for future tools.\starfootnote{*: Equal contribution; \textit{Reheat nachos} refers to producing a lesser version of an earlier song or album; illustrates one of a failure case of our participant's incorrect usage.}\footnote{\url{https://github.com/dayeonki/crosscultural_communication }}


\end{abstract}

\section{Introduction}

Picture this: You are an international student who learned English as a second language.
One day, a native English-speaking friend turns to you: ``\textit{When you're running in the rain and the main character energy starts to hit}.'' You know every word---but what does it actually \textit{mean}? How do you respond?

This scenario illustrates a broader challenge in \textit{informal} cross-cultural communication, between non-native English speakers (NNS) \raisebox{-0.2em}{\includegraphics[height=1.1em]{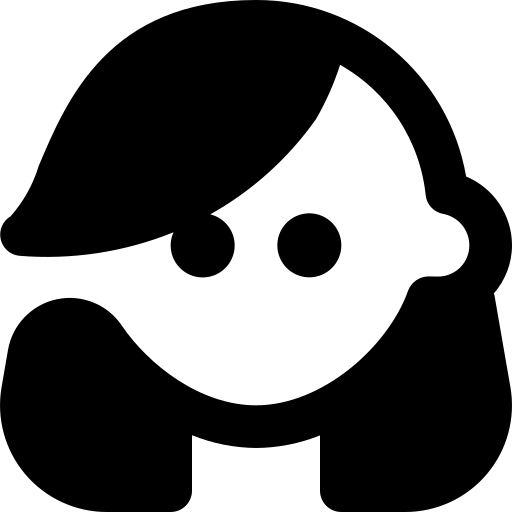}} and native English speakers (NS) \raisebox{-0.2em}{\includegraphics[height=1.1em]{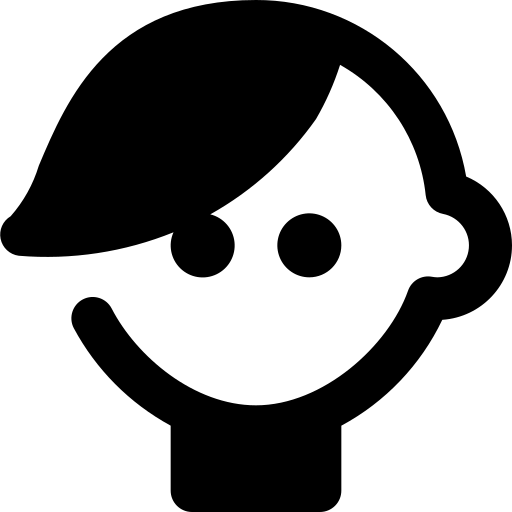}}.\footnote{The visual distinction between the two icons is intended solely to aid readability and carries no implication of gender.}
Contemporary English is rich with neologisms\textemdash newly coined expressions (e.g., main character energy\footnote{This refers to dramatic self-confidence or self-importance.}) or existing forms that have acquired new meanings (e.g., Ohio\footnote{This is often used to describe something that is weird, awkward, or cringeworthy.}) \citep{newmark1988textbook}, particularly in informal, slang, and internet-mediated discourse \citep{mattiello2005pervasiveness}.
These encode practices and cultural contexts, often to fill lexical gaps within a particular speech community \citep{santhi2010translating, khan2013neologisms}. 
As such, these expressions offer insights into everyday experiences within the communities that use them \citep{mcdonald2005meaning}. This sociocultural richness is precisely what makes these expressions challenging for NNS to understand and use appropriately\textemdash yet it is also what makes them appealing to learn.\footnote{As an illustrative example, one study found that 98\% of NNS surveyed reported positive attitudes toward including English neologisms in language lessons \citep{Rets2016TeachingNeologisms}.}


Traditional learning resources, such as printed dictionaries and textbooks, often fail to capture these rapidly evolving, context-dependent meanings, leading NNS to seek alternative support.
They increasingly turn to AI tools for support~\citep{tamkin2024clioprivacypreservinginsightsrealworld}, which have become go-to resources for language learners \citep{singh-etal-2024-translating, saha-etal-2025-reading}, who use them to request definitions, rewrite in simpler words, or ask for explanations.

\begin{figure*}
    \centering
    \includegraphics[width=0.95\linewidth]{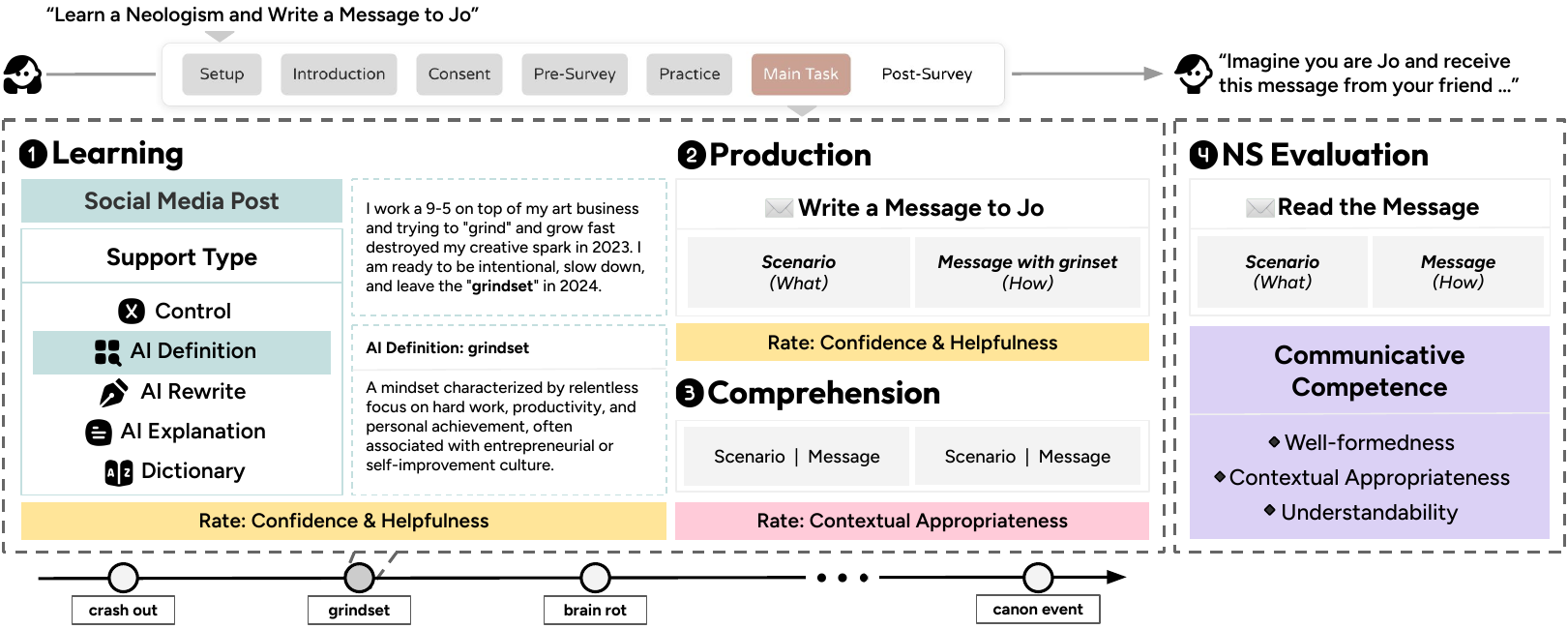}
    \caption{\textbf{Overview of our study design.} 
    We set up the communication scenario by having NNS participants learn neologisms and write messages to a hypothetical NS friend, \textit{Jo}. During the main task, for each of the eight neologisms, they complete a three-stage procedure: \textbf{\ding{202} Learning:} Learn the neologism within a social media post with one of five randomly assigned support types (one control and four treatment); \textbf{\ding{203} Production:} Write a scenario and a message to \textit{Jo} using the learned neologism; \textbf{\ding{204} Comprehension:} Rate the contextual appropriateness of the neologism in two provided writing samples. Each NNS-produced writing sample is subsequently rated by two NS evaluators for communicative competence (\textbf{\ding{205} NS Evaluation}). We color-code the measures throughout the paper using \raisebox{3pt}{\colorbox{ns-eval}{ }} (NS-Rated Competence), \raisebox{3pt}{\colorbox{nns-mcq}{ }} (NNS Comprehension Competence), and \raisebox{3pt}{\colorbox{nns-self}{ }} (NNS Self-Reported Perceptions). 
    }
    \label{fig:main_fig}
\end{figure*}


However, existing research has not kept pace with this real-world adoption. 
Prior studies evaluating AI tools' capability to understand or process neologisms remain in constrained evaluation formats such as multiple-choice questions \citep{deng2024newterm}. 
These settings are far removed from how users actually engage with AI tools when encountering neologisms \citep{liao2023rethinking}. Consequently, it remains unclear how useful such AI tools truly are in supporting \textit{real} users navigating the cross-cultural communication of neologisms.

To address this gap, we conduct a between-subjects human study with 234 NNS participants \raisebox{-0.2em}{\includegraphics[height=1.1em]{figure/logo/nns.png}} whose native languages are Spanish, German, or Chinese, along with 144 NS evaluators \raisebox{-0.2em}{\includegraphics[height=1.1em]{figure/logo/ns.png}} residing in the United States. We simulate a real-world scenario in which an NNS encounters a neologism in a social media post, learns the term with different types of support conditions, and uses it to communicate with an NS friend (\autoref{fig:main_fig}). 
Motivated by how users typically employ AI tools when encountering new words \citep{Xiao2023ChatGPT_EFL, Klimova2024ChatGPTLanguageEducation}, we design four support conditions: (1) AI Definition: providing a dictionary-style definition; (2) AI Rewrite: rewriting the social media post into simpler English; (3) AI Explanation: requesting an explanation of meaning and usage; and (4) Non-AI Dictionary: consulting an official online reference. 
We select eight neologisms from an online dictionary list.
For each neologism, NNS participants first learn it with randomly provided support, then complete tasks assessing their learning goals \citep{pickering2013integrated}: (\textbf{1}) \textbf{Production}, which they describe a brief scenario and write a message to an NS friend using the learned word, and (\textbf{2}) \textbf{Comprehension}, which they judge the contextual appropriateness of the word in two writing samples. 
Each NNS-produced writing from \textbf{Production} is subsequently rated by two NS evaluators along three dimensions of communicative competence: well-formedness, contextual appropriateness, and understandability \citep{Light1989CommunicativeCompetence}.

Our findings show that AI Explanation consistently helps NNS participants achieve higher NS-rated competence across all dimensions compared to no support. Comprehension task performance does not differ significantly across support conditions (\S\ref{res:1}).
We also examine the potential of using NNS' self-reported perceptions as proxies for NS-rated competence, given the practical challenges of directly involving NS in the evaluation.
However, what \textit{feels} communicatively effective from the NNS perspective generally overestimates how NS \textit{actually} interprets it (\S\ref{res:2}).
We further observe a significant gap between NNS-produced writing and those written by NS evaluators (\S\ref{res:3}). 
Comments from both parties indicate that NNS' lack of a mechanism to reliably assess imperfect AI outputs and limited contextualized support are potential causes of this gap (\S\ref{discussion:1}).
Taken together, our findings highlight limitations of current AI tools and inform design for future tools, including designing better support, training on data reflecting real user needs and usage, and communicating uncertainty to help NNS calibrate their reliance on AI (\S\ref{discussion:2}; \S\ref{discussion:3}). 

\section{Background \& Research Questions}

\subsection{Neologisms}

Neologisms are newly coined lexical units or existing forms that have acquired new meanings \citep{newmark1988textbook}. 
We focus on internet-derived neologisms (i.e., contemporary slang terms that emerge and evolve in informal discourse) due to both the challenges they pose \citep{charteris1998compound, reima_2010} and their appeal to NNS \citep{Rets2016TeachingNeologisms}.
Prior work in the context of AI has largely focused on detecting and collecting neologisms \citep{Nguyen2018UrbanDictionary, tomaszewska2025neontoolautomateddetection} or evaluating AI tools' capability to understand and process them \citep{zheng-etal-2024-neo, llm_chinese_neologism_2025}. 
However, despite the widespread use of AI by users for informal language learning \citep{terzimehić2025conversationalaicatalystinformal}, most existing benchmarks operationalize neologism understanding through constrained evaluation formats, such as multiple-choice questions \citep{deng2024newterm} or machine translation \citep{Awadh2020NeologismsTranslation, zheng-etal-2024-neo}.
These setups fail to adequately capture how users engage with AI tools \textit{in the wild} (i.e., limited ecological validity \citep{ethayarajh2020utility, liao2023rethinking}. As a result, it remains unclear how effectively AI supports users in learning neologisms for informal cross-cultural communication. Our work aims to address this gap.

\subsection{Cross-Cultural Communication}

Cross-cultural communication concerns the way people from different cultures interact \citep{tannen1983cross, hurn2013cross}. We define culture broadly to include \textit{any} communicative context in which interlocutors do \textbf{not} share a common \textbf{linguistic} or \textbf{cultural} background in a given context (e.g., a university lecturer teaching new undergraduate students, or interactions between a native and a non-native speaker) \citep{thomas_1983}. 
In such settings, the lack of shared common ground between interlocutors makes communication particularly challenging \citep{meyer2016culture, korkut2018study, carpuat-etal-2025-interdisciplinary}.
In response, lay users increasingly turn to AI tools to overcome these communicative gaps \citep{yue_2024, Sarwari2024AIInterculturalCommunication}. 
Prior work has proposed a range of interventions to bridge these gaps \citep{bourges1998meaning, heimgartner2017culturally}, including AI-generated explanations \citep{ki-etal-2025-share, saha-etal-2025-reading, zhao2026pragmaticsmeetscultureculturallyadapted} or question-answering chatbots \citep{Zhang_2025}.
However, human-centered evaluations of such interventions\textemdash particularly in \textit{informal} cross-cultural communication settings\textemdash remain underexplored.
Our work builds on this line of work with a specific focus on neologism communication between non-native and native English speakers.
We focus on English, as current AI tools provide stronger support for English than other languages~\citep{emily_parrot_2021} and English speakers are more readily accessible for participant recruitment.

\subsection{Communicative Competence}
\label{subsec:communicative_competence}

As defined by~\citet{Light1989CommunicativeCompetence}, communicative competence is the ability to be \textit{functionally adequate} in daily communication, encompassing sufficient \textit{knowledge}, \textit{judgment}, and \textit{skill} to communicate effectively. 
In this work, we focus on three widely recognized dimensions of communicative competence in the field of language development: (1) grammatical or discourse competence, (2) sociolinguistic competence \citep{hymes1972communicative}, and (3) strategic competence, referring to compensatory strategies for making the most of one's knowledge \citep{savignon1976communicative, canale-swain-1980-theoretical}.
Since communication effectiveness is context-dependent \citep{hymes1972communicative}, evaluations should be based on outcomes in realistic situations \citep{canale2014communicative}. 
Adopting this perspective, we simulate a cross-cultural communication scenario between NNS and NS and assess NNS' communicative competence through tasks reflecting real-world interactions, with outcomes subsequently rated by NS evaluators. 
Although directly involving NS is ideal, practical constraints (e.g., limited opportunities) make this challenging \citep{adaptive_second_2024}. Consequently, NNS often must rely on their own judgment to decide if they are confident in using the neologism in conversations. To approximate this, we also collect NNS self-reported perceptions as proxy measures of competence.

\subsection{Research Questions}
Informed by the literature, we formalize three research questions (RQs):
\begin{itemize}[leftmargin=*, itemsep=2pt, parsep=-1pt, topsep=0pt]
    \item[] \textbf{RQ1.} How effective are different support in helping NNS learn and communicate neologisms?
    \item[] \textbf{RQ2.} To what extent can NNS' self-reported perceptions serve as reliable proxy of communicative competence?
    \item[] \textbf{RQ3.} How much can AI support close the gap between NNS and NS communicative competence, and what limitations remain?
\end{itemize}

\begin{table*}
\centering
\renewcommand{\arraystretch}{1.2}
\resizebox{\linewidth}{!}{%
    \begin{tabular}{lp{4.4cm}p{3.7cm}p{2cm}p{1.5cm}cc}
    \toprule
    \textbf{Condition} & \textbf{Provided Support} & \textbf{Interaction Patterns} & \textbf{Density} & \textbf{Source} & \textbf{Def.?} & \textbf{Context.?} \\
    \toprule

    \raisebox{-0.2em}{\includegraphics[height=1.1em]{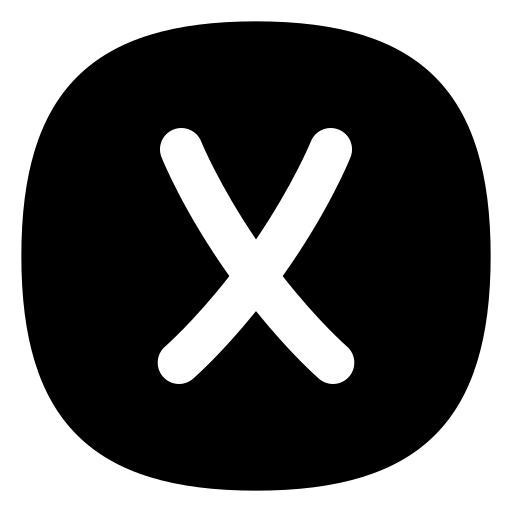}} \textbf{Control}
    & No learning support & - & - & - & - & - \\
    \midrule

    \raisebox{-0.2em}{\includegraphics[height=1.1em]{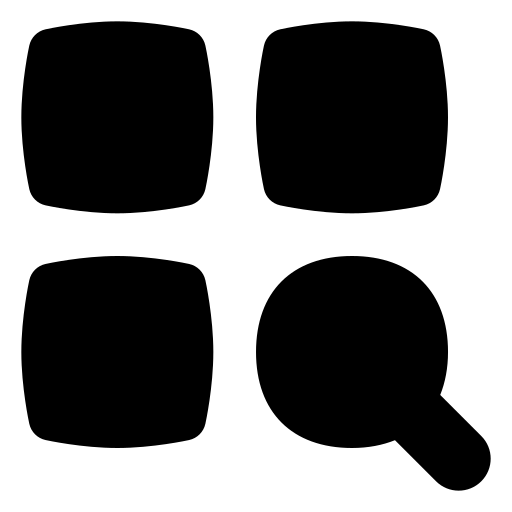}} \textbf{AI Definition}
    & LLM-generated dictionary-style definition of the neologism 
    & AI as traditional dictionaries \citep{lew2024chatgptlexical} 
    & {Low (M=30.5; SD=7.38)} 
    & {AI (Prompt A.1.2)} 
    & \cmark & \xmark \\
    \midrule

    \raisebox{-0.2em}{\includegraphics[height=1.1em]{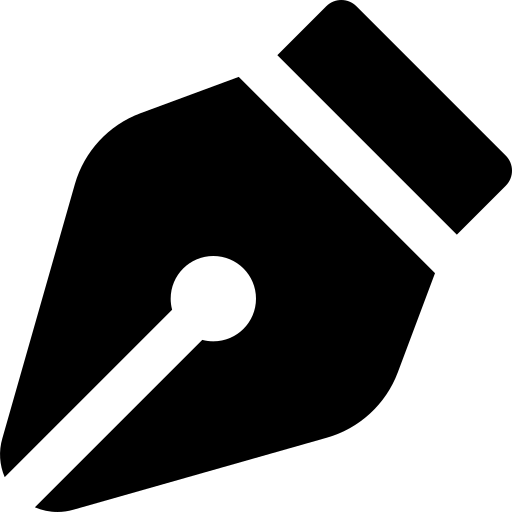}} \textbf{AI Rewrite}
    & LLM-rewritten social media post with the neologism in simpler English 
    & AI-assisted paraphrasing activities \citep{opportunities_2025} 
    & {Low (M=29.5; SD=10.7)} 
    & {AI (Prompt A.1.3)} 
    & \xmark & \xmark \\
    \midrule

    \raisebox{-0.2em}{\includegraphics[height=1.1em]{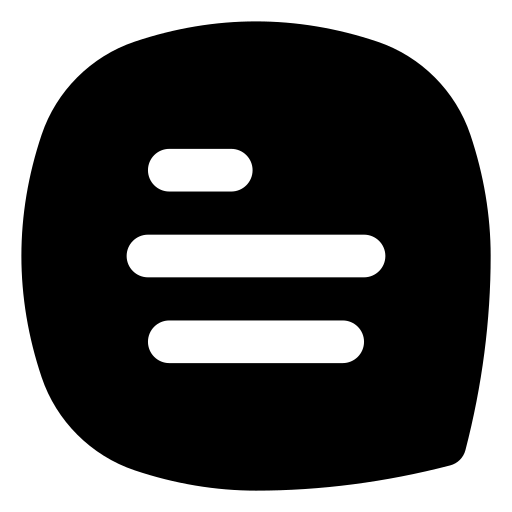}} \textbf{AI Explanation}
    & LLM-generated explanation of the neologism with contextualized usage examples 
    & AI-assisted explainer \citep{kohnke2023chatgpt} 
    & {Mid (M=95.4; SD=5.22)} 
    & {AI (Prompt A.1.4)} 
    & \xmark & \cmark \\
    \midrule

    \raisebox{-0.2em}{\includegraphics[height=1.1em]{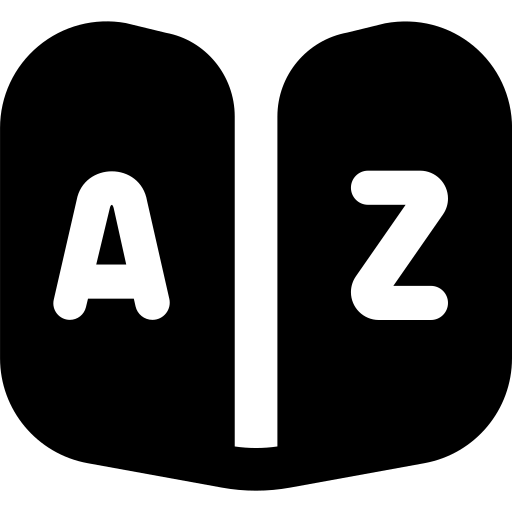}} \textbf{Non-AI Dictionary}
    & Merriam-Webster page for the neologism 
    & Comprehensive official online reference
    & {High (M=412.3; SD=77.4)} 
    & {Human experts} 
    & \cmark & \cmark \\

    \bottomrule
    \end{tabular}
}
\caption{
\textbf{Condition details.}
\textbf{Provided Support} describes the type of learning support given to each condition.
\textbf{Interaction Patterns} characterizes the key interaction attributes of each support type.
\textbf{Density} (information density as low/mid/high) is measured by the mean and standard deviation of word count.
\textbf{Source} indicates where the condition materials are from.
\textbf{Def.?} denotes whether the condition includes a formal definition of the neologism.
\textbf{Context.?} denotes whether the condition includes contextualized usage examples.
}
\label{tab:support_condition_type}
\end{table*}

\section{Methods}
\label{methods}

This section outlines our study design (\S\ref{sec:study_design}), material collection (\S\ref{sec:stimuli_collection}), five types of support conditions (\S\ref{sec:types}), participant recruitment (\S\ref{sec:participants}), and the measures and collected data (\S\ref{sec:dependent_var}).

\subsection{Study Design}
\label{sec:study_design}

We examine how different types of AI support influence NNS' communicative competence with English neologisms using a between-subjects design. NNS participants are situated in a cross-cultural communication scenario where they encounter a neologism in a social media post and wish to communicate with a hypothetical NS friend, \textit{Jo}.\footnote{We use a gender-neutral name to avoid implicit assumptions; \url{https://en.wikipedia.org/wiki/Jo_(given_name)}.}
They first complete a practice session and then rate their familiarity with each neologism on a five-point Likert scale (1:Not at all, 5:Very well) \citep{asahara-2019-word, Zheng2024WordDifficulty}.
For each neologism, they then complete a three-stage procedure to learn and perform tasks assessing their learning goals (\autoref{fig:main_fig}):

 \noindent\textbf{\ding{202} Learning stage.}
NNS participants learn a neologism presented in a social media post under a randomly assigned support condition. 

\noindent\textbf{\ding{203} Production stage.}
Participants write (1) a brief scenario and (2) a message using the learned neologism to their NS friend, \textit{Jo}.
To support understanding, NS-produced writing samples are shown during the practice session, 
and four keywords per neologism, extracted and selected from author-generated writings (Appendix~\ref{appendix:keyword_extraction}), are provided to aid brainstorming without priming participants.
Participants are instructed to make their messages self-contained with at least 10 words, and warnings are displayed for shorter or copy-pasted content.
\looseness=-10

\noindent\textbf{\ding{204} Comprehension stage.}
Participants then rate the contextual appropriateness of the neologism across two provided writing samples by indicating their agreement with ``\textit{The use of the word in this message context is appropriate.}'' on a ten-point Likert scale (1:Not at all, 10:Completely).\footnote{We use a ten-point scale for fine-grained differences, where an one-point improvement is a minimally important difference \citep{Guyatt1987ACO}.} These samples are selected based on 54 NS evaluators' prior ratings, one considered \textit{poor} (M=5.9/10) and the other considered \textit{good} (M=9.6/10).

\ding{202}-\ding{204} process repeats until they finish all eight neologisms. 
Two attention checks are included throughout the study. Detailed study flow and descriptions are provided in Appendices~\ref{appendix:detailed_study} and \ref{appendix:comprehension}.\looseness=-10

\noindent\textbf{Pre-/Post-Task Survey.}
Before beginning the main tasks, NNS participants complete a brief pre-task questionnaire about their demographics and English social media usage.
They later complete a post-task survey asking: (1) perceived confidence, reliance on, and trust in the support for future use \citep{hoffman2019metricsexplainableaichallenges}; (2) perceived mental burden and task difficulty; and (3) open-ended feedback on the support for the \ding{203} \textbf{Production} task.

\subsection{Materials}
\label{sec:stimuli_collection}

\noindent\textbf{Neologisms.}
We compile a pool of 30 neologisms from the Merriam-Webster slang and trending words list,\footnote{\url{https://www.merriam-webster.com/slang}} and filter them by popularity using Google Trends, retaining those that emerged in 2020 or later. This yields 24 candidate neologisms.

\noindent\textbf{Social Media Posts.}
For each neologism, we select one representative social media post from Merriam-Webster to convey the scenario. 
Social media posts provide a realistic, engaging context where neologisms naturally appear and offers an effective, focused learning setting for NNS participants. We present the full list of posts in Appendix \autoref{tab:support_conditions}.

\subsection{Support Condition Types}
\label{sec:types}

We design four support types, in total five conditions as outlined in~\autoref{tab:support_condition_type}.
\textbf{\raisebox{-0.2em}{\includegraphics[height=1.1em]{figure/logo/control.png}} Control} reflects the standalone usefulness of a social media post without any learning support for neologisms.
The three AI-based support conditions differ in the \textit{type} of information provided, corresponding to how people interact with AI in a single turn, and are generated using GPT-4.1.\footnote{\texttt{gpt-4.1-2025-04-14}; prompts in Appendix~\ref{appendix:prompts}.}
\textbf{\raisebox{-0.2em}{\includegraphics[height=1.1em]{figure/logo/definition.png}} AI Definition} illustrates basic cases where NNS use AI as an alternative to dictionaries.
\textbf{\raisebox{-0.2em}{\includegraphics[height=1.1em]{figure/logo/rewrite.png}} AI Rewrite} covers scenarios where users understand neologisms in the context of a specific example.
\textbf{\raisebox{-0.2em}{\includegraphics[height=1.1em]{figure/logo/explanation.png}} AI Explanation} provides explanations in 3--5 sentences of the neologism's usage, including typical contexts, tone, audience, and connotations.
\textbf{\raisebox{-0.2em}{\includegraphics[height=1.1em]{figure/logo/dictionary.png}} Non-AI Dictionary} shows the full content from the Merriam-Webster website, which includes the definition, example usages, word origin, and typical contexts.\footnote{For example, \url{https://www.merriam-webster.com/slang/main-character-energy}; to isolate the text content from distracting layout elements, we extract the text and render it within a controlled, uniform template.}
We use the official dictionary condition, which provides the most comprehensive and accurate information, to compare 
the utility of \textit{imperfect} AI-based support.

From an initial pool of 24 candidate neologisms, we select a final set of eight by ensuring comparable error rates across all three AI-based conditions, measured through both automatic and human annotation against corresponding Merriam-Webster content (Appendix~\ref{appendix:error_annotation}). The final neologisms are: \textit{brain rot}, \textit{canon event}, \textit{cheugy}, \textit{copium}, \textit{crash out}, \textit{delulu}, \textit{grindset}, and \textit{reheat nachos}.\footnote{Our neologisms span diverse linguistic taxonomy \citep{pinter-etal-2020-nytwit, zheng-etal-2024-neo}; more details in Appendix~\ref{appendix:support_conditions}.}

\subsection{Participants}
\label{sec:participants}

We recruit all participants from Prolific and the sample size is determined using power analysis (expected effect size Cohen's $f$=0.25; Appendix~\ref{appendix:power_analysis}).

\paragraph{\raisebox{-0.2em}{\includegraphics[height=1.1em]{figure/logo/nns.png}} NNS Participants.}
We recruit 234 participants from around the world who self-identified Spanish, German, or Chinese as their first and primary language and English as fluent language (104, 95, and 35 participants, respectively). We select languages to represent diverse families: Spanish as Indo-European, German as Germanic, and Chinese as Sino-Tibetan.
We recruit participants aged 18--44, likely to engage with our communicative contexts (i.e., social media posts), and randomly assign them to five conditions. Each receives USD 6 (equivalent to USD 12/hour); 234 (96.3\%) participants who pass both attention checks earn a USD 2 bonus and are included in the analysis.

Of the 234 NNS participants, the mean age was 28.7 years (SD=6.63). 
Gender was: man (57.3\%), woman (38.5\%), non-binary (3.0\%), and prefer not to say (0.9\%).
Monthly English social media use was: never (2), rarely (9), sometimes (24, 10\%), often (38, 16\%), and almost daily (160, 68.7\%).
Average completion time was 49 minutes (SD=26).

\paragraph{\raisebox{-0.2em}{\includegraphics[height=1.1em]{figure/logo/ns.png}} NS Evaluators.}
We recruit 144 native English speakers to evaluate NNS-produced writing samples during the \ding{203} \textbf{Production} stage, simulating the role of the other party in a realistic cross-cultural communication scenario. 
Evaluators are based in the United States and self-identify English as their first, primary, and fluent language. 
Each rates 26 NNS-produced writings for one assigned neologism (randomized across evaluators), and is provided the Merriam-Webster dictionary for reference, with additional tools allowed. 
Evaluators receive USD 6 (equivalent to USD 12/hour), with an average completion time of 32 minutes (SD=9).

Both the NNS and NS tasks were approved by our institution's IRB, and all participants provided consent prior to the study. Details on the NS evaluation survey and participant demographics are provided in Appendices~\ref{appendix:detailed_ns_survey} and \ref{appendix:participants}.

\begin{figure*}[t]
    \centering
    \includegraphics[width=\linewidth]{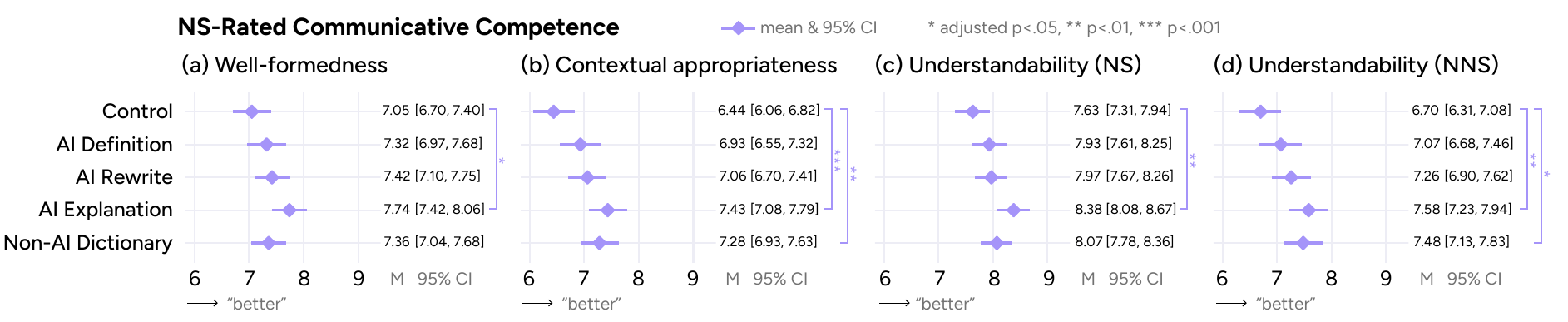}
    \caption{\textbf{RQ1+3: NS-rated communicative competence.} 
    All metrics are measured on a ten-point Likert scale (1:Not at all, 10:Completely).
    $x$-axis is truncated as no ratings fall below 6 or above 9.
    (c) Understandability (NS): how much the NS evaluator understands the message;
    (d) Understandability (NNS): how much the NS evaluator thinks the NNS participant understood the meaning of the neologism through the message.
    AI Explanation significantly improves communicative competence over Control across all dimensions.
    }
    \label{fig:ns-results}
\end{figure*}

\subsection{Measures and Collected Data}
\label{sec:dependent_var}

\paragraph{\textcolor{ns-eval}{\ding{117}} NS-Rated Competence (RQ1, RQ3).}
During the \ding{205} \textbf{NS Evaluation} stage, NS evaluators rate each NNS-produced writing sample using five questions covering three dimensions of communicative competence (\S\ref{subsec:communicative_competence}): \textbf{Well-formedness}, assessing (1) grammatical acceptability and (2) coherence/cohesiveness regardless of meaning, derived from grammatical and discourse competence; (3) \textbf{Contextual appropriateness}, evaluating how appropriately the neologism is used in context, capturing sociolinguistic competence; and \textbf{Understandability}, measuring (4) ease of understanding the message and (5) whether the NNS correctly understood the neologism's meaning, reflecting strategic competence.
In total, this yields 3,744 observations (1,872 writings from 234 participants $\times$ 8 neologisms, each receiving two ratings per question).

We also collect writing samples from each NS evaluator during the \ding{205} \textbf{NS Evaluation} stage, which are subsequently rated by 16 additional NS evaluators on the same dimensions, yielding 288 observations (144 writings $\times$ two ratings each).

\paragraph{\textcolor{nns-mcq}{\ding{115}} NNS Comprehension Competence (RQ1).}
For each writing sample $w$ shown to NNS participants during the \ding{204} \textbf{Comprehension} stage, let $\hat{c}$ be the collected contextual appropriateness rating. We take the difference from the mean of the two NS evaluator ratings, $c^*_1$ and $c^*_2$, collected prior to the main study:
\begin{equation}
    \mathbf{Distance}(w) = \left| \hat{c}-{\frac{(c^*_1+c^*_2)}{2}} \right| \nonumber
\end{equation}
where a lower distance indicates that the NNS participant's rating is closer to the NS evaluators'. Each NNS participant rates two writing samples per neologism, yielding a total of 3,744 observations.

\paragraph{\textcolor{nns-self}{\ding{108}} NNS Self-Reported Perceptions (RQ2).}
We collect NNS participants' self-reported \textit{confidence} in and \textit{helpfulness} of the support during the \ding{202} 
\textbf{Learning} (for correctly understanding the neologism, if any) and \ding{203} \textbf{Production} stages (for correctly writing the message, if any), each on a five-point Likert scale (1:Not confident/helpful at all, 5:Very confident/helpful).
Helpfulness and confidence ratings are not collected for the Control group, resulting in a total of 1,872 observations.

\section{Results}

We assess the effectiveness of each support condition (\S\ref{res:1}), measure NNS participants' self-reported perceptions (\S\ref{res:2}), and compare NS ratings of NNS- to NS-produced writing (\S\ref{res:3}).

\subsection{RQ1: How Effective is Each Support?}
\label{res:1}

We employ linear mixed-effects models, where the dependent variables are NS-rated competence and NNS comprehension competence.
Fixed effects include support condition, language group, their interaction, English social media usage, and initial familiarity with the neologisms (collected prior to the main task).
Random intercepts include NNS participant ID (all models), NS evaluator ID (for NS ratings), and neologism ID (for other outcomes).
We report the average marginal effects with Bonferroni-adjusted \textit{p}-values to address multiple comparison issues.\footnote{We use R: \texttt{lme4} \citep{lme4}, \texttt{lmerTest} \citep{lmerTest}, and \texttt{emmeans} \citep{emmeans}.} To address the limitations of \textit{p}-values, we also report 95\% confidence intervals (CIs)~\citep{dragicevic2016fair, intro_to_new_stats}. All means, \textit{p}-values, and CIs are reported in figures. 

\paragraph{\textcolor{ns-eval}{\ding{117}} \textbf{NS-Rated Competence.}}
We show results in~\autoref{fig:ns-results} and summarize key findings below:
\begin{itemize}[leftmargin=*, itemsep=2pt, parsep=-1pt, topsep=0pt, label={}]
    \item \textbf{(a) Well-formedness.}\footnote{We average (1) grammatical acceptability and (2) coherence/cohesiveness ratings due to high correlation.} AI Explanation significantly improves over Control, while no significant differences are observed for other conditions.
    \item \textbf{(b) Contextual appropriateness}. Both AI Explanation and Non-AI Dictionary significantly improves over Control. AI Definition and AI Rewrite fall in between these support conditions, with no statistically significant differences.
    \item \textbf{(c) Understandability (NS).} AI Explanation significantly helps NNS participants write messages better understood by NS evaluators compared to Control. No significant differences are observed for other treatment support conditions.
    \item \textbf{(d) Understandability (NNS).} As with Contextual appropriateness, both AI Explanation and Non-AI Dictionary significantly outperform Control, while others show no significant differences.
\end{itemize}
Taken together, compared to no support (Control), AI Explanation consistently helps NNS participants achieve higher communicative competence in all dimensions.
In contrast, while Non-AI Dictionary offers the most comprehensive and accurate information, it improves only a subset of dimensions. This is likely due to its high information density (\autoref{tab:support_condition_type}), which is perceived as higher mental burden and task difficulty (\autoref{tab:post_survey}).

\begin{figure}
    \centering
    \includegraphics[width=\linewidth]{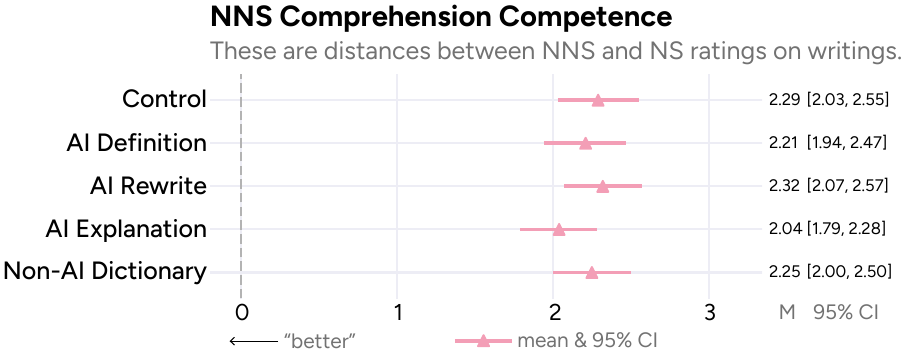}
    \caption{\textbf{RQ1: NNS comprehension distance.} Compared to the Control, none of the treatment support conditions significantly reduce the distance.
    }
    \label{fig:mcq-results}
\end{figure}


\paragraph{\textcolor{nns-mcq}{\ding{115}} \textbf{NNS Comprehension Competence.}}
As shown in~\autoref{fig:mcq-results}, AI Explanation yields the smallest average distance, but the difference is not statistically significant compared to other support conditions. 
The overall average distance remains relatively high (2.22), suggesting that while some support conditions (AI Explanation and Non-AI Dictionary) yield significantly higher NS-rated competence, they do not necessarily help NNS participants to accurately judge the contextual appropriateness of neologisms in writing samples.

\subsection{RQ2: Perceived vs. Actual Competence}
\label{res:2}

\begin{figure}[t!]
    \centering
    \includegraphics[width=\linewidth]{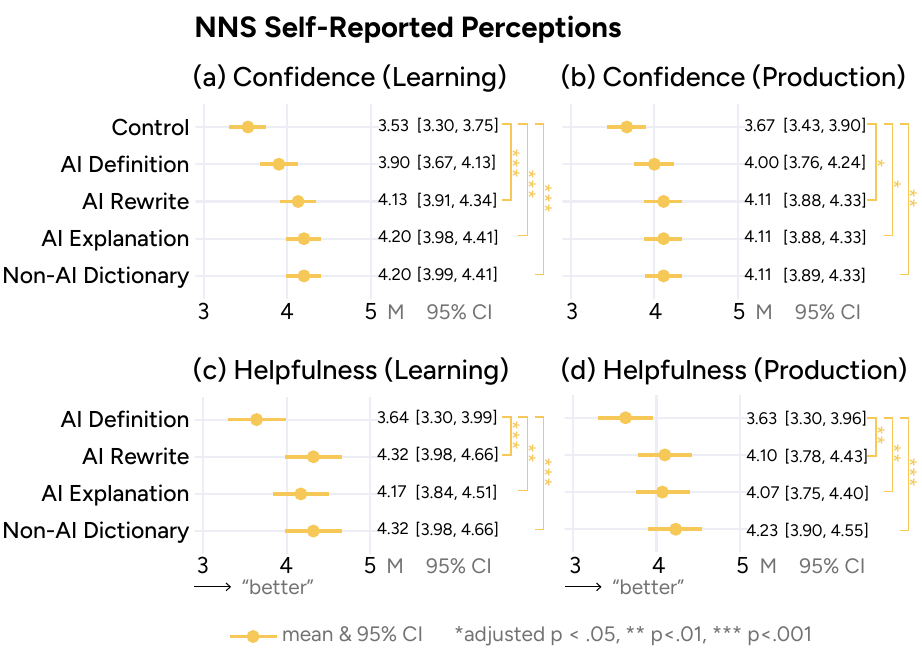}
    \caption{\textbf{RQ2: NNS self-reported confidence and helpfulness.} 
    All support conditions except AI Definition significantly improve confidence over Control; all conditions improve helpfulness over AI Definition.
    }
    \label{fig:nns-self-results}
\end{figure}

We show \textcolor{nns-self}{\ding{108}} \textbf{NNS Self-Reported Perceptions} in \autoref{fig:nns-self-results} and summarize findings below:

\begin{itemize}[leftmargin=*, itemsep=2pt, parsep=-1pt, topsep=0pt, label={}]
    \item \textbf{(a+b) Confidence (Learning+Production).} Non-AI Dictionary, AI Explanation, and AI Rewrite significantly improve confidence over Control, in decreasing order of effect size, while AI Definition shows no significant difference.
    \item \textbf{(c+d) Helpfulness (Learning+Production).} Non-AI Dictionary, AI Rewrite, and AI Explanation all significantly improve over AI Definition, in decreasing order of effect size.
\end{itemize}
Overall, NNS participants' self-reports serve as limited proxy for NS-rated communicative competence.\footnote{We further provide AI-rated communicative competence results using LLM-as-judge in Appendix~\ref{appendix:llm_as_judge}, showing that, in aggregate, they distinguish writings under five conditions while producing higher ratings than NS-rated competence.} While AI Explanation and Non-AI Dictionary align with NS ratings (\S\ref{res:1}), AI Rewrite also boosts confidence and perceived helpfulness without reliably improving actual competence. 
Combined with open-ended comments (Appendix~\ref{appendix:detailed_comments}) and post-task survey results (\autoref{tab:post_survey}), we attribute this over-reliance to its ease of understanding: it presents social media posts in simpler terms, reducing perceived mental burden and task difficulty.\footnote{The effects of the conditions are consistent across all three languages. Results for each measure by language group (Spanish, German, and Chinese) are reported in Appendix~\ref{appendix:lang_group_results}.}

\begin{table}
\centering
\resizebox{\linewidth}{!}{%
    \begin{tabular}{lllllll}
    \toprule
    \textbf{Measures} & \textbf{\raisebox{-0.2em}{\includegraphics[height=1.1em]{figure/logo/control.png}}} & \textbf{\raisebox{-0.2em}{\includegraphics[height=1.1em]{figure/logo/definition.png}}} & \textbf{\raisebox{-0.2em}{\includegraphics[height=1.1em]{figure/logo/rewrite.png}}} & \textbf{\raisebox{-0.2em}{\includegraphics[height=1.1em]{figure/logo/explanation.png}}} & \textbf{\raisebox{-0.2em}{\includegraphics[height=1.1em]{figure/logo/dictionary.png}}} \\
    \toprule

    \textbf{Confidence (↑)} & - & 3.42 & 4.02 & \underline{4.04} & \textbf{4.51} \\
    \textbf{Reliance (↑)} & - & 2.74 & \underline{3.60} & 3.40 & \textbf{4.28} \\
    \textbf{Trust for Future Use (↑)} & - & 3.56 & 3.67 & \underline{3.88} & \textbf{4.44} \\
    
    \midrule
    \textbf{Mental Burden (↓)} & 3.67 & \underline{3.42} & \textbf{3.31} & 3.44 & 3.77 \\
    \textbf{Task Difficulty (↓)} & 3.70 & 3.66 & \underline{3.51} & \textbf{3.46} & 3.65 \\
    
    \bottomrule
    \end{tabular}
}
\caption{\textbf{Post-survey statistics of NNS participants.} \textbf{Best} and \underline{second best} scores for each metric are highlighted. All metrics are measured on a 1--5 scale. \raisebox{-0.2em}{\includegraphics[height=1.1em]{figure/logo/dictionary.png}} Non-AI Dictionary yields the highest confidence, reliance, and trust for future use.
Detailed results per language group are provided in Appendix~\ref{appendix:detailed_post}.}
\label{tab:post_survey}
\end{table}

\definecolor{midgreen}{RGB}{96, 171, 117}

\begin{table*}[]
\centering
\resizebox{\linewidth}{!}{
    \begin{tabular}{ 
        m{6.5cm}
        m{8.4cm}
        >{\centering\arraybackslash}m{2.2cm}
        >{\centering\arraybackslash}m{0.8cm}
        >{\centering\arraybackslash}m{0.8cm}
        >{\centering\arraybackslash}m{0.8cm}
    }
    \toprule
    \textbf{Scenario} & \raisebox{-0.2em}{\includegraphics[height=1.1em]{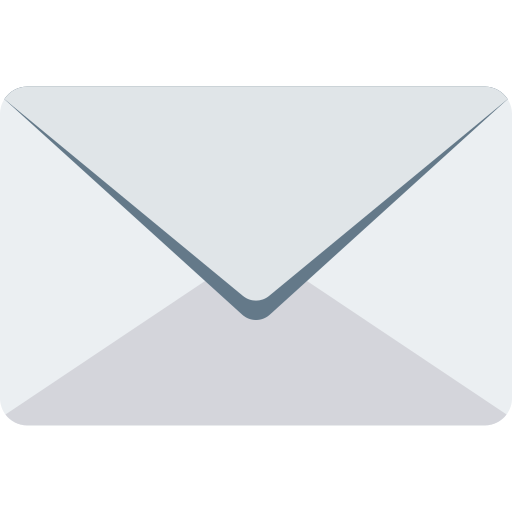}} \textbf{Message to \textit{Jo}} & \textbf{Support} & \textbf{WF} & \textbf{CA} & \textbf{U} \\
    \toprule

    {Telling Jo about a botched pizza delivery on a special occasion}	& The pizza arrived, finally! They took too long so I got it free, which rocks honestly. But I think they pulled a \textbf{reheated nachos} on me, this thing looks like if it has been put through the oven twice! It's texture is crunchy and dried up. Like it's still free but... I expected something better, today it's a big day after all. & \raisebox{-0.2em}{\includegraphics[height=1.1em]{figure/logo/definition.png}} & \cellcolor{red!20} 1 & \cellcolor{red!20} 1.5 & \cellcolor{red!20} 1 \\
    \midrule
    I'm telling Jo about Tim and his essay	& Hey Jo, Tim don't know how to write an essay so he's \textbf{reheating} her roommate's \textbf{nachos}... & \raisebox{-0.2em}{\includegraphics[height=1.1em]{figure/logo/rewrite.png}} & \cellcolor{orange!20} 5 & \cellcolor{orange!20} 6 & \cellcolor{orange!20} 5 \\
    \midrule
    Talking about the entertainment industry	& I feel like it's getting more difficult to create something new - whether that be in music, movies, etc. Everything just sounds and looks like \textbf{reheated nachos} nowadays. & \raisebox{-0.2em}{\includegraphics[height=1.1em]{figure/logo/dictionary.png}} & \cellcolor{midgreen!30} 10 & \cellcolor{midgreen!30} 10 & \cellcolor{midgreen!30} 10 \\
    \midrule
    Someone made something that was not as good as the original version.	& Taylor Swift was totally \textbf{reheating} Sabrina Carpenter's \textbf{nachos} on this new album. & \textsc{NS} & \cellcolor{midgreen!30} 10 & \cellcolor{midgreen!30} 10 & \cellcolor{midgreen!30} 10 \\
    
    \bottomrule
    \end{tabular}
}
\caption{\textbf{NNS- and NS-produced writing samples with NS-rated competence for \underline{reheat nachos}.} \textsc{\textbf{WF}}: Well-formedness; \textsc{\textbf{CA}}: Contextual appropriateness;\textsc{ \textbf{U}}: Understandability (averaged). We color-code the ratings for \raisebox{3pt}{\colorbox{red!20}{ }} (low), \raisebox{3pt}{\colorbox{orange!20}{ }} (mid), and \raisebox{3pt}{\colorbox{midgreen!30}{ }} (high). 
NS denotes NS-produced writing.
Examples for other words are in Appendix~\ref{appendix:detailed_examples}.} 
\label{tab:examples}
\end{table*}

\subsection{RQ3: NNS vs. NS Communicative Competence}
\label{res:3}

We compare NS-rated competence between NNS- and NS-produced writing samples. To assess differences across groups, we perform a Kruskal-Wallis test for ordinal ratings and non-parametric data. 
NS-produced writings receive significantly higher well-formedness (M=7.94, 95\% CI=[7.70, 8.19]) and contextual appropriateness (M=7.67 [7.36, 7.98]) than all NNS support conditions (\textit{p}<.001), except for AI Explanation (n.s.).
NS writing also score higher averaged understandability (M=8.06 [7.81, 8.31]) than all support conditions: Control and AI Rewrite (\textit{p}<.001), AI Definition (\textit{p}=.001), and Non-AI Dictionary (\textit{p}=.011).
Descriptive statistics of NNS- and NS-produced messages are provided in Appendix~\ref{appendix:nns_writings}.

Together, these results show a significant gap between NNS- and NS-produced messages in communicative competence and style, suggesting that no single support alone can fully bridge it.

\section{Discussion}

\subsection{Understanding the NNS-NS Gap}
\label{discussion:1}

To understand the sources of the gap observed in communicative competence between NNS- and NS-produced writing samples (\S\ref{res:3}), we analyze open-ended comments from NS evaluators on the NNS-produced writings.
Many evaluators note that NNS participants appeared to ``\textit{have little understanding of what the neologism actually means}'' as messages often ``\textit{feel awkward},'' contain ``\textit{replaced words}'' (e.g., \underline{copium} $\rightarrow$ cope, \underline{crash out} $\rightarrow$ crashed), or ``\textit{use in secondary meanings}.''

More importantly, evaluators note cases where participants ``\textit{used neologisms literally},'' specifically for the word \underline{reheat nachos}. 
For this word, AI Definition provides only the literal meaning: \textit{reheating prepared nachos to a desirable temperature or texture} (Appendix~\ref{appendix:support_conditions}). 
As such, AI outputs are often imperfect, yet NNS lack the mechanisms to reliably judge their correctness, leading to inaccurate use of neologisms during communication (first example in \autoref{tab:examples}). Feedback from NNS participants confirm this: many noted the support was ``\textit{especially helpful when it's a word that I sort of know, but for an entirely new term, it adds confusion and I cannot judge whether it is correct}.'' This further explains the limitations of using NNS self-reports as a reliable proxy for communicative competence (\S\ref{res:2}).\footnote{Detailed comments are provided in Appendix~\ref{appendix:detailed_comments}.}
Quantitatively, this gap appears in §\ref{res:1}: NNS participants write effective messages yet struggle to judge appropriate neologism use.

\subsection{Implications for Future AI Design}
\label{discussion:2}

How can we bridge this competence gap? 
As NS evaluators note, ``\textit{unless you find yourself or friends using the term, it can be hard to sometimes know if it's being used in the proper way},'' reflecting the rich sociocultural references neologisms carry.
Yet, direct involvement of NS is often limited by practical constraints: NNS have few opportunities for individualized practice \citep{adaptive_second_2024}, social discomfort when speaking awkwardly \citep{Nederhof1985MethodsOC, kim1994cross}, or prefer AI-based learning tools \cite{ngo2024chatgpt}.
Our results show that current AI tools, when used in certain ways, can help improve NNS communicative competence (\S\ref{res:1}).

\paragraph{Richer Contextual Support.}
Feedback from NNS participants in the AI Explanation group highlights the value of ``\textit{various notes on context usage examples, origin, tone, and audience}.''
Together with this group's high NS-rated competence (\S\ref{res:1}), the results suggest that current AI tools can generate informative usage examples that help NNS communicate neologisms effectively.\footnote{See Appendix~\ref{appendix:ai_writing} for NS-rated competence ratings for AI-generated, NNS-, and NS-produced writing samples.}
At the same time, requests for ``\textit{more examples},'' especially for those ``\textit{based on real-world scenarios showing how NS actually uses the word},'' including ``\textit{direct quotes},'' ``\textit{common scenarios},'' or ``\textit{typical mistakes made by NNS}'' (i.e., unnatural usage), were common across all conditions. 
To meet this need, future AI-based support should provide richer, contextualized examples grounded in sociocultural nuance and authentic usage, through: (i) contrastive minimal pairs of natural NS usage and plausible NNS misuses, mined from real-world communities where neologisms appear in situ (e.g., r/Neologisms) or learners seek feedback (e.g., r/EnglishLearning); (ii) structured pragmatic guidance tied to audience, tone, and connotation (e.g., ``Use it when... / Avoid it when...''), making implicit pragmatic cues explicit as AI Explanation already begins to do; and (iii) iterative practice loops in which NNS draft messages and receive targeted feedback, addressing the limitations of single-turn support.
The information density of the support and mental load for the user need to be considered together.


\paragraph{Up-to-date Support for Evolving Language.}

Beyond support design, the time-sensitive nature of neologisms makes continuously updating AI training data uniquely challenging, often leading to outdated knowledge (\S\ref{discussion:1}).
Future tools should address this on two fronts.
First, they should detect emerging neologisms through statistical signals (e.g., atypical likelihood or perplexity patterns) and leverage retrieval-based mechanisms to ground outputs in up-to-date usage examples and definitions.
Second, and equally important, they should explicitly communicate uncertainty \citep{kadavath2022language}, both about neologism meanings and about feedback quality, via confidence scores \citep{lin2022teaching} or verbal hedging \citep{prokofieva2014hedging} (e.g., ``\textit{I'm not sure, but reheat nachos may mean...}''), helping NNS learners calibrate their reliance on AI outputs.
This is especially critical given that our preliminary LLM-as-judge analysis (Appendix~\ref{appendix:llm_as_judge}) suggests LLM judges tend to overestimate NNS communicative competence relative to NS ratings, highlighting the risk of overly affirmative feedback that may reinforce learner overconfidence.

\subsection{Implications for Our Society}
\label{discussion:3}
While there is clearly room to bridge cross-cultural communication gaps, we caution the community to consider the nature of neologisms, or new slang as studied here, when designing AI tools~\cite{DRAKE198063}.
Informal communication plays an important role in how people build social connections, and the language used in those conversations signifies the social identities of the speakers, further influencing how they identify ingroup or outgroup members to coordinate or build social networks~\cite{10.1145/2441776.2441880, Mendelsohn_Ghosh_Jurgens_Budak_2023, mendelsohn-etal-2023-dogwhistles}.
Similar to intergenerational gaps in slang~\cite{43095617, intergenerational-discourse}, newly emerging slang brings challenges in communication, but when the technology becomes good enough so a slang term is universally known across cultures and languages, it loses its function of building unique, shared connections between people or within a community. 
Therefore, carefully evaluating the societal impacts and the degree to which people want AI to help, or not, is necessary when imagining future AI.


\section{Conclusion}

We explore the utility of AI support for helping NNS use English neologisms in informal interactions with NS. 
In a between-subjects study, NNS participants complete production and comprehension tasks, with their writings rated for communicative competence by NS evaluators. 
Across five support conditions, AI Explanation consistently yields the highest NS-rated competence (\S\ref{res:1}). 
NNS self-reported perceptions are assessed as proxies for competence, but their overestimation limits reliability (\S\ref{res:2}).
We further observe competence gaps between NNS- and NS-produced writings (\S\ref{res:3}), largely driven by NNS' difficulty judging imperfect AI outputs and limited contextualized support (\S\ref{discussion:1}). 
Our findings inform design for future AI support and tools (\S\ref{discussion:2}; \S\ref{discussion:3}) to help NNS communicate neologisms more naturally with NS, rather than ``\textit{like my grandma trying to use slang}.''\footnote{Quote adapted from one of our NS evaluators.}

\section*{Limitations}

The scope of our human study is limited to the current experimental setup. We specifically focused on \textit{informal} cross-cultural communication of neologisms between non-native and native English speakers, thus our findings may not generalize to other cross-cultural communication contexts. Additionally, our AI-based support types are limited to English-based interventions (e.g., we did not provide AI Explanation in participants' native language), involved only single-turn interactions, and did not encompass the full range of ways people engage with AI tools (e.g., search engine results), which are valuable directions for future research.

\section*{Ethics Statement}

This study was approved by our Institutional Review Board (IRB). All participants provided informed consent prior to participation and were compensated according to the rates specified in the consent form. Further, NNS participants who passed both attention checks received a bonus.
\section*{Acknowledgments}



We would first like to thank the members of the \textsc{clip} lab at the University of Maryland for their constructive feedback and support.
We especially thank Connor Baumler, Nishant Balepur, and Atrey Desai for revising the AI support condition content, and Lorena Calvo-Bartolomé for piloting our annotation as an NNS participant. 
We also appreciate Yixin Bai, Taehyun Yang, Caleb Holland, Daniel Palamarchuk, Lingjun Zhao, and Navita Goyal for providing early feedback on our project scope and annotation surveys.
We would further like to thank Calvin Bao, Kartik Ravisankar, Osvaldo Quinjica, HyoJung Han, and Eric Bennett for their feedback on the draft.
Last but not least, Dayeon and Yu would like to extend their gratitude to friends who helped test early stages of the pilot studies: Chenxi Cui, Nuan Wen, Chenghao Yang, Yilin Zhang, Yue Feng, Chen Guo, Yu Miao, Paiheng Xu, Yu Duan, Enwei Wu, and Wei Sun for the Chinese version, and Christopher Sunghun Choi, Nakyung Lee, Daeun Jung, Curie Kim, Joohyung Song, Jeongin Kim, Jisoo Lee, Hano Lee, Hanseul Nam, Dakyoung Heo, Gayoung Lee, Nayeon Kim, Hyojin Lim, and Namhee Kim for the Korean version.
We also thank all the Prolific participants who took part in our study.
This material is based upon work partially supported by the NSF under Grant No. 2229885 (NSF Institute for Trustworthy AI in Law and Society, TRAILS), and NSF CAREER Award No. 2339746 (Rudinger).
Any opinions, findings and conclusions or recommendations expressed in this material are those of the author(s) and do not necessarily reflect the views of the National Science Foundation.

\bibliography{custom}

\newpage
\appendix

\section{Prompts}
\label{appendix:prompts}

\subsection{Support Conditions}
\label{appendix:prompts_1}
We show prompts used for generating AI-based support conditions with GPT-4.1 below. \texttt{\{word\}} entry is populated with each neologism.

\begin{prompt}[title={Prompt A.1.1: System Prompt}]
You are a multilingual language expert, who can understand neologisms very well. Neologism is any newly formed word, term, or phrase that has achieved popular or institutional recognition and is becoming accepted into mainstream language. We are particularly interested in internet slangs, which are non-standard or unofficial forms of language used by people on the Internet (such as social media, forums, or messaging apps) to communicate with one another.
\end{prompt}
\begin{prompt}[title={Prompt A.1.2: AI Definition}]
\texttt{\{system prompt\}} \\ \\
Given the word, provide a formal dictionary definition in English. Return only the definition, no other text. \\ \\
\textbf{Word:} \texttt{\{word\}} \\
\textbf{Definition:} 
\end{prompt}

\begin{prompt}[title={Prompt A.1.3: AI Rewrite}]
\texttt{\{system prompt\}} \\ \\
Given the social media post with the word \texttt{\{word\}}, rewrite it into plain English, which is suitable for a general audience. You must rewrite the \texttt{\{word\}} into English. Return only the rewrite, no other text. \\ \\
\textbf{Post:} \texttt{\{social media post\}} \\
\textbf{Rewrite:} 
\end{prompt}

\begin{prompt}[title={Prompt A.1.4: AI Explanation}]
\texttt{\{system prompt\}} \\ \\
Given the word \texttt{\{word\}}, explain in English how it is used, including typical situations, tone, intended audience, and connotations. Return only the explanation in 3–5 sentences, no other text. \\ \\
\textbf{Word:} \texttt{\{word\}} \\
\textbf{Explanation:} 
\end{prompt}

\subsection{AI-Rated Communicative Competence}
\label{appendix:prompts_2}

\begin{prompt}[title={Prompt A.2: AI-Rated Communicative Competence}]
\textbf{Task Instruction:} Imagine your friend is telling you about something that happened in their day, and they use a specific neologism to describe it. \\
Neologisms are newly created terms or phrases that have gained popularity and are starting to be accepted into everyday language. \\
As a native English speaker familiar with neologisms, we’d love your help in evaluating these scenarios. \\
You will be evaluating scenarios for one neologism {word}, so please make sure to understand the neologism using our provided reference dictionary before proceeding. \\

For each scenario, your friend will share: \\
1. What happened (a short description) \\
2. A Message they sent to you using the word. \\
If you’re unsure about the meaning of the neologism, please feel free to check the reference dictionary page provided in the beginning or use any resources you trust! \\

\textbf{Neologism:} \texttt{\{word\}} \\
\textbf{Reference dictionary page:} \texttt{\{dictionary URL\}} \\ \\
\textbf{What happened:} \texttt{\{scenario\}} \\
\textbf{Message from your friend:} \texttt{\{message\}} \\ \\
Please indicate how much you agree with each statement from 1:Not at all to 10: Completely. \\
\textbf{Well-formedness:} Grammatical correctness, coherence, cohesiveness \\
\textbf{Q1.} The grammar of the message is acceptable. \\
\textbf{Q2.} The message is coherent and cohesive in English, regardless of its meaning. \\
\textbf{Contextual appropriateness:} Appropriateness of the word's usage in context \\
\textbf{Q3.} The use of the word ``\texttt{\{word\}}'' in this message context is appropriate. \\
\textbf{Understandability:} Ease of understanding \\
\textbf{Q4.} I understand what my friend is trying to say in this message context. \\
\textbf{Q5.} I think my friend understood the meaning of the word ``\texttt{\{word\}}'' from this message. \\

Also provide your rating for confidence from 1-5: \\
\textbf{Q6.} How confident do you feel in the judgments you provided? \\
5: Very confident \\
4: Confident \\
3: Somewhat Confident \\
2: Not confident \\
1: Not confident at all \\

Return the response in JSON format with keys ``Q1'' through ``Q6'' and values for your judgment only. Do not add any extra text.
\end{prompt}

\subsection{AI-Generated Writing Sample}
\label{appendix:prompts_3}
We use the same system prompt (A.1.1).

\begin{prompt}[title={Prompt A.3: AI-Generated Writing Sample}]
\texttt{\{system prompt\}} \\ \\
Imagine something just happened, and you want to tell your native English-speaking friend Jo about it. Please (1) briefly describe what happened in the What section, and (2) write a message with the word \texttt{\{word\}} you would send to Jo. \\ \\
To get started with the scenario, you can think about: \\
- something that happened to you (or a common acquaintance), when and where? \\
- something you just saw in the news \\
- ... anything you'd love to share with Jo \\ \\
Try to make your message self-contained so that Jo can understand it easily, and write at least 10 words. \\
Please generate 5 scenarios and return the response in JSON format with keys ``what'' and ``message'' for each scenario. \\ \\
\textbf{Output Example:}
\begin{lstlisting}
[    
    {{
        "idx": "1",
        "what": "...",
        "message": "..."
    }},
    ...
]
\end{lstlisting}
\end{prompt}

\section{Study Design Details}
\label{appendix:study_design}

We present details on our human study design.

\subsection{Keyword Extraction Process}
\label{appendix:keyword_extraction}

During the \ding{203} \textbf{Production} stage, participants are instructed to write a brief scenario and a message to their NS friend. Across three pilot studies, NNS participants consistently reported that the writing was the primary bottleneck, leading to completion times that were longer than initially expected.

To support the writing process, we generate five example writing samples per neologism and extract four keywords to present to participants. Keywords are selected according to the following criteria to avoid priming participants:
\begin{itemize}[leftmargin=*, itemsep=2pt, parsep=-1pt]
    \item Avoid revealing or at least implicitly conveying the meaning of the neologism (e.g., the keyword ``phone'' for ``\underline{brain rot}'').
    \item Span diverse everyday contexts, including social, work, and leisure domains.
    \item Support multiple plausible scenarios to avoid constraining to a single predictable narrative.
    \item Exclude sensitive or controversial topics (e.g., politics, gender).
\end{itemize}
NNS participants are not required to use the keywords but are encouraged to draw on them if they run out of ideas (\autoref{fig:study_flow}). The complete set of keywords for each neologism is shown in \autoref{tab:keywords}.

\begin{table}
\centering
\resizebox{\linewidth}{!}{%
    \begin{tabular}{llllllll}
    \toprule
    \textbf{Neologism} & \textbf{Keywords} \\
    \toprule

    \textbf{brain rot} & video, music, song, game \\
    \textbf{canon event} & meme, small talk, gym, boss \\
    \textbf{cheugy} & wedding, cafe, high school reunion, sport \\
    \textbf{copium} & job interview, concert, airport, stock \\
    \textbf{crash out} & during lunch, tournament, moving, amusement park \\
    \textbf{delulu} & office, singer, class, museum \\
    \textbf{grindset} & gym, marathon, classroom, home \\
    \textbf{reheat nachos} & during a trip, roommate, metro, after dinner \\
    
    \bottomrule
    \end{tabular}
}
\caption{\textbf{Extracted keywords for each neologism.} The same set of four keywords per neologism is shown identically to all participants.} 
\label{tab:keywords}
\end{table}

\subsection{Study Interface}
\label{appendix:detailed_study}

We built a custom study interface, with screenshots shown in \autoref{fig:study_flow} following the task flow: \textbf{(1)} Study setup, \textbf{(2)} Introduction with task instructions, \textbf{(3)} Consent to participate, \textbf{(4)} Pre-task survey, \textbf{(5)} Practice session, \textbf{(6)} Familiarity check, \textbf{(7)} Main task (\ding{202} \textbf{Learning} to \ding{204} \textbf{Comprehension}), and \textbf{(8)} Post-task survey. NNS participants were required to complete all pre- and post-task survey questions and a practice session before the main study to ensure they understood the associated tasks.

\paragraph{Pre-Task Survey.}
The pre-task survey includes the following nine questions:
\begin{itemize}[leftmargin=*, itemsep=2pt, parsep=-1pt]
 \item \textbf{Native language:} What are your native language(s)?
 \item \textbf{Age:} What is your age?
 \item \textbf{Gender:} What is your gender?
 \vspace{3pt}
 {\begin{itemize}[label=$\circ$, leftmargin=2em, itemsep=2pt, parsep=0pt]
    \small
    \item Man
    \item Woman
    \item Non-binary
    \item Prefer not to say
  \end{itemize}}
 \item \textbf{Nationality:} What is your nationality?
 
 \item \textbf{English proficiency:} What is your level of proficiency in English? Please select the option that best describes how you use English in your daily life.\footnote{\url{https://en.wikipedia.org/wiki/ILR_scale}}
 \vspace{3pt}
 {\begin{itemize}[label=$\circ$, leftmargin=2em, itemsep=2pt, parsep=0pt]
    \small
    \item 0: No Proficiency
    \item 1: Elementary Proficiency
    \item 2: Limited Working Proficiency
    \item 3: Professional Working Proficiency
    \item 4: Full Professional Proficiency
    \item 5: Native or Bilingual Fluency
  \end{itemize}}

 \item \textbf{Years spent in English-speaking country:} How many years have you lived in an English-speaking country?

  \item \textbf{Information sources:} When you encounter an unfamiliar English neologism or slang term, which sources do you typically use to learn its meaning? Select all that apply.
  \vspace{3pt}
  {\begin{itemize}[label=$\circ$, leftmargin=2em, itemsep=2pt, parsep=0pt]
    \small
    \item AI tools (e.g., AI Overview produced by Google, ChatGPT, Claude)
    \item Translation tools (e.g., Google Translate)
    \item Online dictionaries (e.g., Merriam-Webster, Urban Dictionary)
    \item Reference sites or news outlets (e.g., Wikipedia, BBC, New York Times)
    \item Social media or online communities (e.g., TikTok, X/Twitter, Reddit)
    \item Other (please specify)
  \end{itemize}}

  \item \textbf{English social media usage:} In the past month, how often have you read English posts on social media (e.g., Twitter/X, Reddit, Facebook) in your daily work and life?
  \vspace{3pt}
  {\begin{itemize}[label=$\circ$, leftmargin=2em, itemsep=2pt, parsep=0pt]
    \small
    \item Never: Never in the past month
    \item Rarely: Fewer than once a week
    \item Sometimes: Two or three times a week
    \item Often: More than three times a week, but not every day
    \item Habitually: Almost everyday
  \end{itemize}}

  \item \textbf{English writing frequency:} In the past month, how often did you write in English for the following activities? (Never/Rarely/Sometimes/Often/Habitually)\footnote{\url{https://bilingualism.northwestern.edu/leapq/}}
  \vspace{3pt}
  {\begin{itemize}[label=$\circ$, leftmargin=2em, itemsep=2pt, parsep=0pt]
    \small
    \item Texting or messaging friends
    \item Academic assignments
    \item Work-related communication
    \item Social media posts
    \item Personal writing (e.g., journaling, creative writing)
  \end{itemize}}
\end{itemize}

\paragraph{Post-Task Survey.} The post-task survey includes three sections. We dynamically replace \underline{\textsc{condition}} with the participant's assigned support condition. The Control group is not asked for the first section and have different set of questions for the third section.

\begin{itemize}[leftmargin=*, itemsep=2pt, parsep=-1pt]
\item \textbf{[1] What do you think about the \underline{\textsc{condition}} assistance?} (1:Strongly disagree, 5:Strongly agree)
\begin{itemize}[leftmargin=*, itemsep=2pt, parsep=-1pt]
 \item I am confident in the \underline{\textsc{condition}}. I feel that it works well.
 \item The \underline{\textsc{condition}} is very reliable. I can count on it to be correct all the time.
 \item I will use the \underline{\textsc{condition}} again in the future.
\end{itemize}

\item \textbf{[2] What do you think about the task?} (1:Very low, 5:Very high)
\begin{itemize}[leftmargin=*, itemsep=2pt, parsep=-1pt]
 \item How mentally demanding was the task?
 \item How hard did you have to work to accomplish your level of performance?
\end{itemize}

\item \textbf{[3] Is there anything else you want to share with us?} (Control)
\begin{itemize}[leftmargin=*, itemsep=2pt, parsep=-1pt]
 \item If you could have had any type of assistance during the writing task, what would have been most useful to you?
\end{itemize}

\item \textbf{[3] Is there anything else you want to share with us?} (Treatment)
\begin{itemize}[leftmargin=*, itemsep=2pt, parsep=-1pt]
 \item If the \underline{\textsc{condition}} assistance was helpful in doing the writing task, what aspects were most helpful to you?
 \item If the \underline{\textsc{condition}} assistance wasn't quite what you needed for the writing task, what would have made it better?
\end{itemize}
\end{itemize}

\subsection{Comprehension Writing Collection}
\label{appendix:comprehension}

In this section, we describe the collection process for the writing samples used in the \ding{204} \textbf{Comprehension} stage. We use writing samples produced by NNS participants during three pilot studies, which were subsequently assessed by two NS evaluators per sample along the same dimensions of NS-Rated competence: well-formedness, contextual appropriateness, and understandability. 
Based on these ratings, we select two writing samples per neologism that exhibit comparable levels of well-formedness but markedly different levels of contextual appropriateness\textemdash yielding one \textit{good} sample with higher contextual appropriateness rating and one \textit{poor} sample with lower rating.
On average, the two samples received well-formedness ratings of 8.3/10 and 9.4/10, respectively, and contextual appropriateness ratings of 5.9/10 and 9.6/10, respectively.

\subsection{Error Rate Annotation}
\label{appendix:error_annotation}

From an initial pool of 24 candidate neologisms, we select eight by controlling error rates across the three AI-based support conditions (AI Definition, AI Rewrite, AI Explanation). Specifically, we ensure that each condition includes a relatively balanced number of examples with both correct and incorrect AI outputs. Since none of the conditions directly predict the task outcomes\textemdash both NS-rated competence and NNS comprehension competence\textemdash we employ both automatic and human annotation to obtain proxy measures of support quality, as described below.
 
\paragraph{Automatic annotation.}
We first assess the \textit{relevance} of each AI-based support condition using cosine similarity between its output and a corresponding reference from the Merriam-Webster dictionary.\footnote{For example, \url{https://www.merriam-webster.com/slang/main-character-energy}.} The reference varies by condition:
\begin{itemize}[leftmargin=*, itemsep=2pt, parsep=-1pt]
    \item \raisebox{-0.2em}{\includegraphics[height=1.1em]{figure/logo/definition.png}} \textbf{AI Definition:} similarity to the dictionary's short definition entry
    \item \raisebox{-0.2em}{\includegraphics[height=1.1em]{figure/logo/rewrite.png}} \textbf{AI Rewrite:} similarity to the original social media post
    \item \raisebox{-0.2em}{\includegraphics[height=1.1em]{figure/logo/explanation.png}} \textbf{AI Explanation:} similarity to a GPT-4.1- generated summary of the full dictionary content
\end{itemize}

\begin{figure*}
    \centering
    \setcounter{subfigure}{0}
    \subfigure[\textbf{Introduction (Pre-screening)}]{\includegraphics[width=0.5\textwidth]{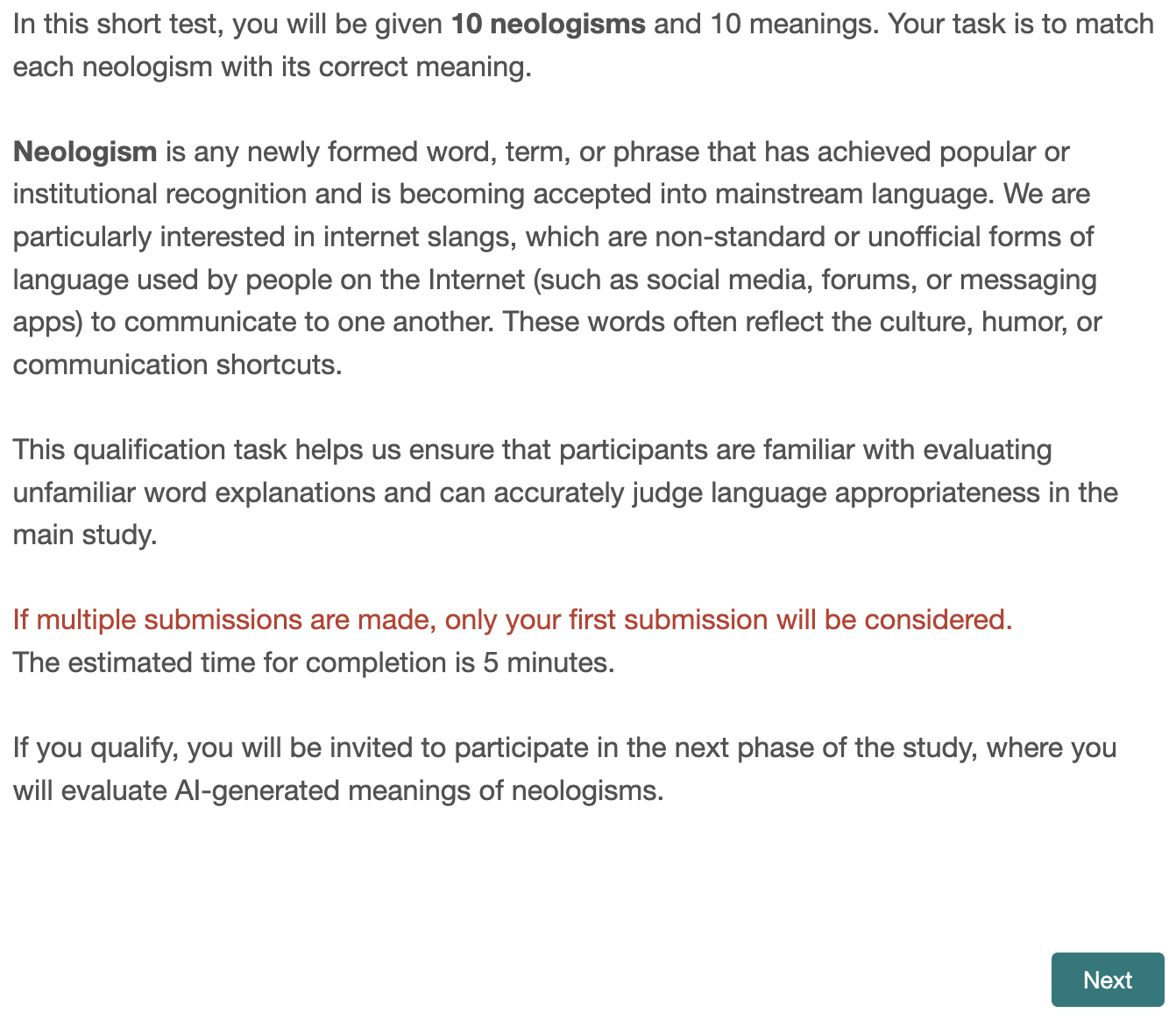}}
    \subfigure[\textbf{Task (Pre-screening)}]{\includegraphics[width=0.5\textwidth]{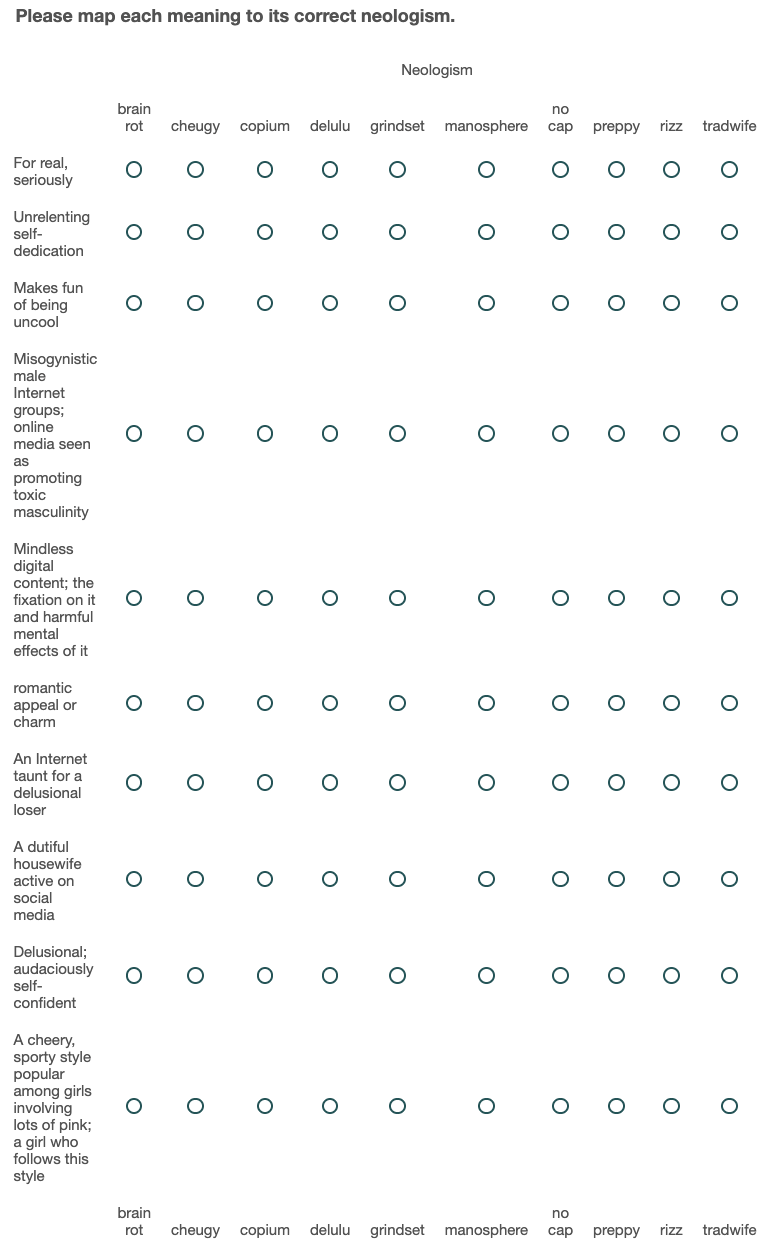}} 
    \caption{\textbf{Screenshots of our NS annotator survey used to compute error rates.} We show examples from the AI Definition condition, including task instructions and main study content for the pre-screening quiz (a–b) and the main annotation task (c–d).}
    \label{fig:ns_error_rate}
\end{figure*}

\paragraph{Human annotation.}
We then evaluate the \textit{accuracy} of each AI-based support condition through human annotation. Recruitment proceeded in two stages to ensure annotator familiarity with the neologisms: (1) a pre-screening meaning-mapping quiz covering ten neologisms, and (2) a main annotation task restricted to participants achieving at least 80\% accuracy in pre-screening. For each condition, we recruit three NS annotators (nine total) to rate how accurately the AI output matches the Merriam-Webster dictionary content on a five-point Likert scale (1:Bad, 5:Perfect). \autoref{fig:ns_error_rate} shows the pre-screening quiz and the main annotation task, which are both implemented in Qualtrics.\footnote{\url{https://www.qualtrics.com/}}

Finally, we combine relevance and accuracy scores from automatic and human annotations to cluster AI support into three quality groups (high, mid, low) per condition. We apply proportional sampling across clusters to select a balanced set of eight neologisms. Results are shown in \autoref{tab:error_rates}.

\definecolor{midgreen}{RGB}{96, 171, 117}

\begin{table}
\centering
\resizebox{\linewidth}{!}{%
    \begin{tabular}{lllllllll}
    \toprule
    \multirow{2}{*}{\textbf{Neologism}} & \multicolumn{2}{c}{\textbf{AI Definition}} & \multicolumn{2}{c}{\textbf{AI Rewrite}} & \multicolumn{2}{c}{\textbf{AI Explanation}} \\
    \cmidrule(lr){2-3}
    \cmidrule(lr){4-5}
    \cmidrule(lr){6-7}
    & \textbf{Auto} & \textbf{Human} & \textbf{Auto} & \textbf{Human} & \textbf{Auto} & \textbf{Human} \\
    \toprule

    \textbf{brain rot} & \cellcolor{red!20} 0.345 & \cellcolor{orange!20} 4.67 & \cellcolor{midgreen!30} 0.943 & \cellcolor{orange!20} 4.33 & \cellcolor{midgreen!30} 0.889 & \cellcolor{midgreen!30} 5.00 \\
    
    \textbf{canon event} & \cellcolor{orange!20} 0.410 & \cellcolor{midgreen!30} 5.00 & \cellcolor{red!20} 0.801 & \cellcolor{orange!20} 4.33 & \cellcolor{red!20} 0.863 & \cellcolor{orange!20} 4.67 \\
    
    \textbf{cheugy} & \cellcolor{red!20} 0.130 & \cellcolor{midgreen!30} 5.00 & \cellcolor{orange!20} 0.867 & \cellcolor{midgreen!30} 5.00 & \cellcolor{orange!20} 0.893 & \cellcolor{midgreen!30} 5.00 \\
    
    \textbf{copium} & \cellcolor{red!20} 0.282 & \cellcolor{orange!20} 4.67 & \cellcolor{orange!20} 0.883 & \cellcolor{orange!20} 4.67 & \cellcolor{midgreen!30} 0.914 & \cellcolor{orange!20} 4.33 \\

    \textbf{crash out} & \cellcolor{midgreen!30} 0.512 & \cellcolor{red!20} 1.00 & \cellcolor{orange!20} 0.903 & \cellcolor{orange!20} 4.67 & \cellcolor{red!20} 0.826 & \cellcolor{red!20} 1.00 \\

    \textbf{delulu} & \cellcolor{orange!20} 0.370 & \cellcolor{midgreen!30} 5.00 & \cellcolor{midgreen!30} 0.991 & \cellcolor{midgreen!30} 5.00 & \cellcolor{orange!20} 0.912 & \cellcolor{red!20} 4.00 \\

    \textbf{grindset} & \cellcolor{red!20} 0.281 & \cellcolor{midgreen!30} 5.00 & \cellcolor{orange!20} 0.924 & \cellcolor{midgreen!30} 5.00 & \cellcolor{midgreen!30} 0.918 & \cellcolor{midgreen!30} 5.00 \\

    \textbf{reheat nachos} & \cellcolor{red!20} 0.295 & \cellcolor{red!20} 1.33 & \cellcolor{red!20} 0.701 & \cellcolor{midgreen!30} 5.00 & \cellcolor{red!20} 0.847 & \cellcolor{red!20} 4.00 \\
    
    \bottomrule
    \end{tabular}
}
\caption{\textbf{Automatic and human annotation results of error rates.} \textbf{Auto:} Automatic annotation (0--1 scale); \textbf{Human:} Human annotation (1--5 scale). We color-code the ratings for each quality group: \raisebox{3pt}{\colorbox{red!20}{ }} (low), \raisebox{3pt}{\colorbox{orange!20}{ }} (mid), and \raisebox{3pt}{\colorbox{midgreen!30}{ }} (high).} 
\label{tab:error_rates}
\end{table}

\subsection{Study Material Details}
\label{appendix:support_conditions}

\begin{figure*}
    \centering
    \setcounter{subfigure}{0}
    \subfigure[\textbf{\raisebox{-0.2em}{\includegraphics[height=1.1em]{figure/logo/definition.png}} AI Definition}]{\includegraphics[width=0.6\textwidth]{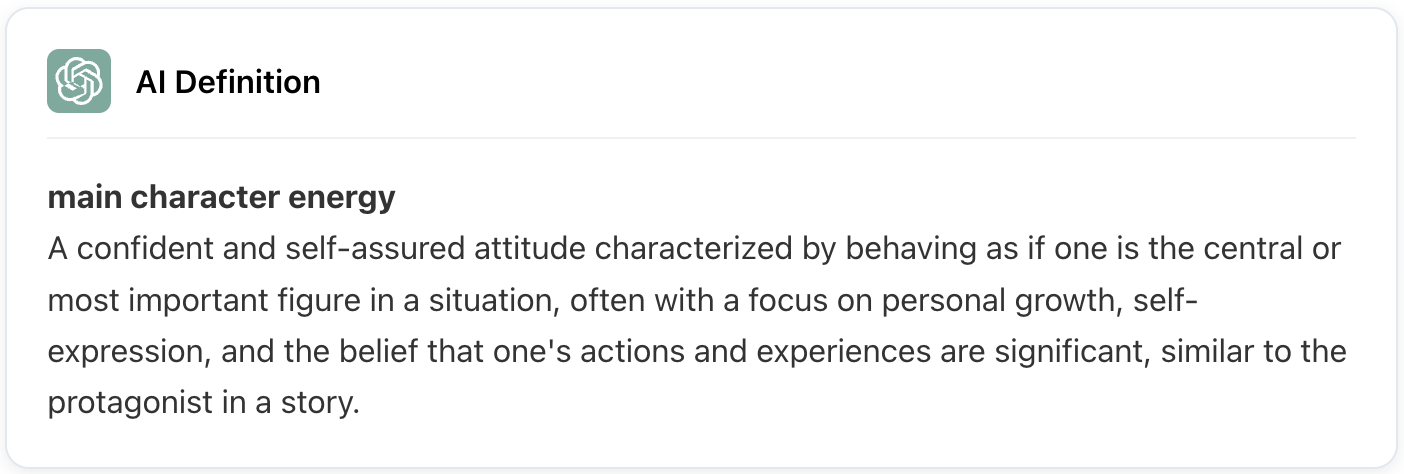}}
    \subfigure[\textbf{\raisebox{-0.2em}{\includegraphics[height=1.1em]{figure/logo/rewrite.png}} AI Rewrite}]{\includegraphics[width=0.6\textwidth]{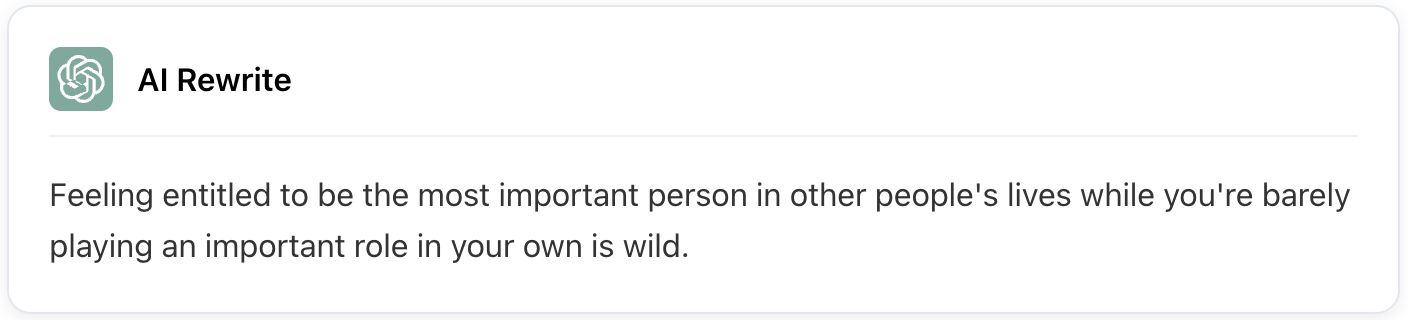}} 
    \subfigure[\textbf{\raisebox{-0.2em}{\includegraphics[height=1.1em]{figure/logo/explanation.png}} AI Explanation}]{\includegraphics[width=0.6\textwidth]{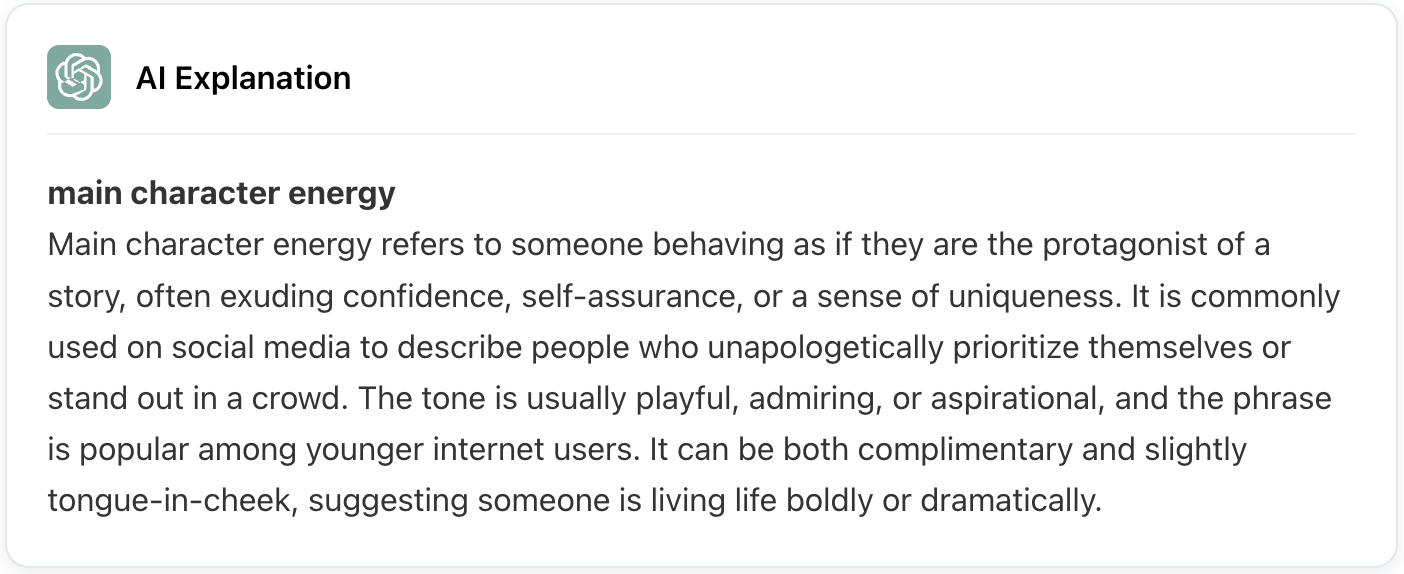}} 
    \subfigure[\textbf{\raisebox{-0.2em}{\includegraphics[height=1.1em]{figure/logo/dictionary.png}} Non-AI Dictionary}]{\includegraphics[width=0.6\textwidth]{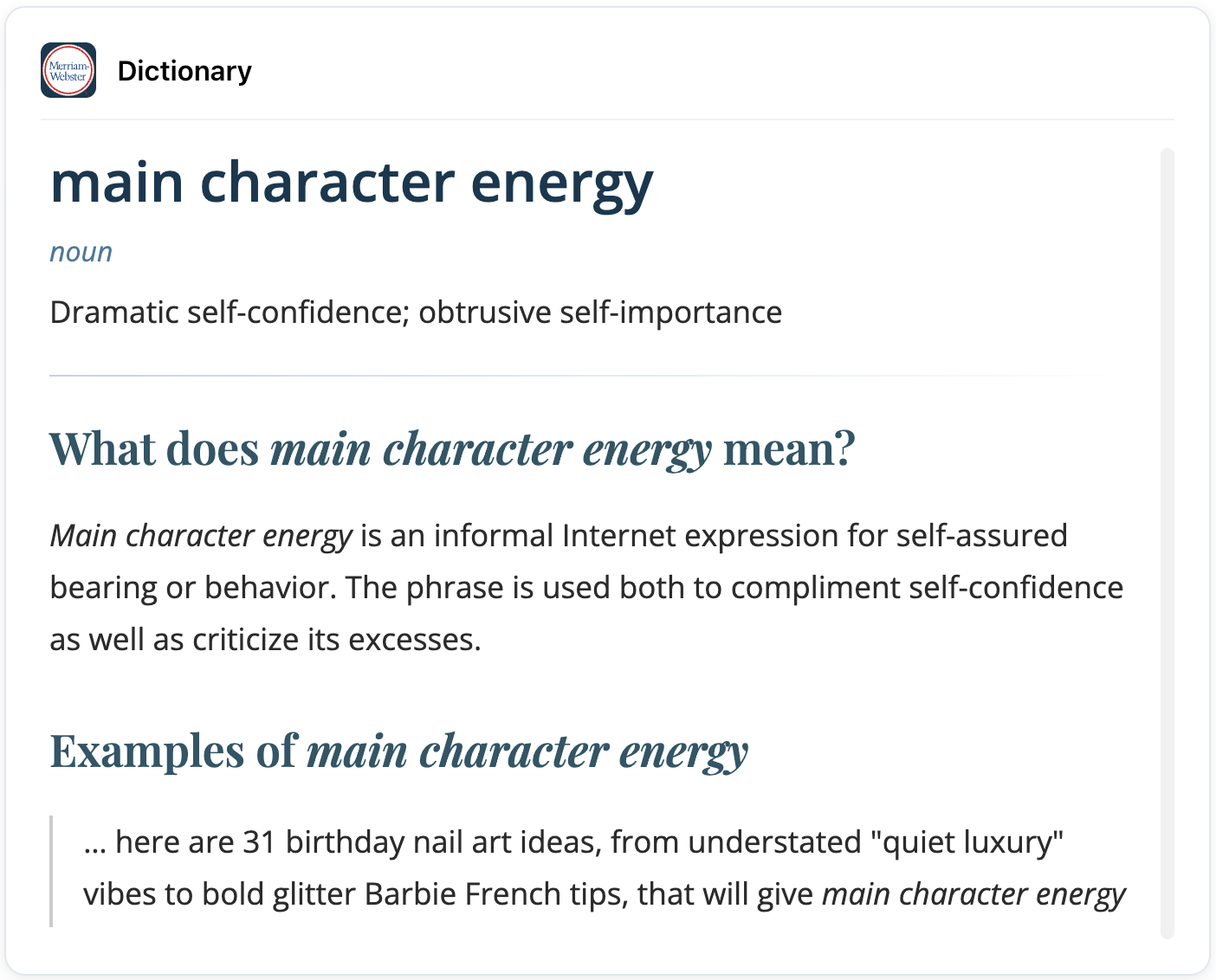}} 
    \caption{\textbf{Support conditions provided to NNS participants.} We illustrate each condition using the practice-session neologism ``\underline{main character energy}.''}    
    \label{fig:support_conditions}
\end{figure*}

We illustrate how each type of support is presented to NNS participants in \autoref{fig:support_conditions}. We also show the AI-based support conditions (AI Definition, AI Rewrite, AI Explanation) and original social media posts used for each neologism in \autoref{tab:support_conditions}. Non-AI Dictionary group participants are provided with the corresponding Merriam-Webster dictionary page.

\subsection{Power Analysis}
\label{appendix:power_analysis}

We determine the sample size for the main human study with NNS participants using a power analysis. Given five conditions (one control and four treatments), we use \texttt{FTestAnovaPower}\footnote{\url{https://www.statsmodels.org/dev/generated/statsmodels.stats.power.FTestAnovaPower.html}} with $\alpha$=0.05 and power=0.8. Assuming a medium effect size (Cohen's $f$=0.25), the required total sample size is 196, corresponding to approximately 40 participants per condition.

\subsection{NS Evaluator Survey Details}
\label{appendix:detailed_ns_survey}

We detail the survey used in the \ding{205} \textbf{NS Evaluation} stage, which was implemented in Qualtrics.
Each NS evaluator rates 26 NNS-produced writing samples associated with one neologism; the neologism assignment is randomized across evaluators.
\autoref{fig:ns_survey} illustrates the task instructions and main workflow. Evaluators first familiarize themselves with the target neologism using a provided Merriam-Webster dictionary entry, along with any additional tools as needed. They then write their own scenario and message using the neologism (later used to compare NNS- and NS-produced writing samples; \S\ref{res:3}). Finally, evaluators rate each NNS-produced writing sample on communicative competence across five questions and report their confidence. After the main evaluation task, we also collect open-ended comments on their general thoughts and reactions while reading the messages.

\paragraph{Pre-Task Survey.}
The pre-task survey includes the following five questions:
\begin{itemize}[leftmargin=*, itemsep=2pt, parsep=-1pt]
 \item \textbf{Age:} What is your age?
 \item \textbf{Gender:} What is your gender?
 \vspace{3pt}
 {\begin{itemize}[label=$\circ$, leftmargin=2em, itemsep=2pt, parsep=0pt]
    \small
    \item Man
    \item Woman
    \item Non-binary
    \item Prefer not to say
  \end{itemize}}
 \item \textbf{Nationality:} What is your nationality?
 \item \textbf{Native language:} What are your native language(s)?
  \item \textbf{English social media usage:} In the past month, how often have you read English posts on social media (e.g., Twitter/X, Reddit, Facebook) in your daily work and life?
  \vspace{3pt}
  {\begin{itemize}[label=$\circ$, leftmargin=2em, itemsep=2pt, parsep=0pt]
    \small
    \item Never: Never in the past month
    \item Rarely: Fewer than once a week
    \item Sometimes: Two or three times a week
    \item Often: More than three times a week, but not every day
    \item Habitually: Almost everyday
    \end{itemize}}
\end{itemize}

\subsection{Participants Details}
\label{appendix:participants}

All human studies, including the main study with NNS participants (\S\ref{appendix:detailed_study}), human annotation for error rate control (\S\ref{appendix:error_annotation}), and the NS evaluation (\S\ref{appendix:detailed_ns_survey}) are conducted on the Prolific platform.\footnote{\url{https://www.prolific.com/}}

\paragraph{\raisebox{-0.2em}{\includegraphics[height=1.1em]{figure/logo/nns.png}} NNS Participants.}
To ensure high-quality responses, we limit participation to Prolific users with an approval rate above 95\% and at least 10 prior submissions. Participants receive a base payment of USD 6 for 30 minutes of participation, with an additional USD 2 performance-based bonus for those who passed both attention checks. Including Prolific platform fees, the total cost of the main task was USD 2,400. Detailed pre-survey statistics by language group are reported in \autoref{tab:pre_survey}.

\paragraph{\raisebox{-0.2em}{\includegraphics[height=1.1em]{figure/logo/ns.png}} NS Evaluators.}
We restrict participation to Prolific users with an approval rate above 98\% and at least 100 prior submissions, and limit each evaluator to a single evaluation task for one neologism. Evaluators receive a base payment of USD 6 for 30-minute task. Including Prolific platform fees, the total cost was USD 1,411.

Among the 160 NS evaluators (144 for evaluating NNS-produced writing and 16 for NS-produced writing), the mean age was 32.8 years (SD=6.46). Gender identities are reported as man (81, 50.6\%), woman (75, 46.9\%), non-binary (3, 1.9\%), and prefer not to say (1, 0.6\%). Reported nationalities included American (148, 92.5\%), Caucasian (3, 1.9\%), African American (3, 1.9\%), Mexican (2, 1.3\%), Fijian (1, 0.6\%), Nigerian (1, 0.6\%), Peurto Rican (1, 0.6\%), and Asian (1, 0.6\%). Native language were reported as English (156, 97.5\%), English and Spanish (3, 1.9\%), and English and Telugu (1, 0.6\%).
Monthly use of English social media content was heavily skewed toward higher frequencies: no evaluators reported never or rarely engaging, 10 (6.3\%) reported sometimes, 17 (10.6\%) often, and 133 (83.1\%) reported using English social media almost daily.

\section{Detailed Results}
\label{appendix:detailed_results}

\subsection{AI-Rated Communicative Competence}
\label{appendix:llm_as_judge}

\begin{figure*}
    \centering
    \setcounter{subfigure}{0}
    \subfigure[\textbf{Introduction}]{\includegraphics[width=0.8\textwidth]{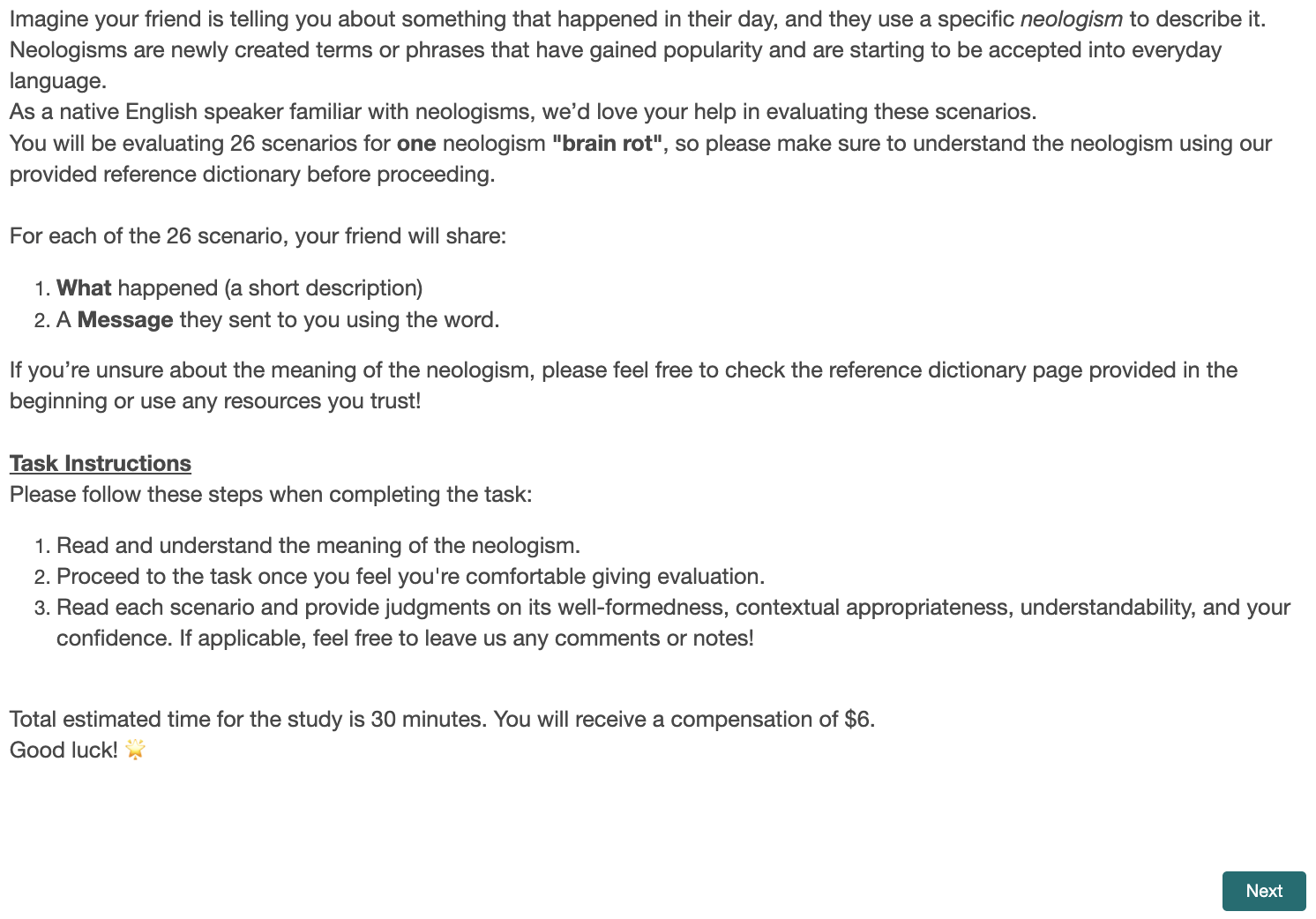}}
    \subfigure[\textbf{Familiarize with Neologism}]{\includegraphics[width=0.8\textwidth]{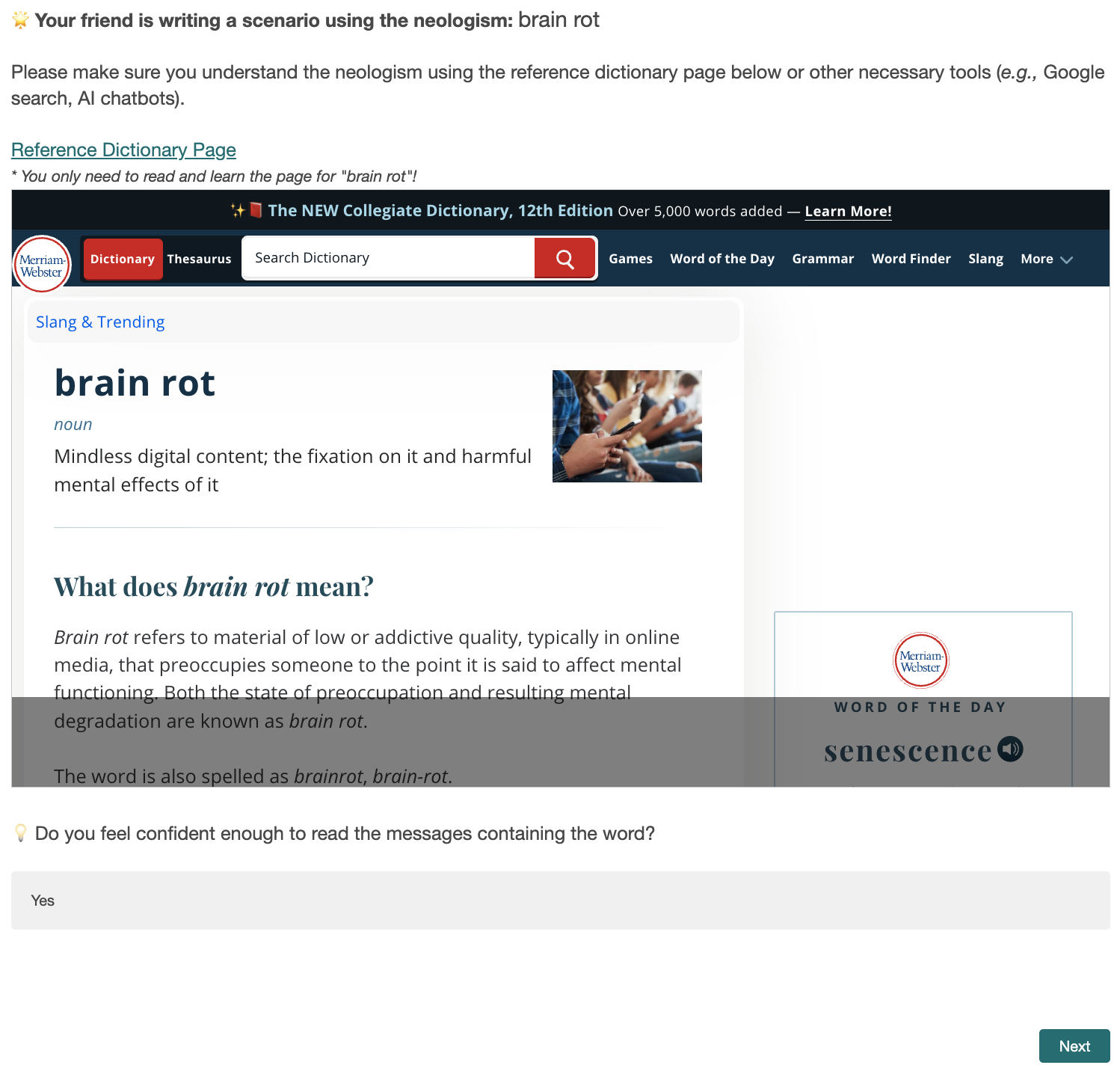}} 
    \caption{\textbf{Screenshots of our NS evaluation survey, organized according to the task flow.} NS evaluators first consult the dictionary page to familiarize themselves with the neologism (b), then provide their own writing sample (c) and evaluate NNS-produced writing samples (d).}
    \label{fig:ns_survey}
\end{figure*}

Having NS evaluators to rate the NNS-produced writing samples is ideal, but involving them is also practically challenging. 
We investigate using NNS self-reported perceptions as an alternative measure of communicative competence (\S\ref{res:2}), but find they are not reliable proxies.
As a middle ground, we test AI tools (i.e., LLM-as-judge) for rating NNS writing. 
We prompt GPT-5\footnote{\texttt{gpt-5-2025-08-27}} to rate each of the three communicative competence dimensions by answering the same five questions used in the \ding{205} \textbf{NS Evaluation} stage (\S\ref{sec:dependent_var}). We collect two ratings per question, resulting in 3,744 observations. 
The prompt is in Appendix~\ref{appendix:prompts_2}. We use the instructions and questions given to NS evaluators (\autoref{fig:ns_survey}).
 
In \autoref{tab:llm_as_judge}, we compare the mean NS- and AI-rated competence scores for NNS-produced writing samples across support conditions. AI ratings are higher on average for well-formedness, contextual appropriateness, and understandability, whereas NS evaluators report higher confidence. Although the exact rankings differ, AI ratings correctly identify the best (AI Explanation) and worst (Control) conditions, matching NS-rated competence.

\begin{table}
\centering
\resizebox{\linewidth}{!}{%
    \begin{tabular}{llllllllll}
    \toprule
    \multirow{2}{*}{\textbf{Condition}} & \multicolumn{4}{c}{\textbf{NS-rated}} & \multicolumn{4}{c}{\textbf{AI-rated}} \\
    \cmidrule(lr){2-5}
    \cmidrule(lr){6-9}
    & \textbf{WF} & \textbf{CA} & \textbf{U} & \textbf{C} & \textbf{WF} & \textbf{CA} & \textbf{U} & \textbf{C} \\
    \toprule

    \textbf{Control} & \underline{7.05} & \underline{6.44} & \underline{7.17} & \underline{4.17} & \underline{8.93} & \underline{7.52} & \underline{8.18} & 4.10 \\
    \textbf{AI Definition} & 7.32 & 6.93 & 7.50 & 4.23 & 9.03 & 8.16 & 8.56 & \textbf{4.20} \\
    \textbf{AI Rewrite} & 7.42 & 7.06 & 7.62 & 4.24 & 9.04 & 7.77 & 8.31 & 4.12 \\
    \textbf{AI Explanation} & \textbf{7.74} & \textbf{7.43} & \textbf{7.98} & \textbf{4.23} & \textbf{9.18} & \textbf{8.24} & \textbf{8.59} & 4.19 \\
    \textbf{Non-AI Dictionary} & 7.36 & 7.28 & 7.78 & 4.23 & 9.02 & 7.72 & 8.32 & 4.12 \\
    \rowcolor{gray!15}
    \textbf{Avg.} & 7.19 & 7.04 & 7.34 & 4.17 & 9.04 & 7.88 & 8.39 & 4.15 \\

    \bottomrule
    \end{tabular}
}
\caption{\textbf{NS-rated vs. AI-rated competence.} \textbf{WF:} Well-formedness; \textbf{CA:} Contextual appropriateness; \textbf{U:} Understandability (averaged); \textbf{C:} Confidence. \textbf{Best} and \underline{worst} scores for each metric are highlighted.}
\label{tab:llm_as_judge}
\end{table}

\subsection{Language Group-specific Results}
\label{appendix:lang_group_results}

\begin{figure}
    \centering
    \subfigure[\textbf{Spanish}]{\includegraphics[width=0.5\textwidth]{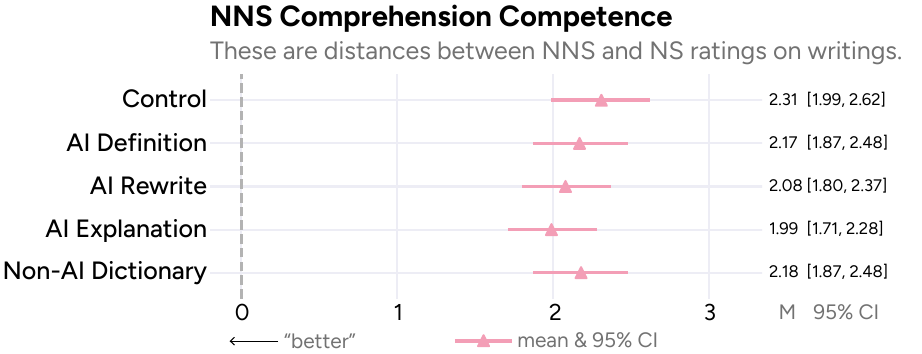}}
    \subfigure[\textbf{German}]{\includegraphics[width=0.5\textwidth]{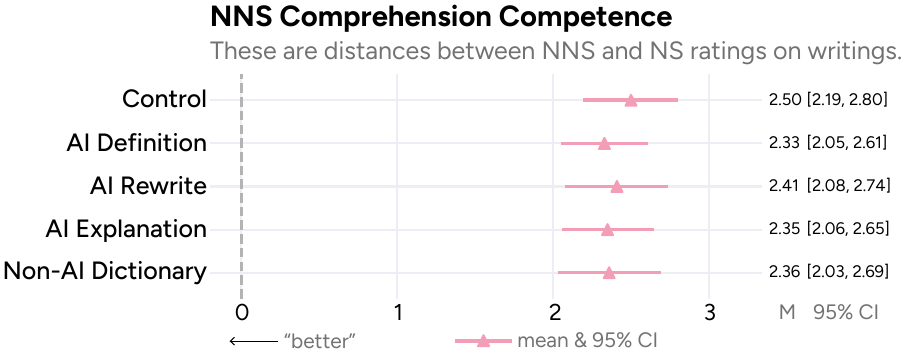}}
    \subfigure[\textbf{Chinese}]{\includegraphics[width=0.5\textwidth]{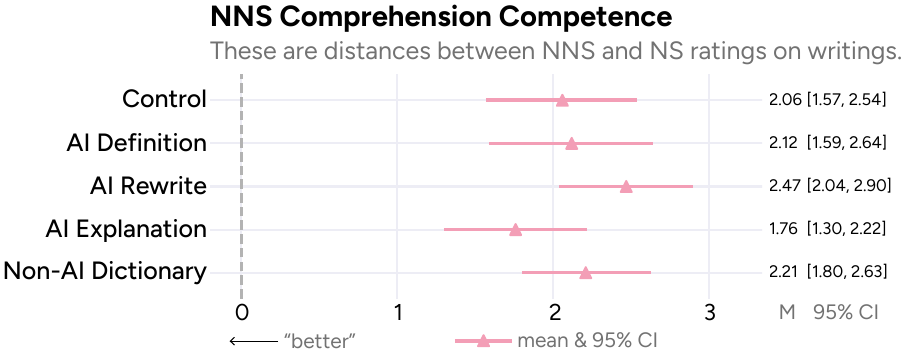}}
    \caption{\textbf{NNS comprehension distance for each language group.}}
    \label{fig:mcq-results-lang}
\end{figure}

\begin{figure}
    \centering
    \subfigure[\textbf{Spanish}]{\includegraphics[width=0.5\textwidth]{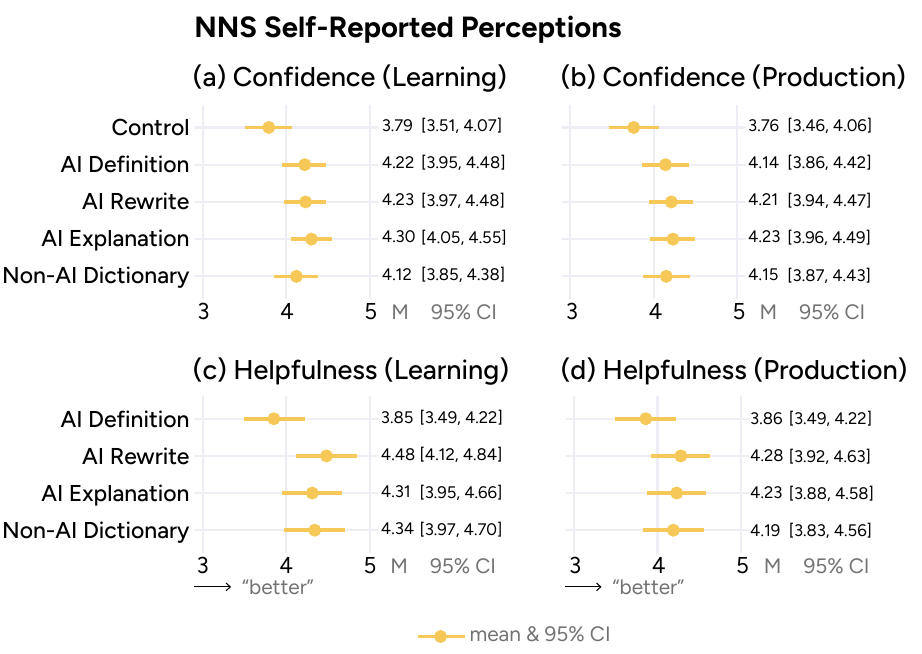}}
    \subfigure[\textbf{German}]{\includegraphics[width=0.5\textwidth]{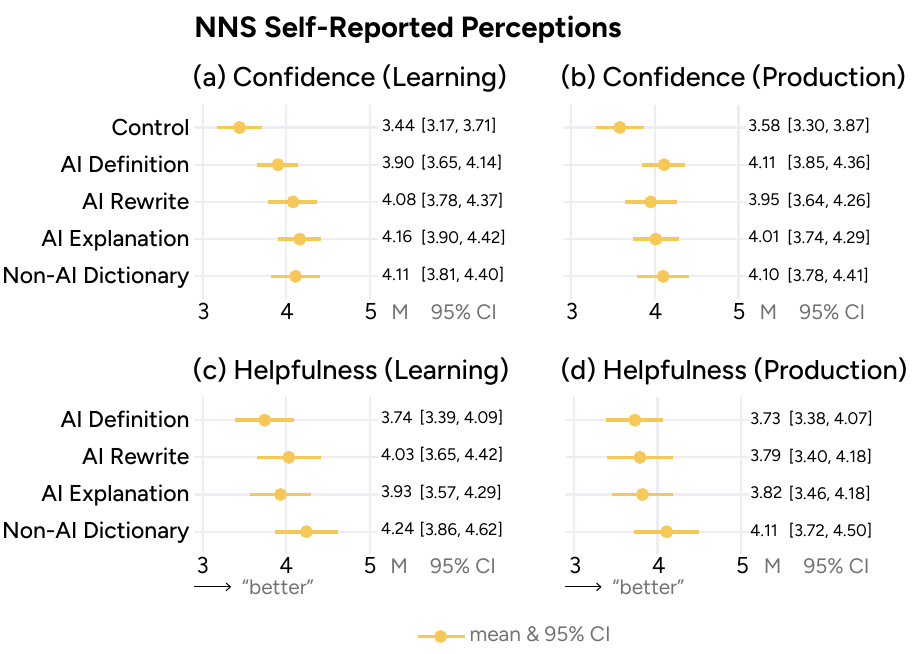}}
    \subfigure[\textbf{Chinese}]{\includegraphics[width=0.5\textwidth]{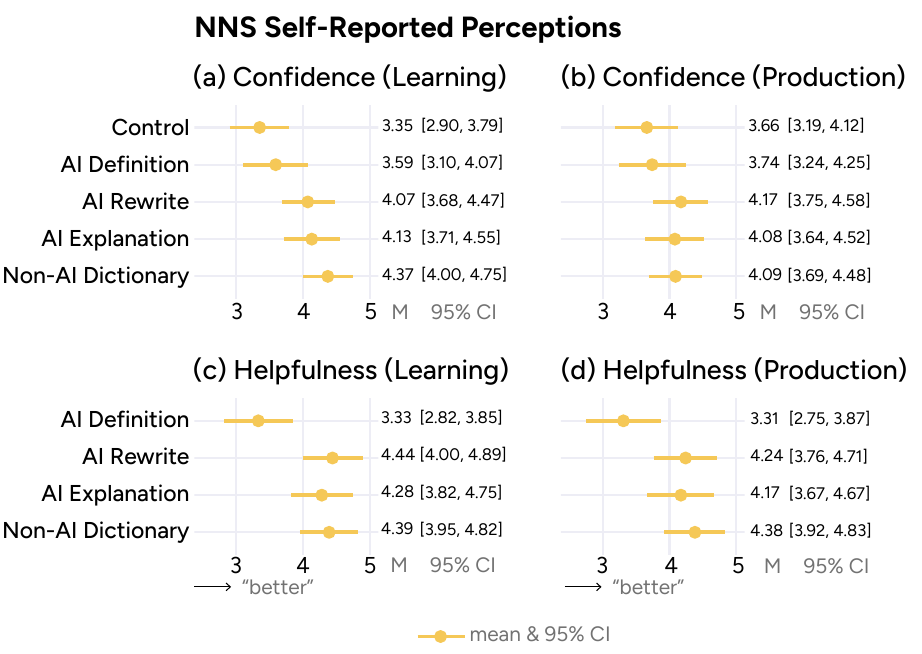}}
    \caption{\textbf{NNS self-reported confidence and helpfulness perceptions for each language group.}}
    \label{fig:nns-self-results-lang}
\end{figure}

\begin{figure*}
    \centering
    \setcounter{subfigure}{0}
    \subfigure[\textbf{Spanish}]{\includegraphics[width=\textwidth]{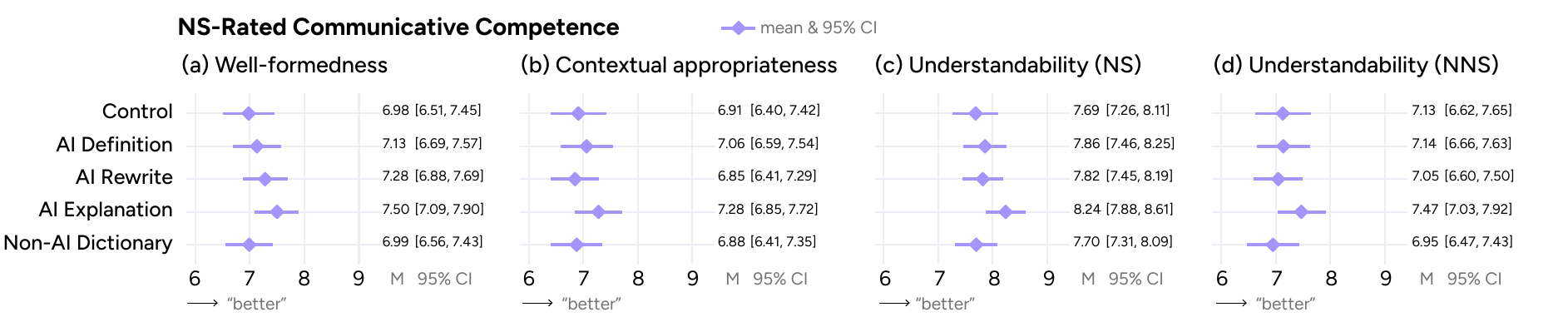}}
    \subfigure[\textbf{German}]{\includegraphics[width=\textwidth]{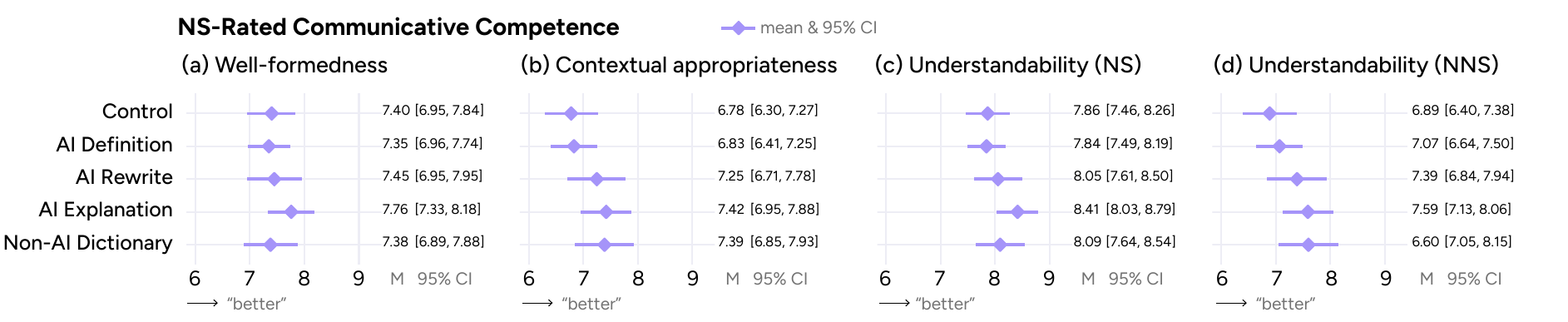}}
    \subfigure[\textbf{Chinese}]{\includegraphics[width=\textwidth]{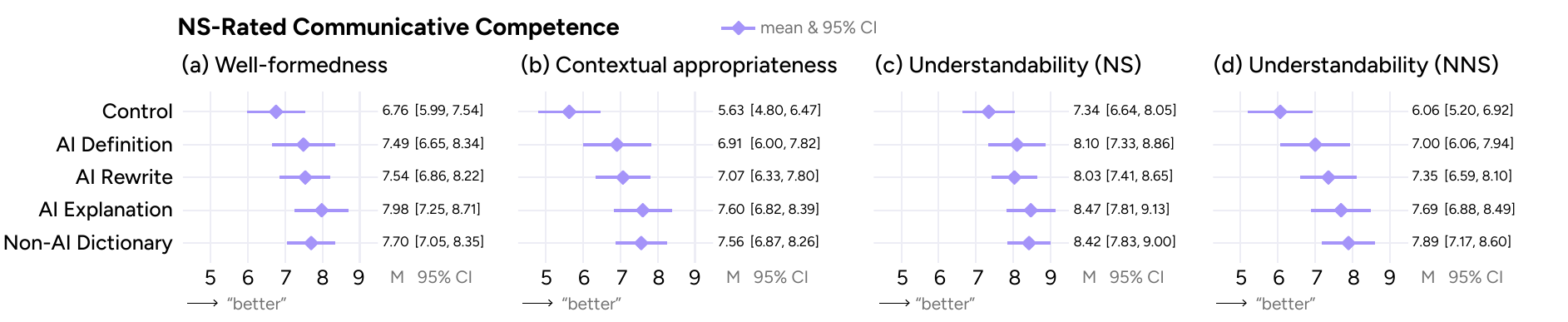}}
    \caption{\textbf{NS-rated competence for each language group.}}
    \label{fig:ns-results-lang}
\end{figure*}

We perform a Type III Analysis of Variance (ANOVA) using Satterthwaite's method. The interaction between condition and language is not significant for all the mixed-effects models, indicating that the effect of the condition is consistent across three languages.
We present results specific to each language group on NS-rated communicative competence (\autoref{fig:ns-results-lang}), NNS comprehension competence (\autoref{fig:mcq-results-lang}), and NNS self-reported perceptions (\autoref{fig:nns-self-results-lang}).

\begin{table*}
\centering
\resizebox{\linewidth}{!}{%
    \begin{tabular}{lllllll}
    \toprule
    \textbf{Measures} & \textbf{Language} & \textbf{Control} & \textbf{AI Definition} & \textbf{AI Rewrite} & \textbf{AI Explanation} & \textbf{Non-AI Dictionary} \\
    \toprule

     \multirow{3}{*}{\textbf{Confidence (↑)}} & Spanish & -& 3.50 & 4.27	& 4.29	& 4.60 \\
     & German & -&3.48&3.53&3.76&4.57 \\
     & Chinese & -&2.80&4.25&4.00&4.22 \\
     \midrule

     \multirow{3}{*}{\textbf{Reliance (↑)}} & Spanish & -&3.00&3.90&3.75&4.40 \\
     & German & -&2.60&3.13&2.90&4.14 \\
     & Chinese & -&2.40&3.63&3.71&4.22 \\
    \midrule

    \multirow{3}{*}{\textbf{Trust for Future Use (↑)}} & Spanish & -&3.75&4.05&3.92&4.65 \\
     & German & -&3.32&3.00&3.71&4.29 \\
     & Chinese & -&4.00&3.88&4.29&4.22 \\
    \midrule

     \multirow{3}{*}{\textbf{Mental Burden (↓)}} & Spanish & 3.94&3.30&3.23&3.38&3.75 \\
     & German & 3.30&3.44&3.33&3.43&3.79 \\
     & Chinese & 4.17&3.80&3.50&3.71&3.78 \\
    \midrule

    \multirow{3}{*}{\textbf{Task Difficulty (↓)}} & Spanish & 4.00&3.80&3.59&3.46&3.70 \\
     & German & 3.20&3.48&3.47&3.33&3.79 \\
     & Chinese & 4.50&4.00&3.38&3.86&3.33 \\
    
    \bottomrule
    \end{tabular}
}
\caption{\textbf{Post-survey statistics for each language group.} All metrics are measured on a 1--5 scale.} 
\label{tab:post_survey_lang}
\end{table*}

\subsection{Post-task Survey}
\label{appendix:detailed_post}
We present detailed post-task survey statistics for each language group in \autoref{tab:post_survey_lang}.

\begin{table*}
\centering
\resizebox{\linewidth}{!}{%
    \begin{tabular}{llllllll}
    \toprule
    \textbf{Measures} & \textbf{Control} & \textbf{AI Definition} & \textbf{AI Rewrite} & \textbf{AI Explanation} & \textbf{Non-AI Dictionary} \\
    \toprule

    \textbf{Keyword Inclusion Rate} & \textbf{0.262} & 0.125 & 0.239 & 0.144 & 0.173 \\
    \textbf{Similarity to Social Media Post} & \textbf{0.313} & 0.297 & 0.311 & 0.299 & 0.304 \textbf{}\\

    \bottomrule
    \end{tabular}
}
\caption{\textbf{Descriptive statistics for NNS-produced messages.} \textbf{Highest} scores for each metric are highlighted.} 
\label{tab:qualitative}
\end{table*}

\subsection{Descriptive Statistics of Messages}
\label{appendix:nns_writings}

We report descriptive statistics for NNS- and NS-produced messages in terms of length and lexical diversity, measured via Corrected Type Token Ratio (CTTR) \citep{Carroll-cttr}. NNS-produced messages are generally longer than those produced by NS: the average word count is 27.0 for Control, 34.2 for AI Definition, 29.5 for AI Rewrite, 32.7 for AI Explanation, and 27.4 for Non-AI Dictionary, compared to 22.0 for NS messages.
NS-produced messages are also lexically less diverse overall (CTTR=2.93) than NNS messages across all conditions (3.19 for Control, 3.50 for AI Definition, 3.34 for AI Rewrite, 3.44 for AI Explanation, and 3.25 for Non-AI Dictionary).

We also compute two metrics for NNS-produced messages: (1) Keyword inclusion rate, which quantifies the proportion of messages containing any of the keywords we provide to support writing during the \ding{203} \textbf{Production} stage (\autoref{tab:keywords}), using exact string matching; and (2) Similarity to the social media post (\autoref{tab:support_conditions}), which we compute via embeddings\footnote{\url{https://huggingface.co/sentence-transformers/all-MiniLM-L6-v2}} and measure with cosine similarity.

As shown in \autoref{tab:qualitative}, the keyword inclusion rate is highest for the Control group (0.262), likely because participants, having no additional support, rely more on the keywords to initiate their writing. Similarly, message similarity to the social media post is highest for the Control group (0.313), followed by AI Rewrite (0.311) and Non-AI Dictionary (0.304).

\subsection{Qualitative Examples}
\label{appendix:detailed_examples}

In \autoref{tab:detailed_examples}, we present examples for each neologism, showing NNS- and NS-produced writing that received differing NS-rated competence.

\subsection{Open-ended Feedback}
\label{appendix:detailed_comments}

We present detailed comments from NNS participants on (1) which aspects of each condition were helpful and (2) which aspects fell short and how they could be improved. Representative feedback for both parts per condition is shown in \autoref{tab:detailed_comments}.

\begin{table}
\centering
\resizebox{\linewidth}{!}{%
    \begin{tabular}{llllllll}
    \toprule
    \textbf{Condition} & $\mathbf{N}$ & \textbf{Mean} & \textbf{95\% CI} & \textbf{\textit{p}-value} \\
    \toprule

    \textbf{NNS (Control)} & 656 & 6.44 & [6.06, 6.82] &  \\
    \textbf{NNS (AI Definition)} & 800 & 6.93 & [6.55, 7.32] & n.s. \\
    \textbf{NNS (AI Rewrite)} & 736 & 7.06 & [6.70, 7.41] & n.s. \\
    \textbf{NNS (AI Explanation)} & 848 & 7.43 & [7.08, 7.79] & \textit{p}<.001 \\
    \textbf{NNS (Non-AI Dictionary)} & 704 & 7.28 & [6.93, 7.63] & \textit{p}=.028 \\
    \textbf{NNS (Avg.)} & 3,744 & 7.04 & [6.99, 7.29] & n.s. \\
    \textbf{AI} & 80 & 7.12 & [6.37, 7.87] & n.s. \\
    \textbf{NS} & 288 & 7.67 & [7.36, 7.98] & \textit{p}<.001 \\

    \bottomrule
    \end{tabular}
}
\caption{\textbf{Contextual appropriateness dimension of NS-rated competence for AI-generated vs. NNS- vs. NS-produced writing samples.} $\mathbf{N}$: Number of observations; \textbf{Mean:} Mean values; \textbf{95\% CI:} 95\% Confidence Intervals; \textbf{\textit{p}-value:} \textit{p}-values in comparison to the \textbf{NNS (Control)} condition.} 
\label{tab:ai_nns_ns_writing}
\end{table}

\subsection{AI vs. NNS vs. NS Writing Samples}
\label{appendix:ai_writing}

\begin{figure*}
    \centering
    \setcounter{subfigure}{0}
    \subfigure[\textbf{Study Setup} (\textit{All subsequent pages display the character selected by the NNS participant.})]{\includegraphics[width=\textwidth]{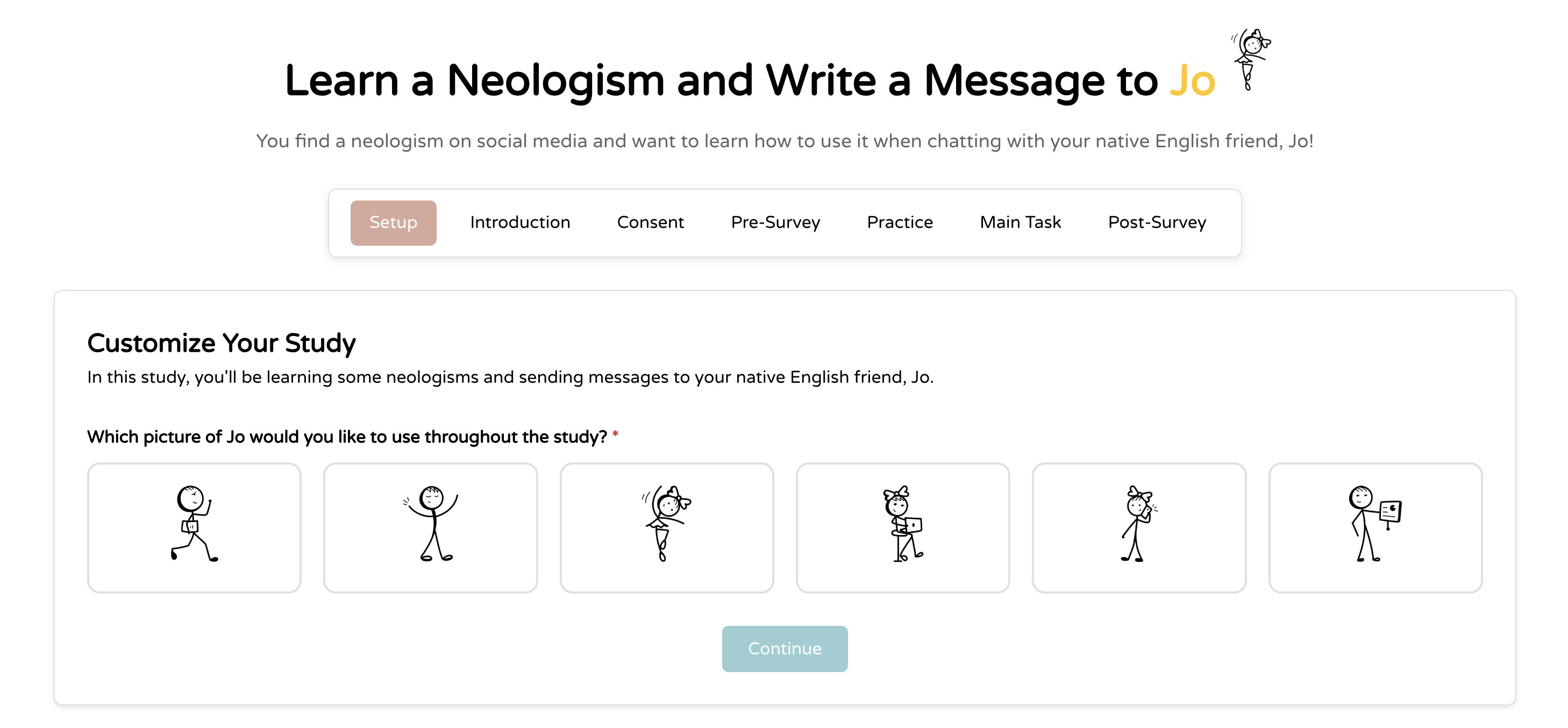}}
    \subfigure[\textbf{Introduction}]{\includegraphics[width=\textwidth]{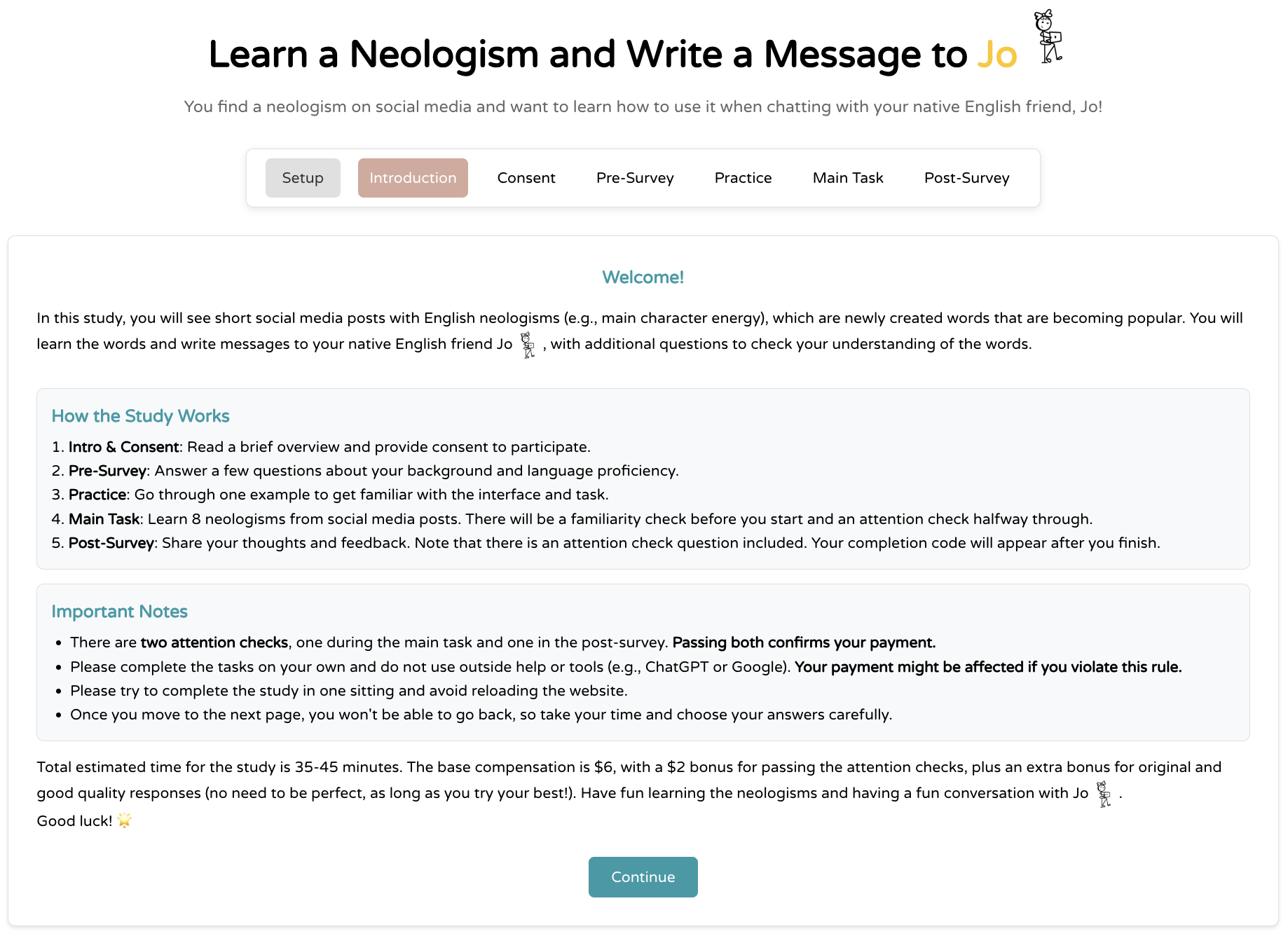}} 
    \caption{\textbf{Screenshots of our annotation interface, organized according to the task flow.} We exclude Consent to Participate page for anonymization.}
    \label{fig:study_flow}
\end{figure*}

To evaluate the effectiveness of current AI tools in generating contextual usage examples, we extend our comparison of NS-rated competence (\S\ref{res:3}) to include AI-generated writing samples. Using GPT-5, we generate five samples per neologism, each rated twice, yielding 80 observations. The prompt is shown in Appendix~\ref{appendix:prompts_3}. We use the same instructions given to NNS participants during the \ding{203} \textbf{Production} stage (\autoref{fig:study_flow}, (g)).

As shown in \autoref{tab:ai_nns_ns_writing}, we compare NS-rated contextual appropriateness across all NNS condition groups, AI-generated, and NS-produced writing samples. Differences are assessed using a Kruskal-Wallis test for ordinal, non-parametric data.
AI-generated samples have a higher average than Control, AI Definition, and AI Rewrite, but the difference is not statistically significant, likely due to their high variance (i.e., wider 95\% CI). This suggests that, despite current AI tools generating plausible examples, these are in fact perceived as similarly or even less communicatively competent than the top-performing NNS support condition (AI Explanation).

\section{Usage of Large Language Models}
We used LLMs to support and refine the writing of our work. Importantly, we did not rely on them to generate content or sentences from scratch. Instead, we employed them primarily to polish the clarity and expression of how we presented our results. In addition, we used them for stylistic adjustments, such as improving readability and removing layout issues (e.g., widows and orphans).

\begin{figure*}
    \setcounter{subfigure}{2}
    \centering
    \subfigure[\textbf{Introduction (NS Annotation)}]{\includegraphics[width=0.6\textwidth]{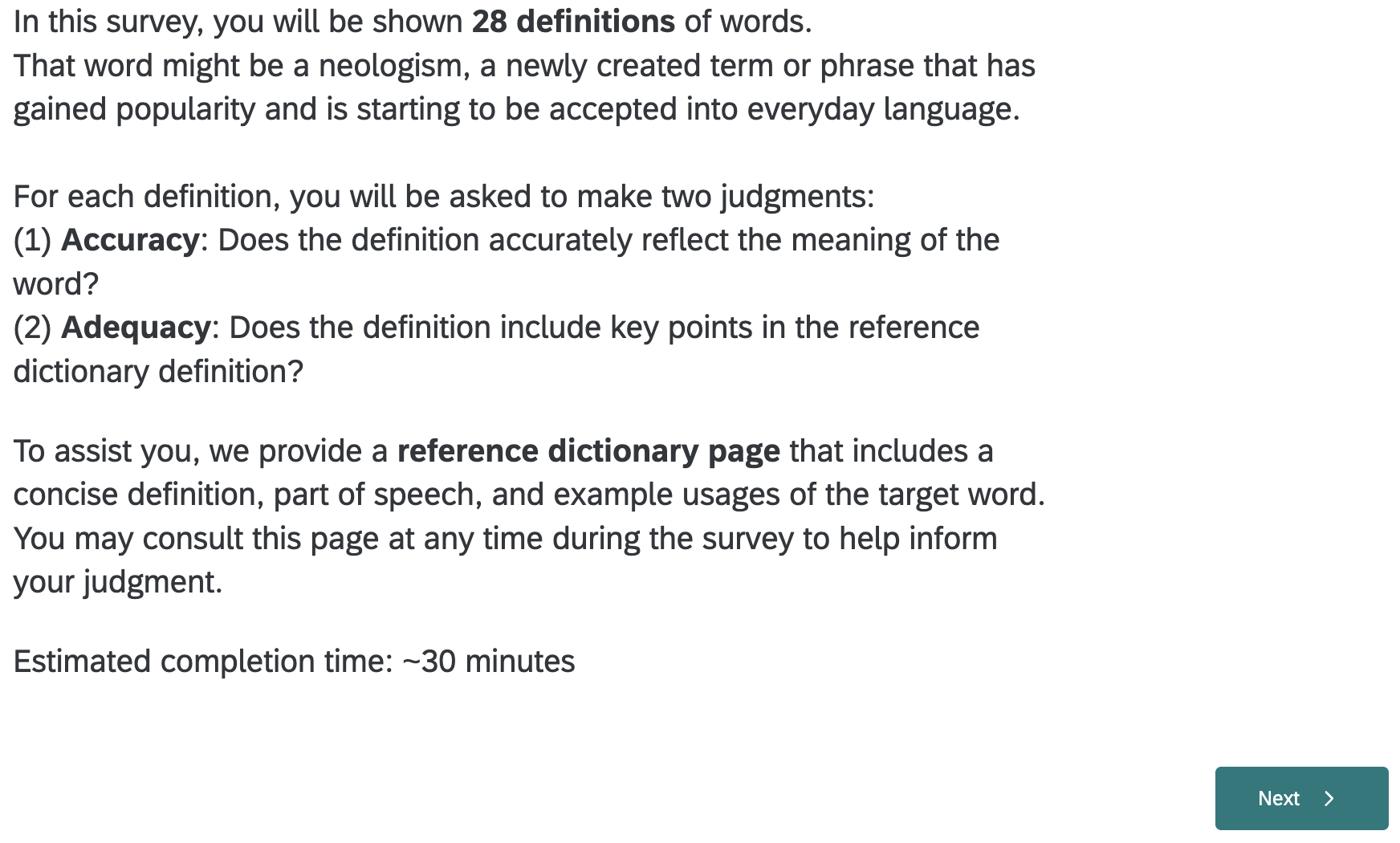}} 
    \subfigure[\textbf{Main Annotation} (\textit{left}: annotation material, \textit{right}: questions)]{
        \begin{minipage}{\linewidth}
            \centering
            \includegraphics[width=0.49\linewidth]{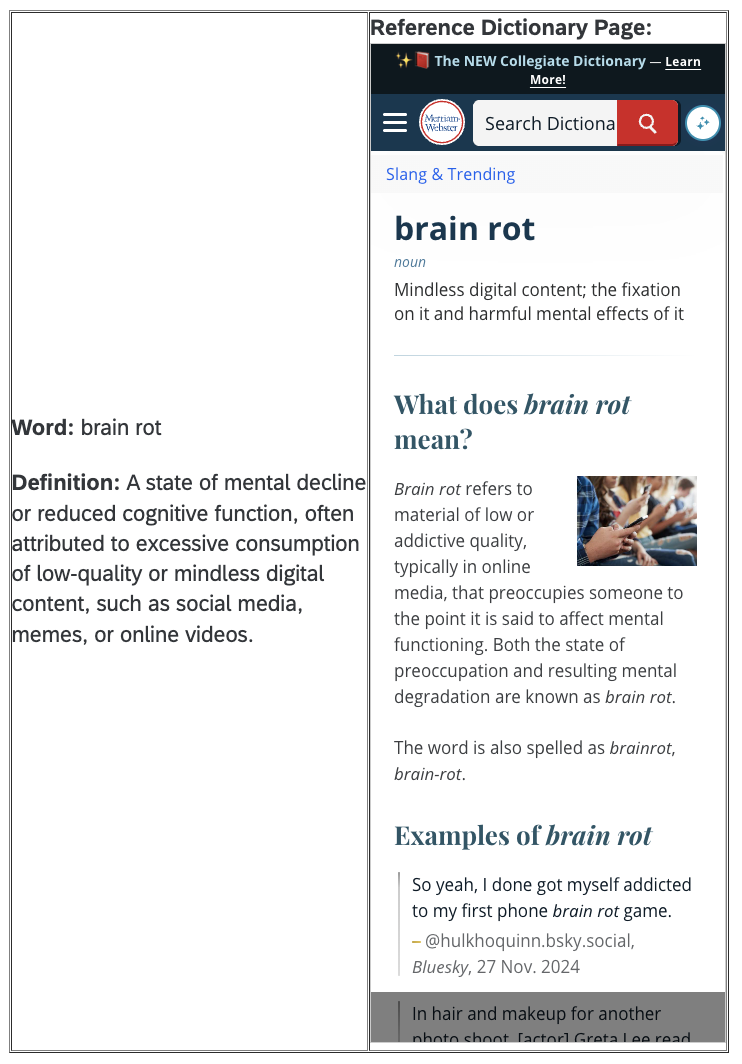}
            \includegraphics[width=0.49\linewidth]{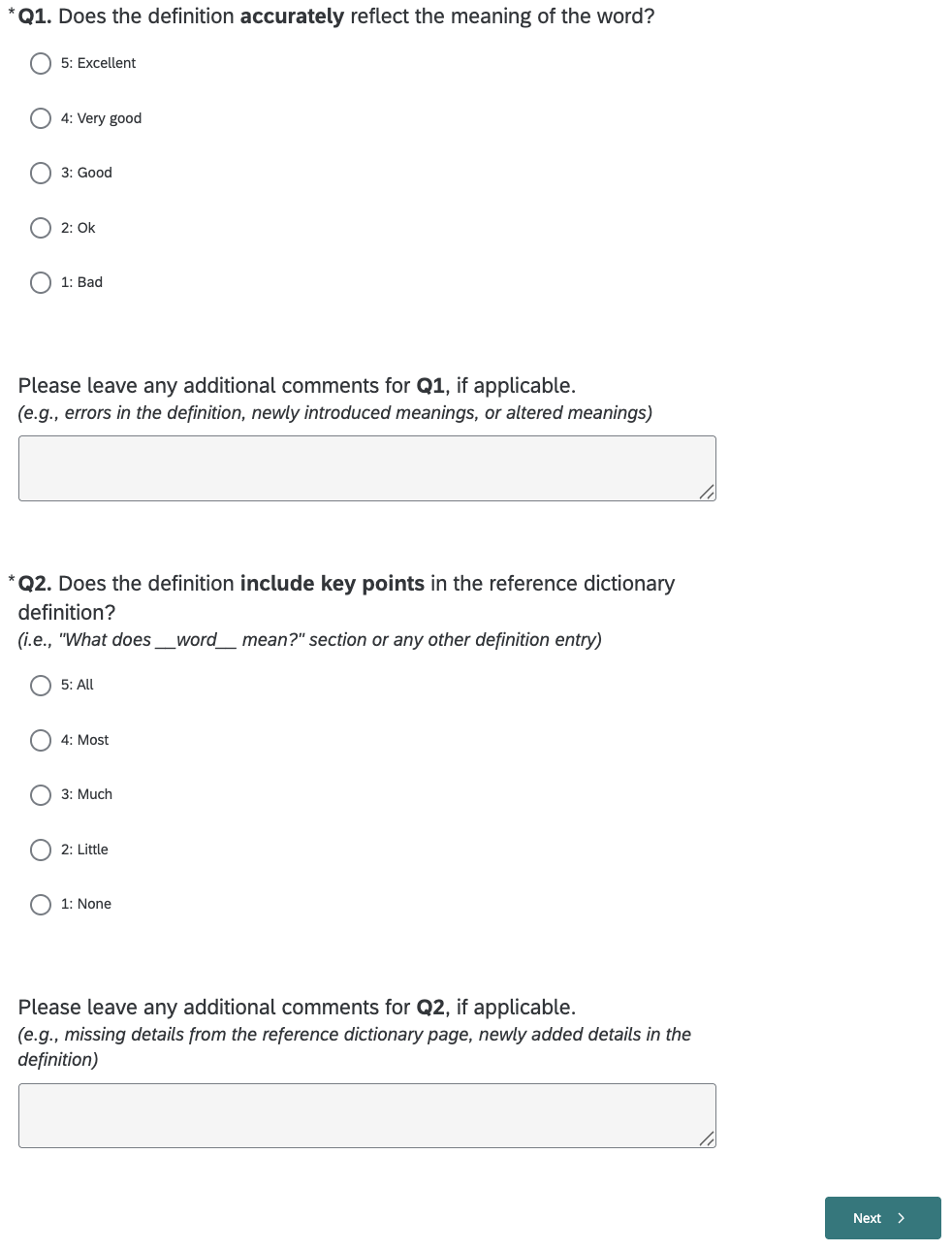}
        \end{minipage}
    }
\end{figure*}

\begin{table*}
\centering
\resizebox{\linewidth}{!}{%
    \begin{tabular}{p{2.5cm} p{5cm} p{5cm} p{5cm} p{9.5cm}}
    \toprule
    \textbf{Neologism} & \textbf{Social Media Post} & \textbf{AI Definition} & \textbf{AI Rewrite} & \textbf{AI Explanation} \\
    \toprule

    \textbf{brain rot} & So yeah, I done got myself addicted to my first phone brain rot game. & A state of mental decline or reduced cognitive function, often attributed to excessive consumption of low-quality or mindless digital content, such as social media, memes, or online videos. & So yeah, I ended up getting myself addicted to my first mindless phone game. & Brain rot is internet slang used to describe the feeling of mental numbness or decline after consuming excessive, often low-quality or repetitive online content, such as endless social media scrolling, memes, or mindless videos. It is typically used in a humorous or self-deprecating tone among younger audiences, especially on platforms like TikTok, Twitter, or Discord. The term carries a negative connotation, implying that the content is ``rotting'' one's brain or making them less productive or intelligent. It is often used to jokingly criticize one's own habits or the addictive nature of certain online trends. \\
    \midrule
    
    \textbf{canon event} & i think my october canon event was someone spreading a rumor that i died and then having someone i knew for 3 days confess their love to me & A significant or pivotal occurrence, often unavoidable, that fundamentally shapes a person's character or life trajectory, typically referenced in online discussions to describe moments perceived as essential to someone's personal development or storyline. & I think the important turning point in my life in October was when someone spread a rumor that I died, and then someone I had only known for three days confessed their love to me. & Canon event refers to a significant or pivotal moment in someone's life that is seen as inevitable or necessary for their personal development, often drawing from storytelling where ``canon'' means official or essential to the plot. It is typically used humorously or ironically online, especially among younger audiences, to describe awkward, embarrassing, or formative experiences (like a bad breakup or a fashion disaster) that everyone seems destined to go through. The tone is usually lighthearted or self-aware, and the phrase often carries a sense of resigned acceptance, as if these moments are unavoidable parts of growing up or being human. \\
    \midrule
    
    \textbf{cheugy} & I don't care how cheugy it is, I'm sharing my favorite quotes! & A term used to describe someone or something that is out of date, trying too hard to be trendy, or no longer considered cool, especially in reference to millennial trends or aesthetics that have fallen out of favor. & I don't care how outdated or uncool it is, I'm sharing my favorite quotes! & Cheugy is a slang term used to describe someone or something that is outdated, trying too hard, or no longer considered cool—often in reference to trends, fashion, or behaviors that were once popular but now seem passé. It is typically used by younger people, especially Gen Z, to gently poke fun at millennials or anyone perceived as out of touch with current trends. The tone is usually playful or teasing rather than mean-spirited, and it often carries a sense of lighthearted judgment. Common situations include commenting on decor, clothing, or social media habits that feel behind the times. \\
    \midrule
    
    \textbf{copium} & Victors writing history is the biggest lie ever. Frequently the losers are the ones who feel the need to write the history because they are huffing extreme amounts of copium. & A fictional substance or concept humorously referred to as something people metaphorically ``inhale'' or rely on to cope with disappointment, failure, or unfavorable outcomes, especially by rationalizing or denying reality; often used online to mock or satirize someone perceived as making excuses. & Victors writing history is the biggest lie ever. Frequently the losers are the ones who feel the need to write the history because they are desperately trying to comfort themselves and cope with their loss. & Copium is a blend of ``cope'' and ``opium,'' used online to mockingly describe someone who is using denial or rationalizations to deal with disappointment or failure. It's often employed in a sarcastic or teasing tone, especially in gaming, sports, or political discussions when someone refuses to accept an unfavorable outcome. The term implies that the person is metaphorically ``inhaling copium'' to numb the pain of reality. Its intended audience is typically internet-savvy users familiar with meme culture. The connotation is generally dismissive or mocking, suggesting that the person is deluding themselves rather than facing facts. \\
    \midrule
    
    \textbf{crash out} & there's a dude crashing out on twitter because he can't understand why so many colleges found him stuck up and rejected him, it's so funny & To fall asleep suddenly and deeply, often due to exhaustion or fatigue, typically in an unplanned or informal setting. & There's a guy having a meltdown on Twitter because he can't understand why so many colleges thought he was arrogant and rejected him, it's so funny. & Crash out is an informal slang phrase used to describe falling asleep suddenly or passing out from exhaustion, often after a long day or intense activity. It is commonly used among friends or peers, especially in casual conversations, texts, or social media posts. The tone is relaxed and colloquial, with no negative connotations—it's typically used to convey relief or humor about being extremely tired. Typical situations include someone coming home late and immediately falling asleep, or dozing off unexpectedly at a party or while studying. \\
    \midrule
    
    \textbf{delulu} & They believe in and live by a sci fi book, I'm not surprised that they're this delulu and make up weird scenarios. & A slang term derived from ``delusional,'' used to describe someone who holds unrealistic or overly optimistic beliefs, often about themselves, situations, or relationships, especially in a humorous or self-aware manner. & They believe in and live by a sci-fi book, I'm not surprised that they're this delusional and make up weird scenarios. & Delulu is a playful slang term derived from ``delusional,'' often used online to describe someone who has unrealistic or overly optimistic beliefs, especially about relationships, celebrities, or fandoms. It is typically used in a lighthearted, self-aware, or teasing tone, rather than as a harsh insult. The term is popular among younger internet users, particularly in K-pop and stan communities, to poke fun at themselves or others for entertaining far-fetched fantasies. While it can be affectionate or humorous, it sometimes carries a mildly mocking connotation. \\
    \midrule
    
    \textbf{grindset} & I work a 9-5 on top of my art business and trying to ``grind'' and grow fast destroyed my creative spark in 2023. I am ready to be intentional, slow down, and leave the ``grindset'' in 2024. & A mindset characterized by relentless focus on hard work, productivity, and personal achievement, often associated with entrepreneurial or self-improvement culture. & I work a 9-5 on top of my art business and trying to work relentlessly and focus only on hustling and productivity destroyed my creative spark in 2023. I am ready to be intentional, slow down, and leave the mindset of constant hard work and hustle in 2024. & Grindset is a blend of ``grind'' and ``mindset,'' used to describe a mentality focused on relentless hard work, hustle, and self-improvement, often in the context of entrepreneurship or personal success. It is commonly used on social media, especially among young adults and those interested in business or self-development. The tone can be both aspirational and ironic—sometimes genuinely praising dedication, other times mocking the excessive glorification of nonstop work. The intended audience is typically ambitious individuals or those involved in ``hustle culture.'' Connotations range from admiration for determination to criticism of unhealthy work-life balance. \\
    \midrule
    
    \textbf{reheat nachos} & ``Reheating nachos'' is so funny because some nachos are somebody else's reheated nachos. Pharrell reheated Off The Wall \& prince's nachos for this sound. Lmao. & To attempt to restore previously prepared nachos to a desirable temperature and texture, typically using a microwave or oven, often resulting in diminished quality compared to their original state. & ``Reusing old ideas'' is so funny because some ideas are just someone else's reused ideas. Pharrell reused Michael Jackson's ``Off The Wall'' and Prince's ideas for this sound. That's hilarious. & Reheat nachos is used online to describe a situation or content that is being recycled, reused, or brought up again after losing its original appeal, much like reheating leftover nachos that are no longer fresh. It often carries a mildly humorous or dismissive tone, suggesting that the repeated topic or meme is stale or unoriginal. The phrase is typically aimed at internet-savvy audiences familiar with meme culture and online trends. It can be used to gently mock someone for bringing up old news or to comment on the lack of novelty in a discussion. \\
    
    \bottomrule
    \end{tabular}
}
\caption{\textbf{AI-based support conditions and social media posts for each neologism.} \textbf{AI Definition:} Formal dictionary definition of the neologism; \textbf{AI Rewrite:} Rewrites of the original social media posts into simpler English; \textbf{AI Explanation:} Natural language explanations of the neologism's meaning and usage.} 
\label{tab:support_conditions}
\end{table*}

\begin{figure*}
    \setcounter{subfigure}{2}
    \centering
    \subfigure[\textbf{Collect NS-produced Writing Sample}]{\includegraphics[width=0.8\textwidth]{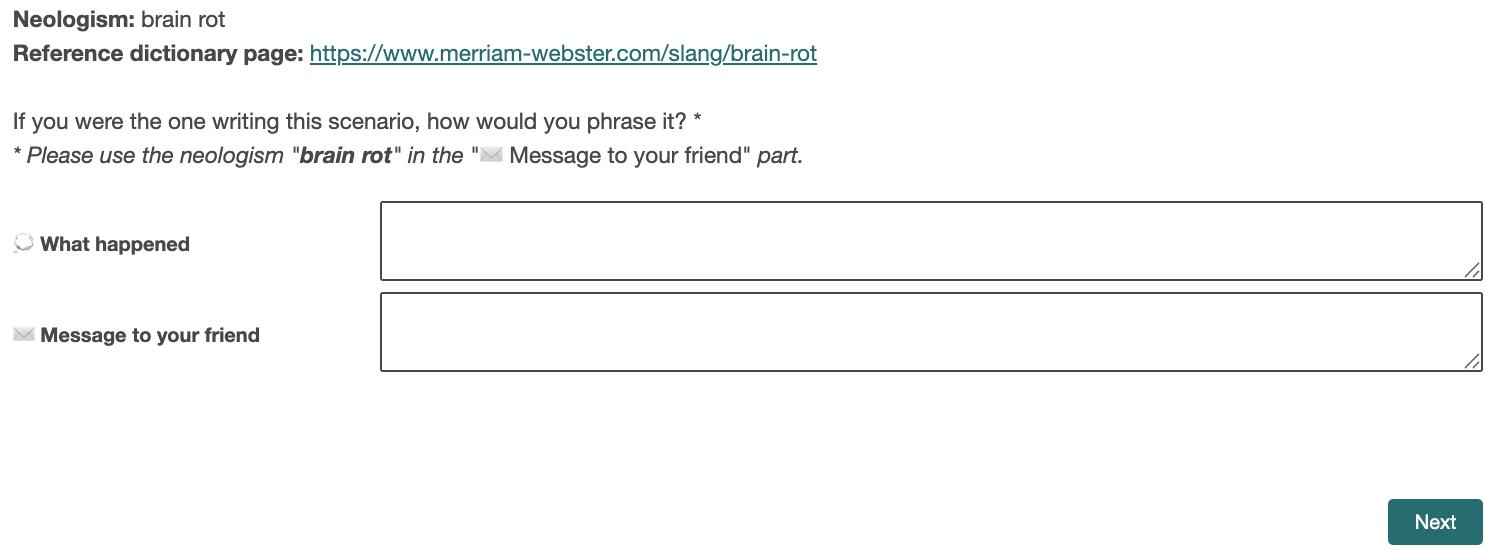}} 
    \subfigure[\textbf{Main Evaluation}]{
        \begin{minipage}{\linewidth}
            \centering
            \includegraphics[width=0.8\linewidth]{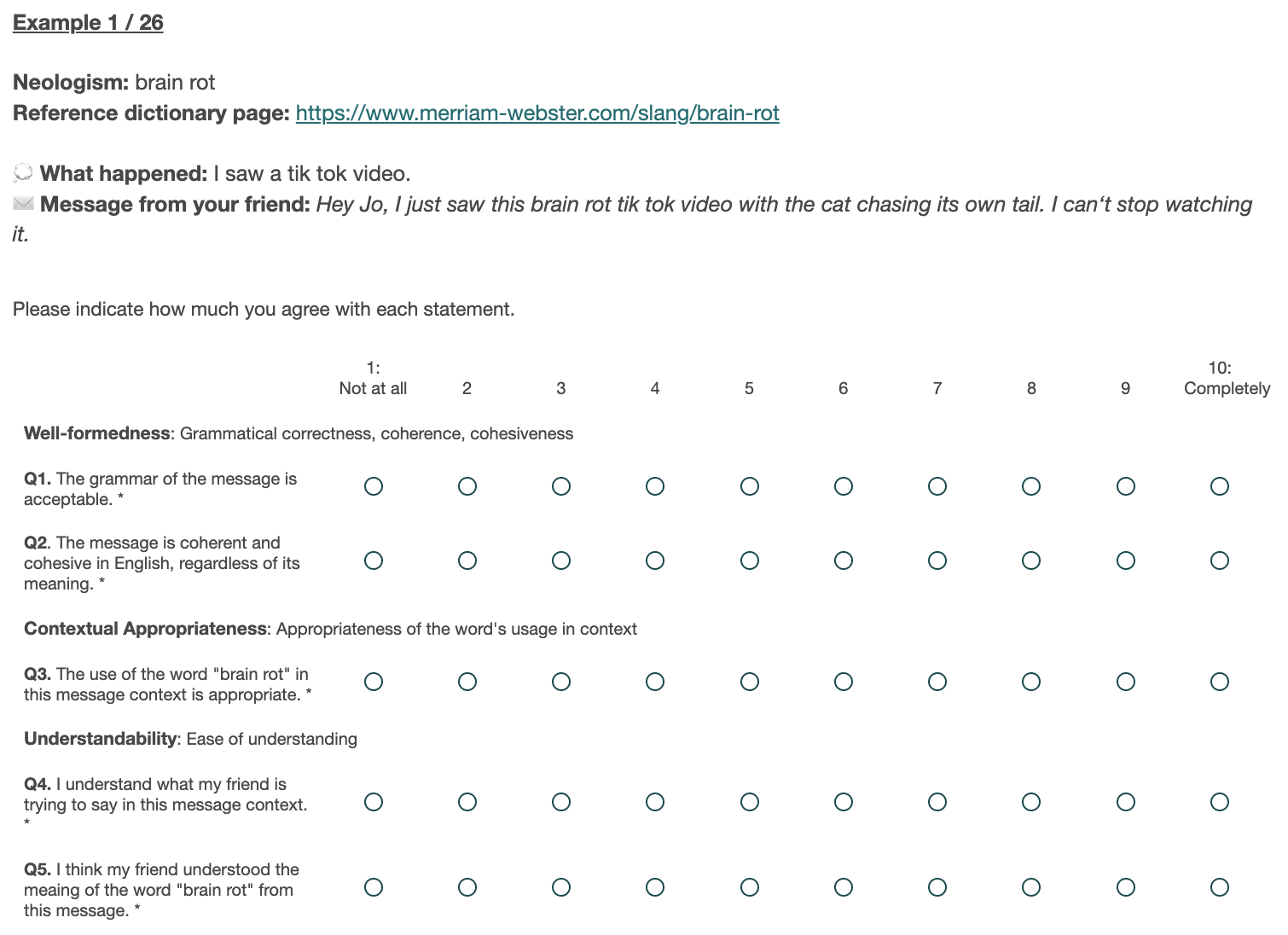}
            \includegraphics[width=0.8\linewidth]{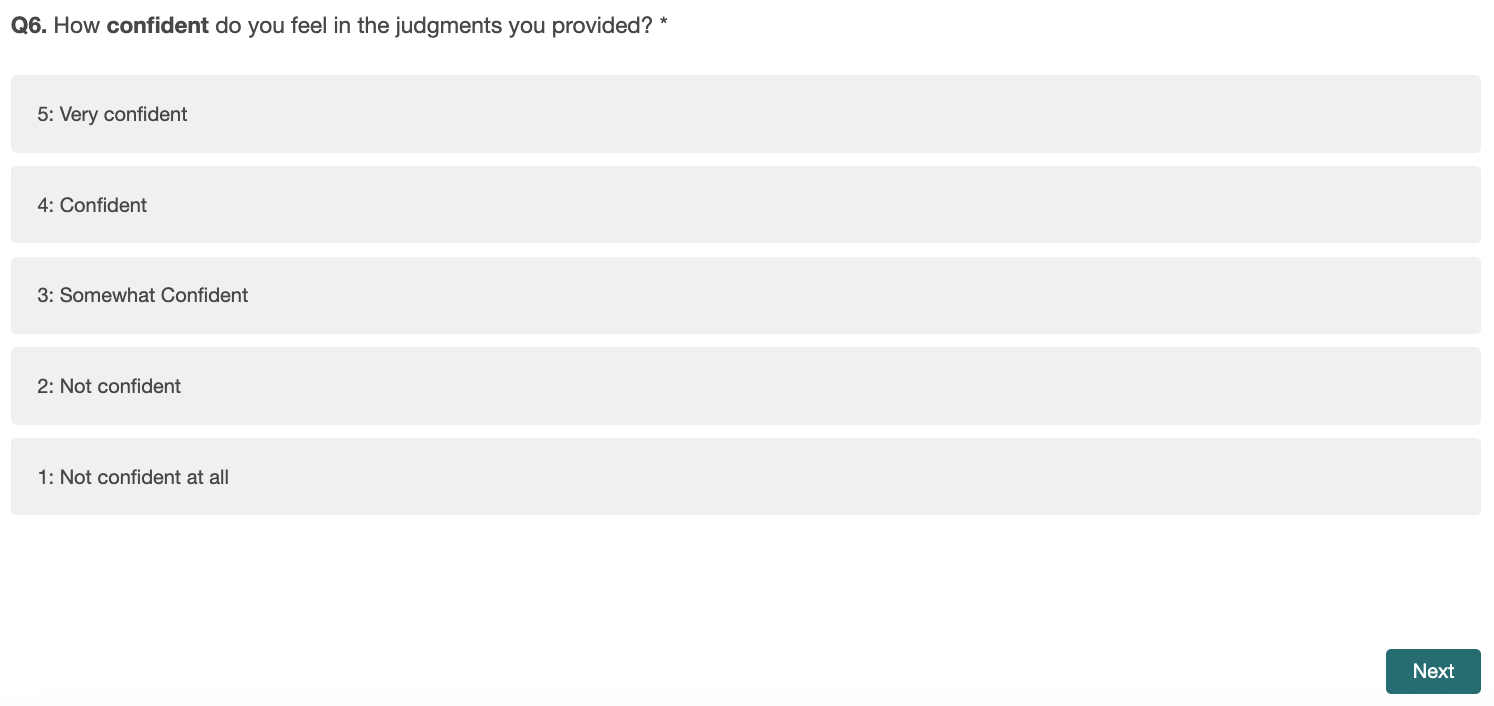}
        \end{minipage}
    }
\end{figure*}

\begin{table*}
\centering
\resizebox{\linewidth}{!}{%
    \begin{tabular}{p{4cm} p{7cm} p{7cm} p{7cm}}
    \toprule
    \textbf{Measures} & \textbf{Spanish} & \textbf{German} & \textbf{Chinese} \\
    \toprule

    \textbf{N} & 104 & 95 & 35 \\
    \midrule
    
    \textbf{Condition Group} & 
    Control (16.3\%), AI Definition (19.2\%), AI Rewrite (22.1\%), AI Explanation (23.1\%), Non-AI Dictionary (19.2\%) & 
    Control (17.1\%), AI Definition (14.3\%), AI Rewrite (22.9\%), AI Explanation (20.0\%), Non-AI Dictionary (25.7\%) & 
    Control (18.9\%), AI Definition (26.3\%), AI Rewrite (15.8\%), AI Explanation (23.2\%), Non-AI Dictionary (15.8\%) \\
    \midrule
    
    \textbf{Native Language(s)} & Spanish (89.4\%), Spanish \& English (7.69\%), Spanish \& Catalan (2.88\%) & German (79.0\%), German \& English (4.21\%), German \& Russian (3.16\%), German \& Polish (2.11\%), German \& Bosnian (2.11\%), German \& Romanian (1.05\%), German \& Spanish (1.05\%), German \& Swiss (1.05\%), German \& Farsi (1.05\%), German \& Turkish (1.05\%), German \& Arabic (1.05\%), German \& Slovak (1.05\%), German \& Albanian (1.05\%), German \& Greek (1.05\%) & Chinese (77.1\%), Chinese \& Cantonese (8.57\%), Cantonese (5.71\%), Chinese \& Taiwanese (2.86\%), Chinese \& English (2.86\%), Chinese \& English \& French (2.86\%) \\
    \midrule
    
    \textbf{Age} & 28.9 (S.D.=5.91) & 27.8 (S.D.=7.05) & 30.5 (S.D.=7.21) \\
    \midrule
    
    \textbf{Gender} & Man (48.5\%), Woman (43.7\%), Non-binary (6.8\%), Prefer not to say (1.0\%) & Man (80.0\%), Woman (20.0\%) & Man (22.9\%), Woman (74.3\%), Prefer not to say (2.9\%) \\
    \midrule
    
    \textbf{Nationality} & Mexican (56.3\%), Spanish (21.4\%), Chilean (15.5\%), Mexican-American (1.9\%), American (1.9\%), Cuban-American (1.0\%), Honduran (1.0\%), Ecuadorian (1.0\%) & 
    German (87.4\%), Austrian (6.3\%), Swiss (2.1\%), German-Albanian (1.1\%), German-Greek (1.1\%), German-Spanish (1.1\%), Romanian (1.1\%) & 
    Chinese (54.3\%), Canadian (11.4\%), Taiwanese (8.6\%), Australian (8.6\%), British (5.7\%), Malaysian (5.7\%), American (1.0\%), Kongese (1.0\%) \\
    \midrule
    
    \textbf{English Proficiency} & 3.56 (S.D.=0.946) & 3.35 (S.D.=0.920) & 3.49 (S.D.=1.07) \\
    \midrule
    
    \textbf{Years} & 2.80 (S.D.=6.77) & 1.08 (S.D.=2.93) & 10.7 (S.D.=9.10) \\
    \midrule
    
    \textbf{Info. Source} & 
    AI tools (18.0\%), Translation tools (22.3\%), Dictionary (28.2\%), Reference/News (6.3\%), Social media (25.2\%) & 
    AI tools (24.4\%), Translation tools (30.5\%), Dictionary (25.0\%), Reference/News (3.7\%), Social media (14.0\%), Ask around (2.4\%) & 
    AI tools (37.5\%), Translation tools (25.0\%), Dictionary (18.8\%), Reference/News (7.8\%), Social media (9.4\%), Ask around (1.6\%) \\
    \midrule
    
    \textbf{Social Media} & 
    Habitually (81.6\%), Often (11.7\%), Sometimes (5.8\%), Rarely (1.0\%), Never (0.0\%) & 
    Habitually (65.3\%), Often (18.9\%), Sometimes (9.5\%), Rarely (4.2\%), Never (2.1\%) & 
    Habitually (40.0\%), Often (22.9\%), Sometimes (25.7\%), Rarely (11.4\%), Never (0.0\%) \\
    \midrule
    
    \textbf{Writing (Texting)} &
    Habitually (25.2\%), Often (19.4\%), Sometimes (26.2\%), Rarely (19.4\%), Never (9.7\%) &
    Habitually (24.2\%), Often (21.1\%), Sometimes (23.2\%), Rarely (16.8\%), Never (14.7\%) &
    Habitually (31.4\%), Often (28.6\%), Sometimes (25.7\%), Rarely (14.3\%), Never (0.0\%) \\
    \midrule

    \textbf{Writing (Academic)} &
    Habitually (15.5\%), Often (17.5\%), Sometimes (22.3\%), Rarely (20.4\%), Never (24.3\%) &
    Habitually (26.3\%), Often (14.1\%), Sometimes (23.2\%), Rarely (22.1\%), Never (13.7\%) &
    Habitually (22.9\%), Often (37.1\%), Sometimes (2.9\%), Rarely (14.3\%), Never (22.9\%) \\
    \midrule

    \textbf{Writing (Work)} &
    Habitually (33.0\%), Often (17.5\%), Sometimes (22.3\%), Rarely (15.5\%), Never (11.7\%) &
    Habitually (20.0\%), Often (24.2\%), Sometimes (25.3\%), Rarely (12.6\%), Never (17.9\%) &
    Habitually (51.4\%), Often (28.6\%), Sometimes (11.4\%), Rarely (2.9\%), Never (5.7\%) \\
    \midrule

    \textbf{Writing (Social Media)} &
    Habitually (39.8\%), Often (20.4\%), Sometimes (18.4\%), Rarely (15.4\%), Never (5.8\%) &
    Habitually (26.3\%), Often (28.4\%), Sometimes (15.8\%), Rarely (11.6\%), Never (17.9\%) &
    Habitually (8.6\%), Often (34.3\%), Sometimes (28.6\%), Rarely (17.1\%), Never (11.4\%) \\
    \midrule

    \textbf{Writing (Personal)} &
    Habitually (20.4\%), Often (24.3\%), Sometimes (24.3\%), Rarely (15.5\%), Never (15.5\%) &
    Habitually (10.5\%), Often (8.4\%), Sometimes (22.1\%), Rarely (28.4\%), Never (30.5\%) &
    Habitually (14.3\%), Often (11.4\%), Sometimes (31.4\%), Rarely (25.7\%), Never (17.1\%) \\
    
    \bottomrule
    \end{tabular}
}
\caption{\textbf{Pre-survey statistics of NNS participants for each language group.} \textbf{Info. Source}: Information sources when participants encounter an unfamiliar English neologism or slang term (includes free-form ``Others'' responses); \textbf{Social Media}: Monthly English social media usage; \textbf{Writing}: Monthly English writing frequency for each activity.} 
\label{tab:pre_survey}
\end{table*}

\definecolor{midgreen}{RGB}{96, 171, 117}

\begin{table*}
\centering
\resizebox{\linewidth}{!}{%
    \begin{tabular}{ 
        m{2.5cm}
        m{7cm}
        m{12.5cm}
        >{\centering\arraybackslash}m{2.5cm}
        >{\centering\arraybackslash}m{0.8cm}
        >{\centering\arraybackslash}m{0.8cm}
        >{\centering\arraybackslash}m{0.8cm}
    }
    \toprule
    \textbf{Neologism} & \textbf{Scenario} & \raisebox{-0.2em}{\includegraphics[height=1.1em]{figure/logo/email.png}} \textbf{Message to \textit{Jo}} & \textbf{Support} & \textbf{WF} & \textbf{CA} & \textbf{U} \\
    \toprule

    \textbf{brain rot} & 
    I use tiktok a lot of times in a day	& The videos I watched on tiktok are so \textbf{brainrot} that make me feel tired & \raisebox{-0.2em}{\includegraphics[height=1.1em]{figure/logo/control.png}} & \cellcolor{red!20} 1.5 & \cellcolor{red!20} 1 & \cellcolor{red!20} 3 \\
    \cmidrule{2-7}
    & Recently I noticed my attention span has decreased and I'm blaming social media.	& So I finally set a timer on all my socials, to make sure I spend less time engaging with \textbf{brain rot} content and more time in something productive. I promise you I will be more attentive, Jo & \raisebox{-0.2em}{\includegraphics[height=1.1em]{figure/logo/definition.png}} & \cellcolor{orange!20}4.5 & \cellcolor{orange!20}5 & \cellcolor{orange!20}5 \\
    \cmidrule{2-7}
    & I carry my phone with me all the time and kept checking on message updates for nothing.	& Hi Jo, I am gonna uninstall Tiktok as it's such a \textbf{brain rot} APP and disrupting my daily work. & \raisebox{-0.2em}{\includegraphics[height=1.1em]{figure/logo/explanation.png}} & \cellcolor{midgreen!30}9 & \cellcolor{midgreen!30}10 & \cellcolor{midgreen!30}10 \\
    \cmidrule{2-7}
    & I spent my entire Sunday afternoon scrolling through short, meaningless videos on social media instead of doing the yard work I had planned.	& I really need to get off my phone. I spent the entire day consuming so much \textbf{brain rot} there there wasn't enough room in my mind for me to remember to take care of the yard work that I had been planning to get done all week. & NS & \cellcolor{midgreen!30}10 & \cellcolor{midgreen!30}10 & \cellcolor{midgreen!30}10 \\
    \midrule

    \textbf{canon event} & 
    doing wrong exercises in the gym is a canon event as a beginner.	& Hey Jo, i looked at your gym plan and there are some exercises i would change. But dont worry its a \textbf{canon event} for beginners. & \raisebox{-0.2em}{\includegraphics[height=1.1em]{figure/logo/definition.png}} & \cellcolor{red!20}2 & \cellcolor{red!20}1 & \cellcolor{red!20}1.5 \\
    \cmidrule{2-7}
    & Describing my health concerns	& I'll be getting my fourth and fifth eye surgeries next month. I guess I can look at it as a \textbf{canon event} & \raisebox{-0.2em}{\includegraphics[height=1.1em]{figure/logo/rewrite.png}} & \cellcolor{orange!20}5 & \cellcolor{orange!20}6 & \cellcolor{orange!20}5 \\
    \cmidrule{2-7}
    & At the gym today, I tripped over a yoga mat right in front of everyone, and instead of being embarrassed, I just laughed it off because it felt like one of those life-defining a "funny momment"	& I totally wiped out at the gym today—guess that’s my \textbf{canon event} for December! & \raisebox{-0.2em}{\includegraphics[height=1.1em]{figure/logo/dictionary.png}} & \cellcolor{midgreen!30}8.5 & \cellcolor{midgreen!30}9 & \cellcolor{midgreen!30}9 \\
    \cmidrule{2-7}
    & this one time I was so late for a class, that I ran into my professor while trying to get to the class, my books fell, and that drew a lot of attention to me, I ended up getting scolded by him which was embarrassing as I had managed to get the attention of a lot of my peers.	& getting scolded by my professor in front of my peers was my \textbf{canon event}, since then I make sure I am always 10 minutes early to every class. & NS & \cellcolor{midgreen!30}10 & \cellcolor{midgreen!30}10 & \cellcolor{midgreen!30}10 \\
    \midrule

    \textbf{cheugy} & 
    i was so cheugy at the wedding of our friend	& hey jo i was so \textbf{cheugy} at the weeding of brian, did you noticed that? & \raisebox{-0.2em}{\includegraphics[height=1.1em]{figure/logo/rewrite.png}} & \cellcolor{red!20}1 & \cellcolor{red!20}3 & \cellcolor{red!20}1 \\
    \cmidrule{2-7}
    & Talking about a gym member	& Hey Jo, i noticed a guy in my gym wearingcrazy gym outfits every day, freaking out about his training plan and protein intake.  Thats just way too \textbf{cheugy} imo. Just train and stop overthinking dude! & \raisebox{-0.2em}{\includegraphics[height=1.1em]{figure/logo/control.png}} & \cellcolor{orange!20}5.5 & \cellcolor{orange!20}4 & \cellcolor{orange!20}4 \\
    \cmidrule{2-7}
    & I went to a high school reunion and noticed some people still dressing and acting exactly like they did back then…	& Hey Jo, the reunion was sooo wild! Some outfits and hairstyles were so \textbf{cheugy}, it felt like time froze back in high school. & \raisebox{-0.2em}{\includegraphics[height=1.1em]{figure/logo/control.png}} & \cellcolor{midgreen!30}10 & \cellcolor{midgreen!30}10 & \cellcolor{midgreen!30}10 \\
    \cmidrule{2-7}
    & We were at a cafe and someone played an old motivational playlist while posting a quote graphic with cursive fonts and hashtags from 2016.	& I didn't want to say anything in the moment, but the playlist and quote posts felt a little \textbf{cheugy}, like trying too hard to be inspirational instead of just being natural. & NS & \cellcolor{midgreen!30}10 & \cellcolor{midgreen!30}10 & \cellcolor{midgreen!30}10 \\
    \midrule
    
    \textbf{copium} & 
    i am going home after a concert & hey wanna hang out? i need a way to \textbf{copium} with the post concert depression & \raisebox{-0.2em}{\includegraphics[height=1.1em]{figure/logo/rewrite.png}} & \cellcolor{red!20}3.5 & \cellcolor{red!20}1 & \cellcolor{red!20}2 \\
    \cmidrule{2-7}
    & i need to tell someone about my job interview	& Hello Jo, so my job interview went a little strange i was really unprepared for the amount of \textbf{copium} in the room. & \raisebox{-0.2em}{\includegraphics[height=1.1em]{figure/logo/control.png}} & \cellcolor{orange!20}5.5 & \cellcolor{orange!20}5 & \cellcolor{orange!20}4 \\
    \cmidrule{2-7}
    & Jo and I are talking about our favorite singer's last concert	& Your are on a dose of \textbf{copium} my friend, his last concert was so ugly I cannot accept we wasted our savings on someone like him & \raisebox{-0.2em}{\includegraphics[height=1.1em]{figure/logo/explanation.png}} & \cellcolor{midgreen!30}10 & \cellcolor{midgreen!30}10 & \cellcolor{midgreen!30}10 \\
    \cmidrule{2-7}
    & I was walking outside when some children started throwing eggs at me! Everyone around me was laughing as I shouted them down and showed how unruly and poorly behaved they were!	& That sounds like pure \textbf{copium} on your part. They were laughing because you lost your cool dealing with admittedly unruly children. You definitely don't come out of this looking like the winner dude. & NS & \cellcolor{midgreen!30}9.5 & \cellcolor{midgreen!30}10 & \cellcolor{midgreen!30}10 \\
    \midrule

    \textbf{crash out} & 
    Was partying with friends and one went totally crash out when his favorite artist entered the stage	& Was partyin with Tom and he went \textbf{crash out} as Jason Derulo entered the stage & \raisebox{-0.2em}{\includegraphics[height=1.1em]{figure/logo/explanation.png}} & \cellcolor{red!20}1 & \cellcolor{red!20}1 & \cellcolor{red!20}1 \\
    \cmidrule{2-7}
    & Some guy lost the TCG Pokemon tournament and started yelling and screaming to everyone that it was rigged &	hey Jo, I just came back from the TCG Pokemon tournament! I had a great time but there was a dude that started legit \textbf{crashing out} when he lost & \raisebox{-0.2em}{\includegraphics[height=1.1em]{figure/logo/explanation.png}} & \cellcolor{orange!20}5 & \cellcolor{orange!20}6 & \cellcolor{orange!20}7 \\
    \cmidrule{2-7}
    & A freelancer that works with us delivered poor work (again)	& Jo you wont believe how bad the footage was this guy send us today. I nearly \textbf{crashed out} in front of my boss. & \raisebox{-0.2em}{\includegraphics[height=1.1em]{figure/logo/dictionary.png}} & \cellcolor{midgreen!30}10 & \cellcolor{midgreen!30}10 & \cellcolor{midgreen!30}10 \\
    \cmidrule{2-7}
    & My professor announced at the very end of class that the midterm was moved up by a full week, even though we already have two other exams.	& I was fine all day, but when my professor moved the midterm up a week at the last minute, I actually \textbf{crashed out} & NS & \cellcolor{midgreen!30}10 & \cellcolor{midgreen!30}10 & \cellcolor{midgreen!30}10 \\
    \midrule

    \textbf{delulu} & 
    Fritzchen got a new job and hopes to drive his new bosses Ferrari	& Hi there Jo! My old friend Fritz (the redhead) got a new job and is completely \textbf{delulu} about being allowed to drive his new bosses Ferrari to get an oil change. & \raisebox{-0.2em}{\includegraphics[height=1.1em]{figure/logo/definition.png}} & \cellcolor{red!20}1 & \cellcolor{red!20}1 & \cellcolor{red!20}1 \\
    \cmidrule{2-7}
    & Rabid fans crafting and basing their lifestyles on a fictional concept that's not even real.	& Hey Jo, I understand you are a Star Wars fan, as I am one too. But there is no need to go all \textbf{delulu} about it and live your life like a hermit. & \raisebox{-0.2em}{\includegraphics[height=1.1em]{figure/logo/rewrite.png}} & \cellcolor{orange!20}5.5 & \cellcolor{orange!20}5 & \cellcolor{orange!20}5.5 \\
    \cmidrule{2-7}
    & I discuss the latest news on the erratic behavior of Katy Perry.	& Jo, she's just \textbf{delulu}: no one in their right mind would feel heroic for being sent to the moon to record tiktok videos. & \raisebox{-0.2em}{\includegraphics[height=1.1em]{figure/logo/dictionary.png}} & \cellcolor{midgreen!30}10 & \cellcolor{midgreen!30}10 & \cellcolor{midgreen!30}10 \\
    \cmidrule{2-7}
    & A guy made eye contact with me and I thought it meant something	& Call me \textbf{delulu} but I made eye contact with a guy and I think hes my soulmate & NS & \cellcolor{midgreen!30}10 & \cellcolor{midgreen!30}10 & \cellcolor{midgreen!30}10 \\
    \midrule

    \textbf{grindset} & 
    A friend of mine had a long sickness and missed some weeks in school. No I tell Jo how he managed to learn all the stuff he had missed	& Hey Jo, you remember Carl? He had a real \textbf{grindset} after his infection, no he gets good grades! & \raisebox{-0.2em}{\includegraphics[height=1.1em]{figure/logo/definition.png}} & \cellcolor{red!20}1 & \cellcolor{red!20}1 & \cellcolor{red!20}1 \\
    \cmidrule{2-7}
    & I am launching a coaching course this fall. I have been working on the final details.	& Hey Jo, next week I am finaly launching my course, remember I you said I had that \textbf{grindset} look. Tanks for the encouraging. & \raisebox{-0.2em}{\includegraphics[height=1.1em]{figure/logo/explanation.png}} & \cellcolor{orange!20}5 & \cellcolor{orange!20}4 & \cellcolor{orange!20}3 \\
    \cmidrule{2-7}
    & I stayed up all night working on a uni project, skipped breakfast, and still went to the gym in the morning	& Bro I pulled an all nighter and still hit the gym pure \textbf{grindset} vibes, but I might pass out soon lol & \raisebox{-0.2em}{\includegraphics[height=1.1em]{figure/logo/dictionary.png}} & \cellcolor{midgreen!30}10 & \cellcolor{midgreen!30}10 & \cellcolor{midgreen!30}10 \\
    \cmidrule{2-7}
    & I got up super early, went to the gym, and then finished a big chunk of my project before most people were even awake.	& I woke up at 5am, hit the gym, then finished my project, \textbf{grindset} is kicking in & NS & \cellcolor{midgreen!30}10 & \cellcolor{midgreen!30}10 & \cellcolor{midgreen!30}10 \\
    
    \bottomrule
    \end{tabular}
}
\caption{\textbf{NNS- and NS-produced writing samples with NS-rated competence.} \textbf{WF:} Well-formedness; \textbf{CA}: Contextual appropriateness; \textbf{U:} Understandability. We color-code the ratings for \raisebox{3pt}{\colorbox{red!20}{ }} (low), \raisebox{3pt}{\colorbox{orange!20}{ }} (mid), and \raisebox{3pt}{\colorbox{midgreen!30}{ }} (high). NS denotes NS-produced writing.} 
\label{tab:detailed_examples}
\end{table*}

\begin{table*}
\centering
\resizebox{\linewidth}{!}{%
    \begin{tabular}{p{3.3cm} p{10cm} p{10cm}}
    \toprule
    \textbf{Condition} & \textbf{Helpful} & \textbf{Needed} \\
    \toprule

    \textbf{Control} 
    & N/A & \textbullet~Video explaining the context and examples about the words. \\
    & & \textbullet~More \textbf{examples}, especially ones showing how the word is used in casual conversations. \\
    & & \textbullet~AI tools (e.g., Google search or ChatGPT), especially in interactive settings. \\
    & & \textbullet~Explanation of the \textbf{context}. \\
    \midrule

    \textbf{AI Definition}
    & \textbullet~Concise explanation, not overwhelming or overly lengthy. & \textbullet~\textbf{Cross-reference} is needed to confirm. \\
    & \textbullet~Words are easily understandable, simple, short, and use basic concepts. & \textbullet~Include examples of usage and \textbf{context} (e.g., ``canon event'' is mostly used for a computer game/fantasy story plot). \\
    & \textbullet~Very much hit/miss, but when hit, nailed the concept according to own knowledge. & \textbullet~AI should also tell where the \textbf{origin} of the word, or the \textbf{metaphoric use} came from (e.g., ``copium''). \\
    & \textbullet~Definitions got me something in the right direction, but not really about the usage. & \textbullet~It could have been more \textbf{concise} since it sometimes over-explained. \\
    & \textbullet~Helpful to grasp general meaning quickly and gave good starting point for using new words in context. & \textbullet~30\% of the times it mistook the concept or explained it wrongly made me trust it little, after all I can't tell if it's lying or not, especially when it's a word I don't know anything about. \\
    \midrule

    \textbf{AI Rewrite}
    & \textbullet~Changed the neologisms to words that I know. & \textbullet~Perhaps present 2 alternative versions of rewrites. \\
    & \textbullet~Helped in understanding the slang tone and adapting the writing to sound natural for a native English speaker. & \textbullet~AI was too formal, I would prefer to have some \textbf{real people explanations}. \\
    & \textbullet~I think it did a good job of only changing the word in question, while keeping the rest of the statement. & \\
    \midrule

    \textbf{AI Explanation}
    & \textbullet~Helped a lot when mentioned the audiences of the words. & \textbullet~\textbf{Visual examples}. \\
    & \textbullet~Examples, colloquial language in explaining the meaning. & \textbullet~Perhaps also give dictionary definition. \\
    & \textbullet~Context and tone was helpful\textemdash it was helpful in explaining not only the meaning but also who uses the term, and the context in which it is used. & \textbullet~Want to search for \textbf{real world examples} because some didn't sound real. Ideally some \textbf{direct quotes} of more real life cases (like talking to a real person). \\
    & \textbullet~Good explanations, starting with the origin or etymology of the word, explaining in a couple of sentences and then talking about the usual use nowadays. & \textbullet~Sometimes it gave too many meanings and tried to be safe, like saying that words gets mostly used in a joking manner when in reality they are mostly used to be mean. \\
    & & \textbullet~It could be even shorter to reduce the information to a minimum and make the explanation even more effective. \\
    & & \textbullet~Could have included at least a couple of examples very different to each other to have a \textbf{good spectrum of the meaning}. \\
    & & \textbullet~I didn't like that in almost every explanation it mentioned that the words were used by ``younger audiences.'' \\
    \midrule

   \textbf{ Non-AI Dictionary}
    & \textbullet~Examples were really helpful. & \textbullet~Make it more \textbf{concise}. \\
    & \textbullet~Context, evolution of the term (origin) was really helpful. & \textbf{Actual personal explanation} from someone who uses those slang terms. \\
    & & \textbullet~Maybe specifically mention if the word is towards negative or positive \textbf{tone}? Some words will be confusing if they are used in sarcasm. \\
    & & \textbullet~More \textbf{cultural notes}, such as examples from social media, explanations of the playful/sarcastic tone slangs often carry (some section like ``Common scenarios'' or ``How people use this online'' would make it really easier to apply the word naturally in writing context). \\    

    \bottomrule
    \end{tabular}
}
\caption{\textbf{Feedback from NNS participants for each support condition.} \textbf{Helpful:} Aspects of the support participants found helpful; \textbf{Needed:} Aspects they felt were missing or would have improved the support. We do not ask the \textbf{Helpful} question for the Control group.}
\label{tab:detailed_comments}
\end{table*}

\begin{figure*}
    \setcounter{subfigure}{2}
    \centering
    \subfigure[\textbf{Pre-task Survey}]{\includegraphics[width=\textwidth]{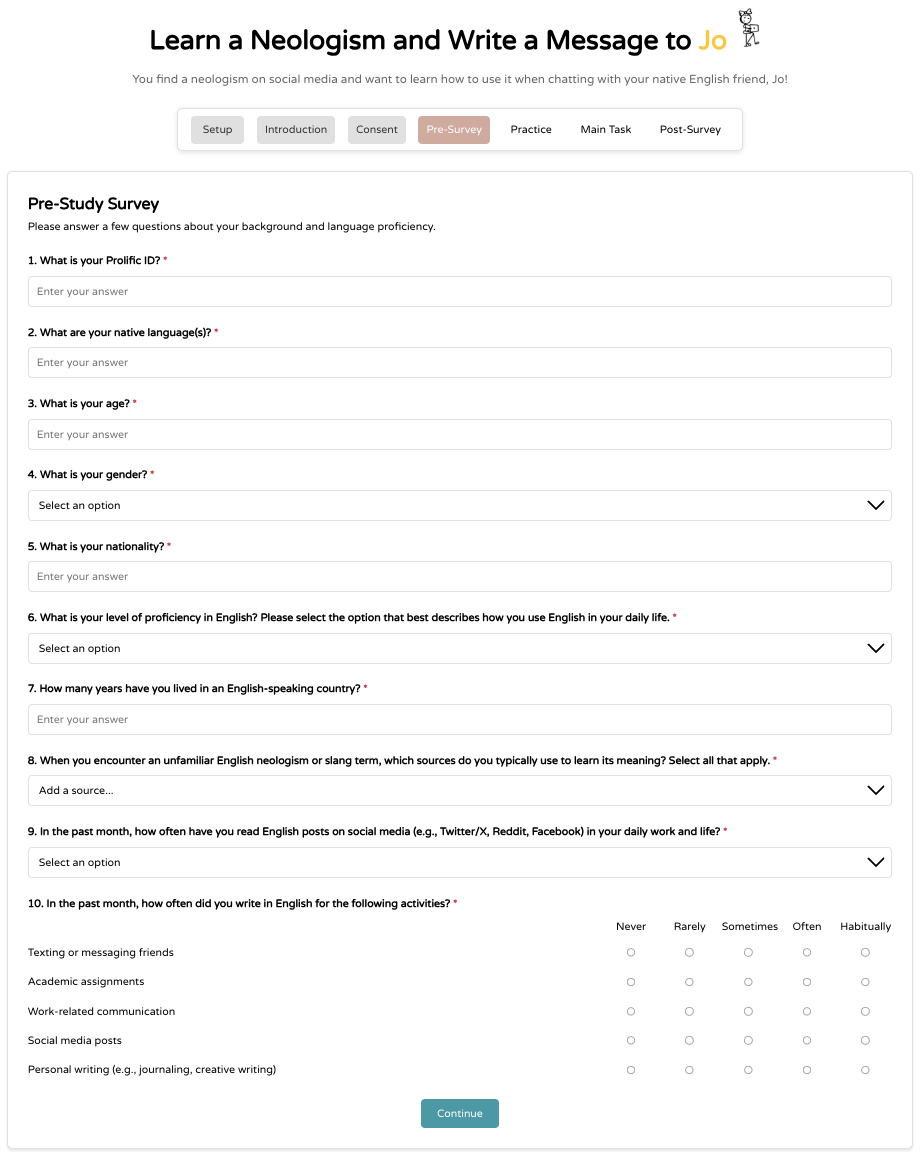}} 
\end{figure*}

\begin{figure*}
    \setcounter{subfigure}{3}
    \centering
    \subfigure[\textbf{Practice Session}]{\includegraphics[width=\textwidth]{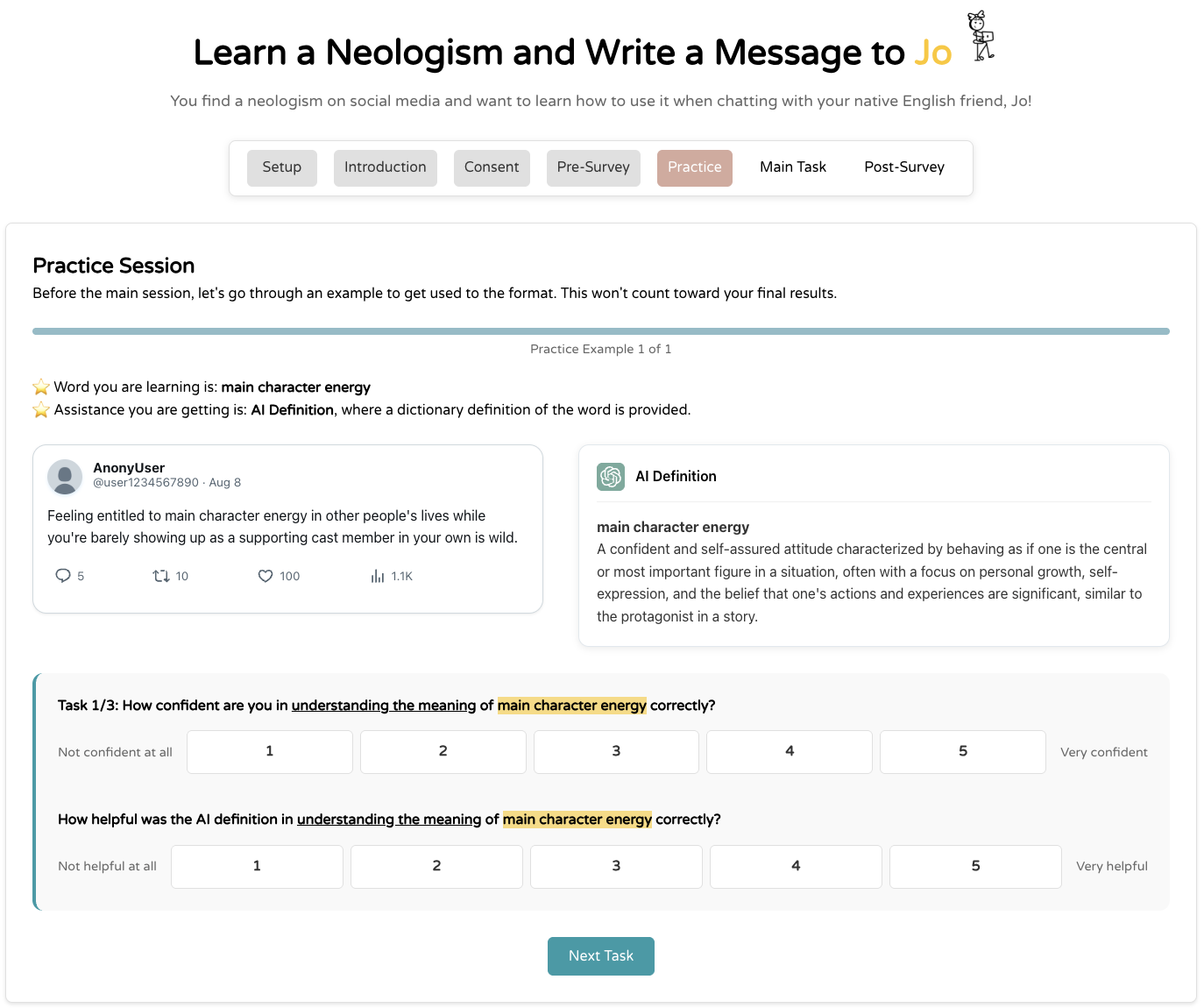}} 
\end{figure*}

\begin{figure*}
    \setcounter{subfigure}{4}
    \centering
    \subfigure[\textbf{Familiarity Check}]{\includegraphics[width=\textwidth]{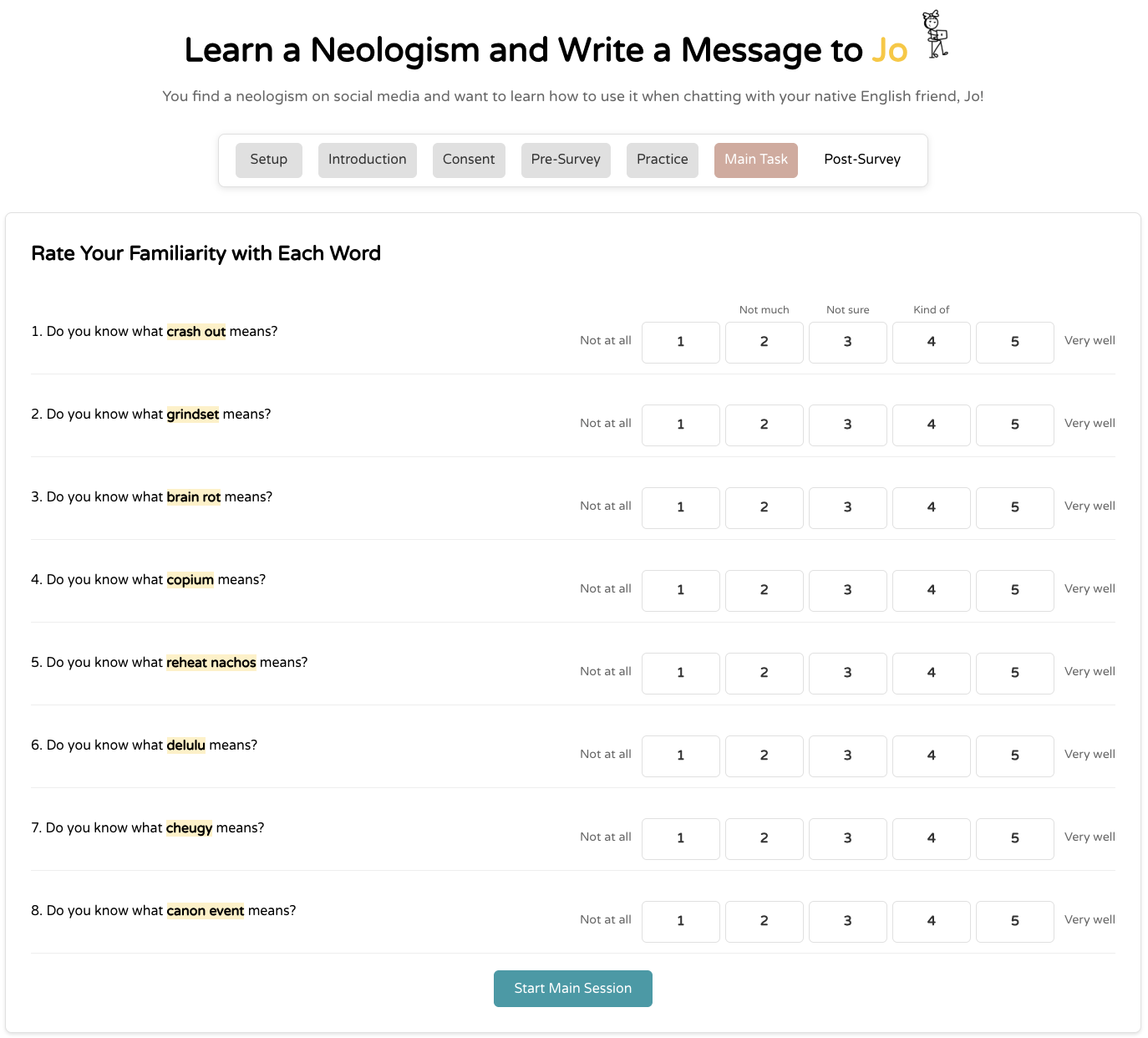}} 
\end{figure*}

\begin{figure*}
    \setcounter{subfigure}{5}
    \centering
    \subfigure[\textbf{Main Task (\ding{202} Learning)}]{\includegraphics[width=\textwidth]{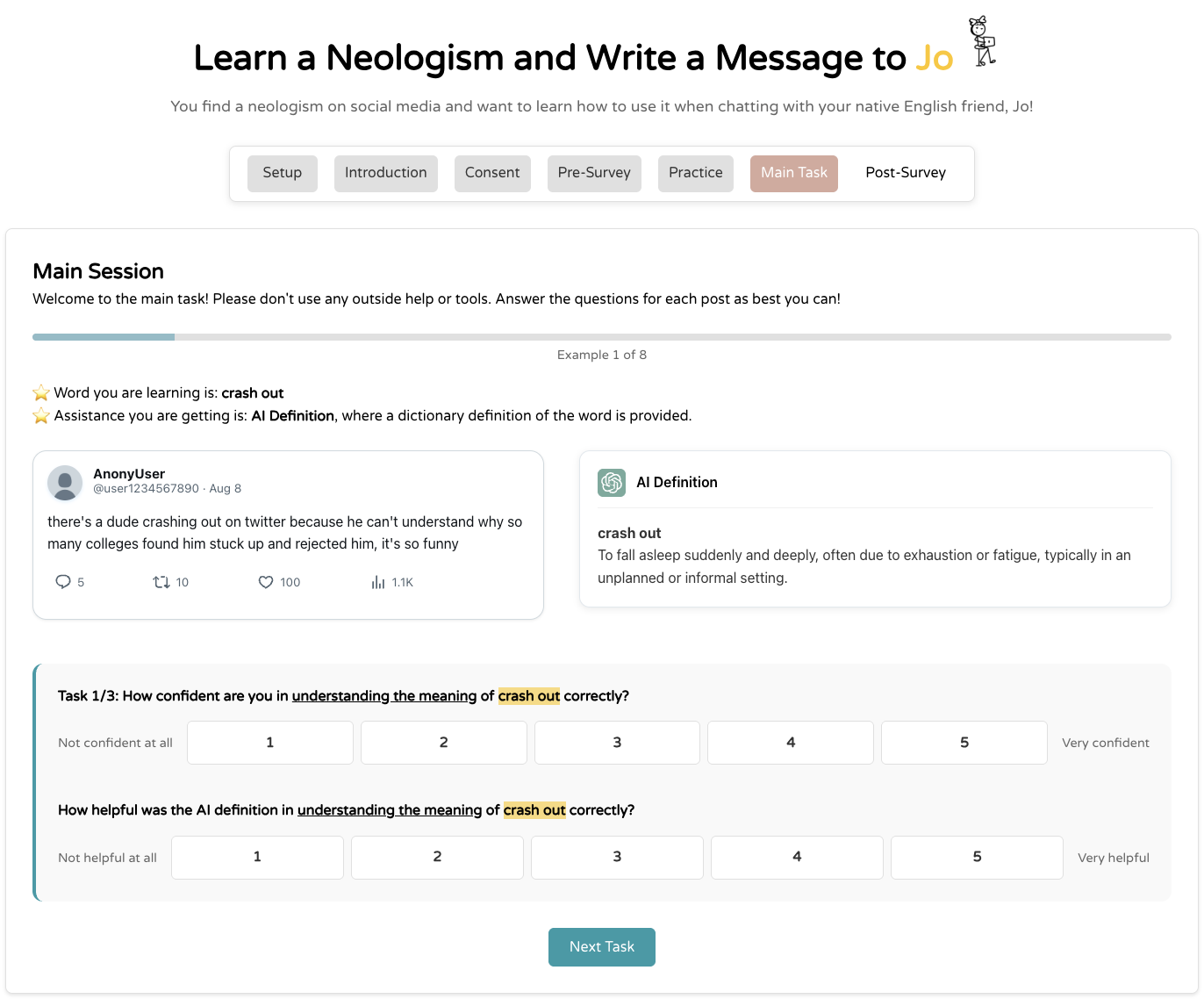}} 
\end{figure*}

\begin{figure*}
    \setcounter{subfigure}{6}
    \centering
    \subfigure[\textbf{Main Task (\ding{203} Production)}]{\includegraphics[width=\textwidth]{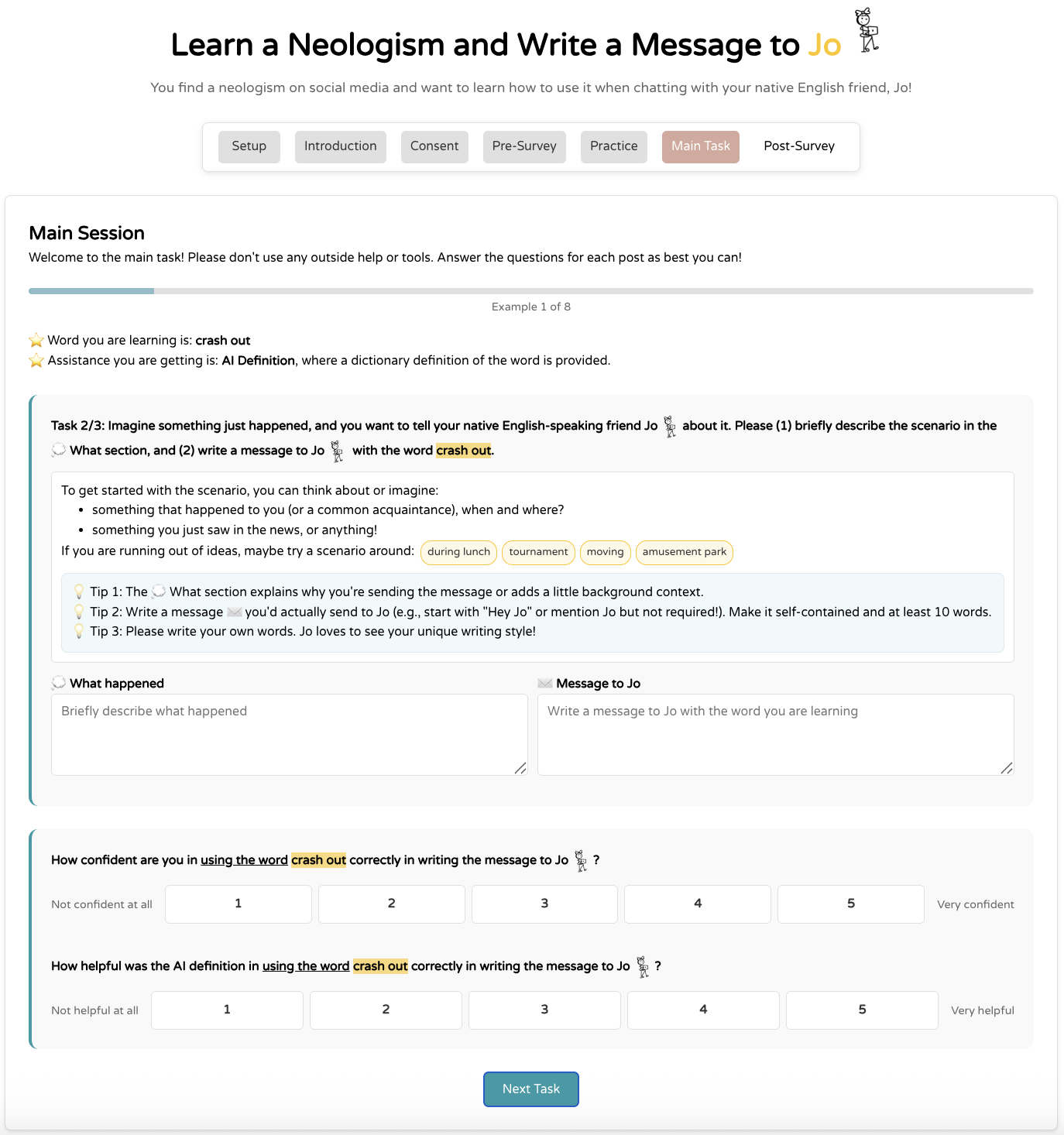}} 
\end{figure*}

\begin{figure*}
    \setcounter{subfigure}{7}
    \centering
    \subfigure[\textbf{Main Task (\ding{204} Comprehension)}]{\includegraphics[width=\textwidth]{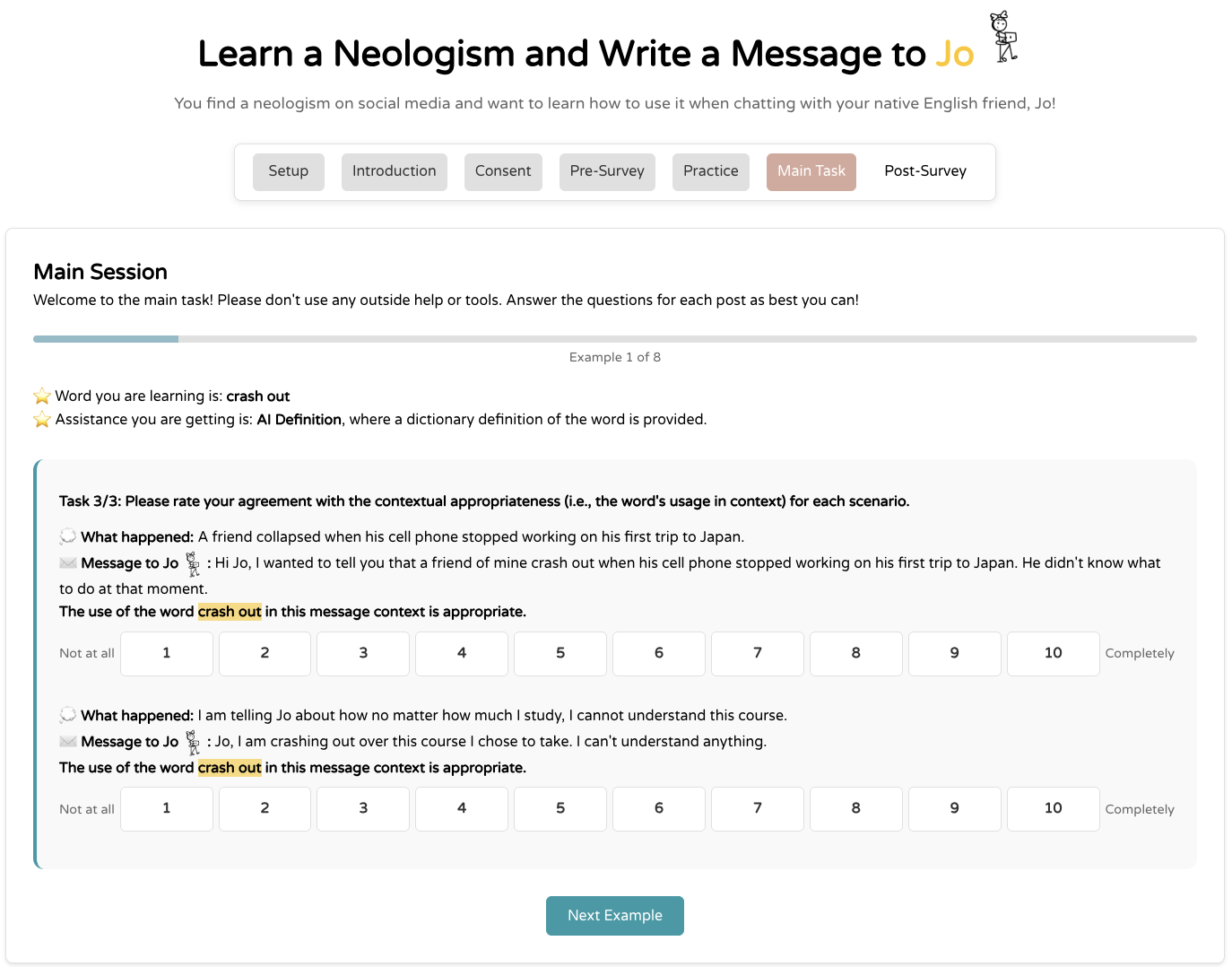}} 
    \subfigure[\textbf{Main Task (Attention Check)}]{\includegraphics[width=\textwidth]{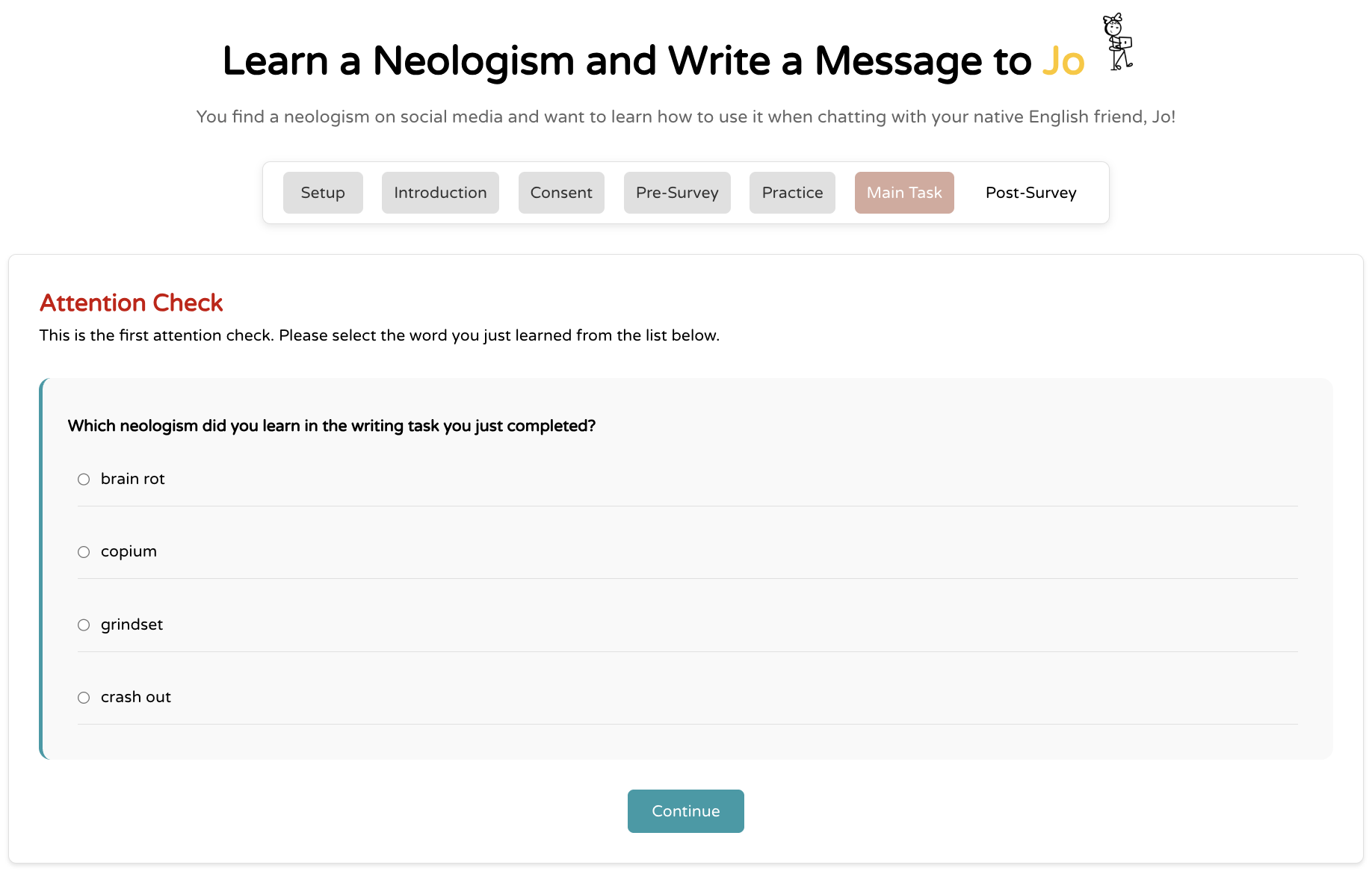}} 
\end{figure*}

\begin{figure*}
    \setcounter{subfigure}{9}
    \centering
    \subfigure[\textbf{Post-task Survey}]{\includegraphics[width=\textwidth]{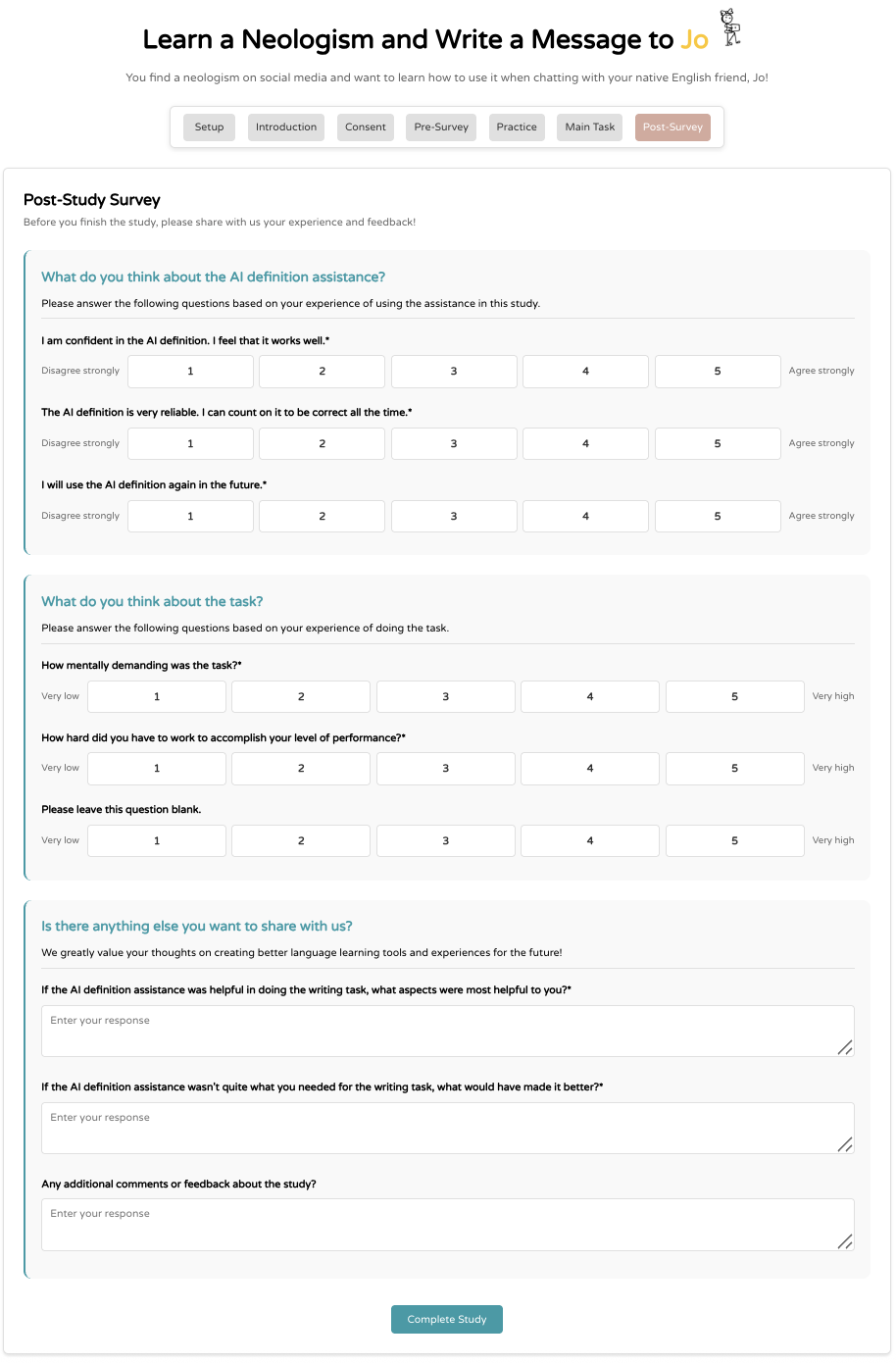}} 
\end{figure*}

\end{document}